\pdfoutput=1

\documentclass[journal]{IEEEtran}
\ifCLASSINFOpdf
\else
\fi

\usepackage{graphicx}
\usepackage{subfigure}
\usepackage{footmisc}
\usepackage{url}
\usepackage{amssymb}
\usepackage{diagbox}
\usepackage{multirow}
\usepackage{array}
\usepackage{booktabs}
\usepackage{epstopdf}
\usepackage{cite}

\begin{document}
%
\title{Automatic Pavement Crack Detection \\Based on Structured Prediction \\with the Convolutional Neural Network}
%
%
%

\author{Zhun~Fan,~\IEEEmembership{Senior~Member,~IEEE,}
        Yuming~Wu,
        Jiewei~Lu, and Wenji~Li
\thanks{Zhun~Fan, Yuming~Wu and Jiewei~Lu are with the Guangdong Provincial
Key Laboratory of Digital Signal and Image Processing, College of Engineering, Shantou University, Shan’tou 515063, China (email: zfan, 16ymwu1,
12jwlu1@stu.edu.cn).}}
\maketitle

\begin{abstract}
Automated pavement crack detection is a challenging task that has been researched for decades due to the complicated pavement conditions in real world. In this paper, a supervised method based on deep learning is proposed, which has the capability of dealing with different pavement conditions. Specifically, a convolutional neural network (CNN) is used to learn the structure of the cracks from raw images, without any preprocessing. Small patches are extracted from crack images as inputs to generate a large training database, a CNN is trained and crack detection is modeled as a multi-label classification problem. Typically, crack pixels are much fewer than non-crack pixels. To deal with the problem with severely imbalanced data, a strategy with modifying the ratio of positive to negative samples is proposed. The method is tested on two public databases and compared with five existing methods. Experimental results show that it outperforms the other methods.
\end{abstract}


%
\IEEEpeerreviewmaketitle

\section{Introduction}
%
%
%
%
\IEEEPARstart{P}{avement} distress detection is a significant step for pavement management \cite{zakeri2017image}. The goal is to obtain the pavement condition information for maintenance. Cracking is the most common type of pavement distress. Since automated crack detection systems are safer, lower costing, more efficient, and more objective, the research about them has attracted wide attention from both the academy and the industry \cite{zou2012cracktree, zhang2017efficient}.

Automated crack detection is a challenging task. In the past few decades, image based algorithms of crack detection have been widely discussed. In early studies, methods are mostly based on combination or improvement of conventional digital image processing techniques, like thresholding \cite{Oliveira2009Automatic}, mathematical morphology \cite{tanaka1998crack}, and edge detection \cite{zhao2010improvement}. These methods are generally based on photometric and geometric hypotheses about properties of crack images \cite{chambon2011automatic}. The most distinguished photometric property is that crack pixels are the darker pixels in an image. Based on that, threshold value is determined globally or locally to segment cracks and background \cite{tang2013automatic, Oliveira2009Automatic, li2008novel}. However, these approaches are very sensitive to noise as they are performed on individual pixels. In order to overcome this problem, other methods take geometric information into account. For instance, the continuity property of cracks is considered to reduce the false detection \cite{tanaka1998crack}. Local binary pattern operator is used to determine whether a pixel belongs to cracks based on the local orientation \cite{hu1local}. Wavelet transform is applied to separate crack regions and crack free regions with multi-scale analysis \cite{subirats2006automation}. These methods detect cracks efficiently, but they are not precise enough for finding all the cracks in an image. More recent works have been proposed to improve the accuracy of the detection method, and can be sorted into the following several branches.

\emph{Minimal Path Based Methods: } Minimal path problem is to find best paths between nodes of a graph. Several methods based on the minimal path principle are proposed. Kaul \emph{et al.} propose an algorithm that can find an open curve without knowing either the endpoints or topology of the curve \cite{kaul2012detecting}. Nguyen \emph{et al.} take brightness and connectivity into account simultaneously to provide features of anisotropic cracks along free-form paths \cite{nguyen2011free}. Amhaz \emph{et al.} propose strategies to select endpoints at a local scale and to select minimal paths at the global scale \cite{amhaz2016automatic}. These methods utilize distinguished features of crack pixels in a global view. However, the drawback is that they are too computationally intensive for practical applications.

\emph{Machine Learning: } With the development of machine learning, several methods focusing on feature extraction and pattern recognition have been proposed for crack detection \cite{delagnes1995markov, Oliveira2013Automatic, cord2012automatic, shi2016automatic}. In CrackIT \cite{Oliveira2013Automatic}, Oliveira \emph{et al.} use the mean and the standard deviation for unsupervised learning to distinguish blocks with crack from blocks without crack. Cord \emph{et al.} use AdaBoost to select textural descriptors that can describe crack images \cite{cord2012automatic}. In CrackForest \cite{shi2016automatic}, Shi \emph{et al.} propose a new descriptor based on random structured forests to characterize cracks. The performance of these methods are great but very dependent on the extracted features. However, due to complicated pavement conditions, it is hard to find features effective for all pavements.

\begin{figure*}[!t]
\centering
\includegraphics[width=17.8cm]{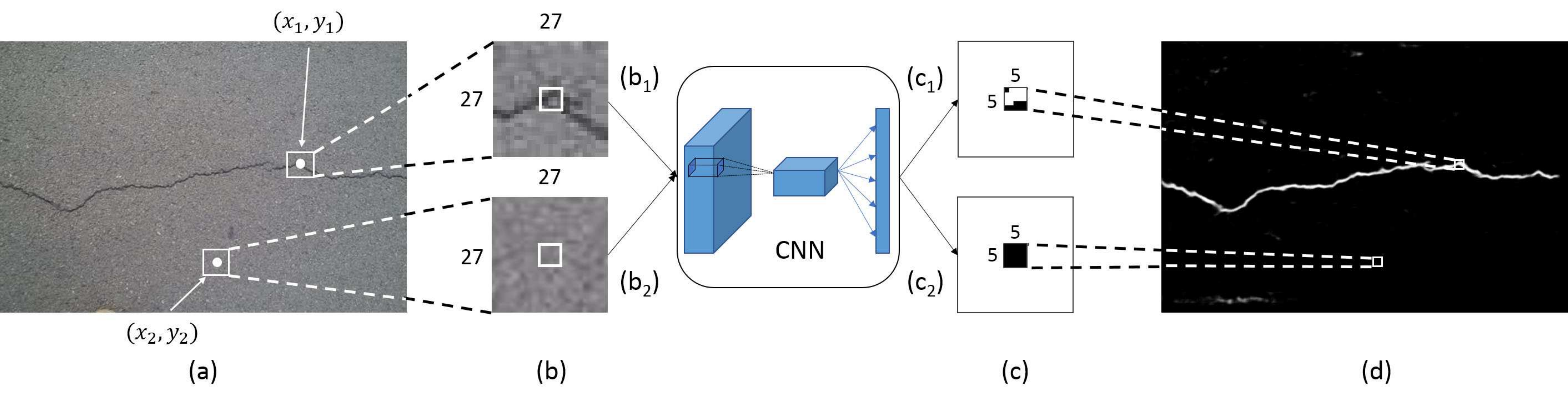}
\caption{Two examples of structured prediction based on CNN. (a) Two pixel examples of the original image. $(x_1,y_1), (x_2,y_2)$ are the coordinate of the pixel. (b) The extracted patches centered on the pixels. In the method the size of the patches is $27\times27$. (c) The structured prediction results of CNN with the patches input. (c$_1$) is obtained from (b$_1$) and (c$_2$) is obtained from (b$_2$). In the method the size of the output structure is $5\times5$. (d) The probability map obtained by applying structured prediction on all pixels.}
\label{Fig. structure_example}
\end{figure*}

\emph{Deep Learning: } For image classification tasks, deep learning has been proven to work better than traditional machine learning methods \cite{krizhevsky2012imagenet}. Recently, deep learning methods are successfully applied on damage and distress detection \cite{Cha2017Deep, Zhang2017Automated, gopalakrishnan2017deep, zhang2016road}. Cha \emph{et al.} use a sliding window to divide the image into blocks and CNN is used to predict whether the block contains cracks or not \cite{Cha2017Deep}. But the method can only find patch level cracks without considering the pixel level. In \cite{zhang2016road}, Zhang \emph{et al.} use CNN to predict whether an individual pixel belongs to crack based on the local patch information. However, the method ignores the spatial relations between pixels and overestimates crack width. In \cite{Zhang2017Automated}, Zhang \emph{et al.} use CNN to predict class for each pixel of the image. However, it still needs manually designed feature extractors for preprocessing, and CNN is only used as a classifier. Besides, its network architecture is strictly related to the input image size, which prevents generalization of the method.

Since CNN can extract good features from raw data, this paper propose a CNN based method that can learn the crack structure of a small patch within an image to find the whole crack on pixel level without preprocessing. Specifically, patches are exacted on each individual pixel of a raw image to build a training database. Then a CNN architecture is proposed for structured prediction as a multi-label problem \cite{liskowski2016segmenting}. Since crack pixels are much less than non-crack pixels in a typical image, applying structured prediction directly can hardly obtain a satisfied result. To solve the multi-label classification problem with imbalanced samples, we propose a strategy with modifying the ratio of positive to negative training samples. After training, the CNN can predict the centered structure of each input patch. Then all structure results of an image are summed to obtain a probability output, which reflects the condition of pavement generally, as shown in Fig.~1. Finally a decision threshold is set to obtain a binary output. The proposed method is compared with traditional methods like Canny \cite{canny1986computational} and local thresholding \cite{niblack1985introduction}, and the state-of-art methods like Free-Form Anisotropy (FFA) \cite{nguyen2011free}, CrackForest \cite{shi2016automatic}, and Minimal Path Selection (MPS) \cite{amhaz2016automatic}) to validate its effectiveness. The contributions of the paper are the following: 1) apply CNN based structured prediction on crack detection and obtain an outstanding performance; 2) propose a strategy to solve the multi-label classification problem with imbalanced samples and analyze its influence.

This paper is organized as the following. Section II describes the implementing details of the convolutional neural network, including data preparation, the architecture of CNN and training details. Section III gives the results and evaluations. Section IV describes the structured prediction in detail. Section V describes the influence of the imbalanced samples and the strategy to deal with it. The generalization of the method is discussed in Section VI. Conclusions and perspectives are given in Section VII.

\section{The Method}

\subsection{Data Preparation}

The proposed method is trained and tested on two databases: CFD\cite{shi2016automatic}\footnote{\url{https://github.com/cuilimeng/CrackForest-dataset}} with RGB images and AigleRN\cite{chambon2011automatic}\footnote{\url{https://www.irit.fr/~Sylvie.Chambon/Crack_Detection_Database.html}} with gray-level images. The CFD database contains 118 RGB images with a resolution of 320 $\times$ 480 pixels. All images are taken by an iphone5 from pavements of Beijing, China. These images contain noises such as shadows, oil spots and water stains, and have non-uniform illumination. We use 72 images for training and 46 images for testing. The AigleRN database contains 38 gray-level images, which have been collected on French pavement and have been pre-processed to reduce the non-uniform illumination. Compared to the images in CFD, the pavement images in AigleRN are with more complex texture. We use 24 images for training and 14 images for testing.

The input of the network are patches extracted from images based on each specific pixel, as shown in Fig.~2. In this work, we set the patch size as $27 \times 27$ and thus $h = 13$. Typically, RGB images with 3 channels and gray-level images with 1 channel are used as input of the network. So we propose a network with 3 channels input and a network with 1 channel input respectively.

\begin{figure}[!t]
\centering
\includegraphics[width=8.8cm]{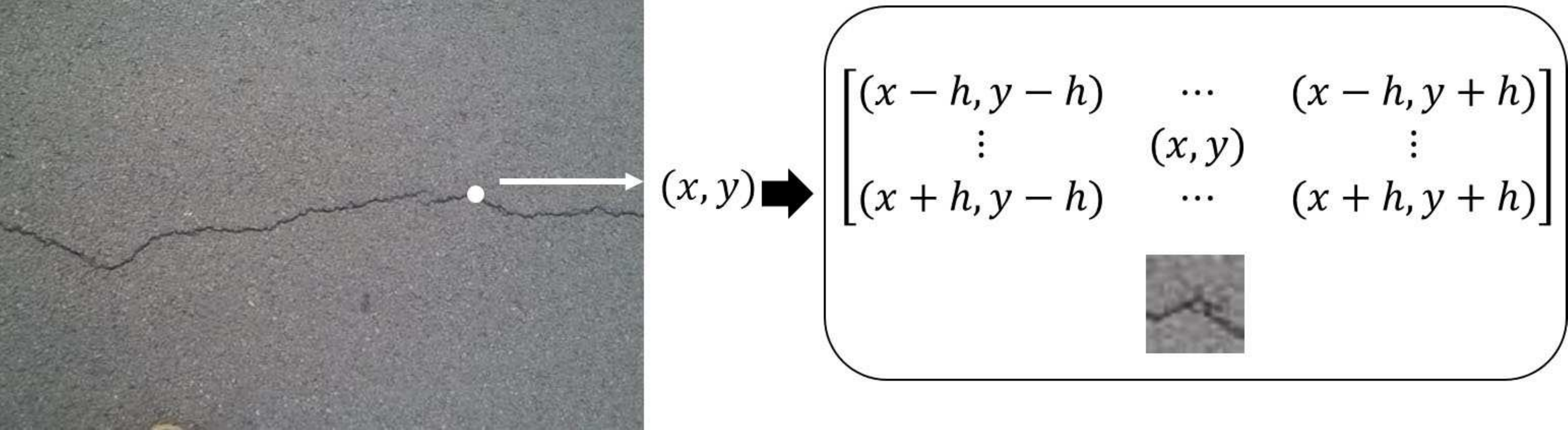}
\caption{The extracted patch centered on pixel $(x,y)$. $h$ is the distance between the edges and the center. Thus the size of the patch is $(2h+1)\times(2h+1)$.}
\label{Fig. preparation}
\end{figure}

In a typical crack image, non-crack pixels are a lot more than crack pixels. For a classification task with imbalanced samples, the proportion of positive and negative samples has a great impact on the network performance. Specifically, according to the manually labeled ground truth, if the center pixel of the extracted patch is a crack pixel, the corresponding patch is a positive sample and vice versa. To modify the proportion and keep the samples uniform, for training, all positive samples are extracted first. Then negative samples are extracted randomly according to the setting proportion. Generally, the ratio of positive to negative samples is set to 1:3, which is discussed in Section V. Then the training database can be obtained as shown in Fig.~3. In testing procedure, all pixels from an image are used and symmetric padding is used to deal with the boundary of the image. As a result, we can obtain all samples from an image in testing. For instance, 153,600 samples can be obtained in a 320$\times$480 image. Details of the training and testing database are shown in Table~I.

\begin{figure}[!t]
\centering
\includegraphics[width=4cm]{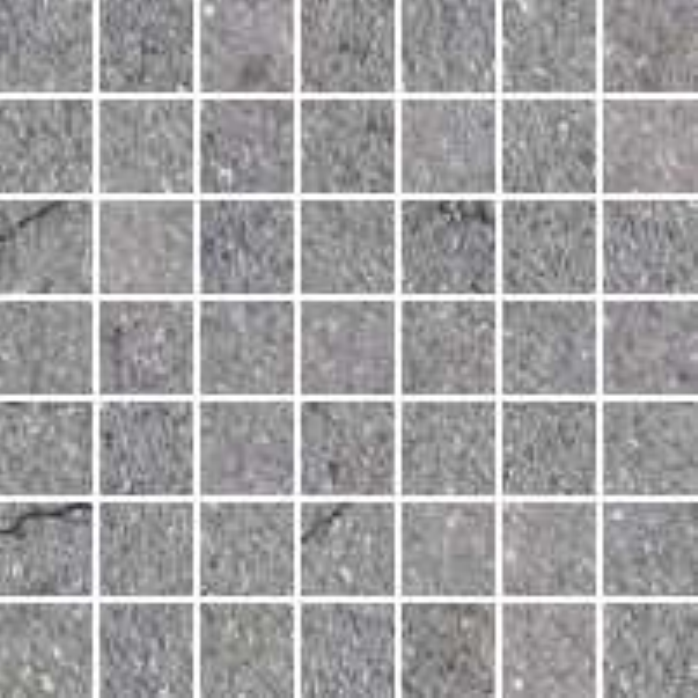}
\includegraphics[width=4cm]{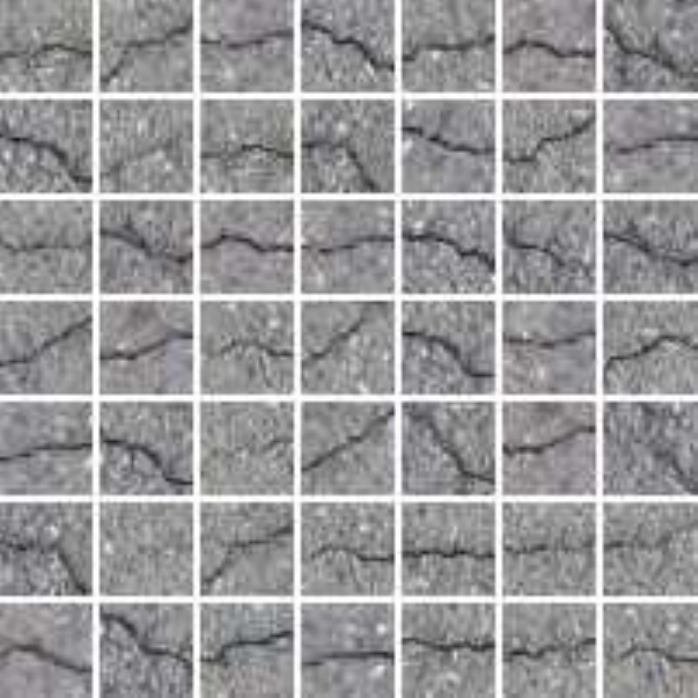}
\caption{Example of negative (left) and positive (right) $27 \times 27$ training patches extracted from the CFD database.}
\label{Fig. database}
\end{figure}

\begin{table}[!t]
\renewcommand{\arraystretch}{1.3}
\caption{Training Database}
\label{Table database}
\centering
\begin{tabular}{c| c c c c}
\hline
\space & \multicolumn{2}{c}{CFD} & \multicolumn{2}{c}{AigleRN}\\
\hline
Image dimensions & \multicolumn{2}{c}{320$\times$480} & \multicolumn{2}{c}{991$\times$462/311$\times$462}\\
\hline
\space & Training & Testing & Training & Testing\\
Images & 72 & 46 & 24 & 14\\
Positive patches & 185,483 & 107,112 & 39,208 & 42,317 \\
Negative patches & 556,449 & 6,958,488 & 117,624 & 4,168,351 \\
Total patches & 741,932 & 7,065,600 & 156,832 & 4,210,668 \\
Positive : Negative & 1:3 & 1:65 & 1:3 & 1:98.5\\
\hline
\end{tabular}
\end{table}

Data with zero mean and equal variance is preferred for optimization in machine learning. For image input, a common practice is scaling the raw data from [0, 255] to [-1, 1]. This preprocessing is adopted our work.

\subsection{Network Architecture}

Typical CNN contains convolutional layers, max-pooling layers and fully-connection (FC) layers.  The proposed network has 4 convolutional layers with 2 max-pooling layers and 3 FC layers. The architecture of the network is illustrated in Fig.~4 and Table II. Extracted patches are used as input of the network with 3 channels in CFD and 1 channel in AigleRN. All convolutional layers are equipped with kernel of $3\times3$ and stride of 1, whose effectiveness has been verified in VGG-Net \cite{simonyan2014very}. Zeros padding on the boundary of each convolutional layer is assigned to preserve the spatial resolution of resulting feature maps after convolution. Max pooling is performed with stride 2 over a $2\times2$ window.

\begin{figure*}[!t]
\centering
\includegraphics[width=16cm]{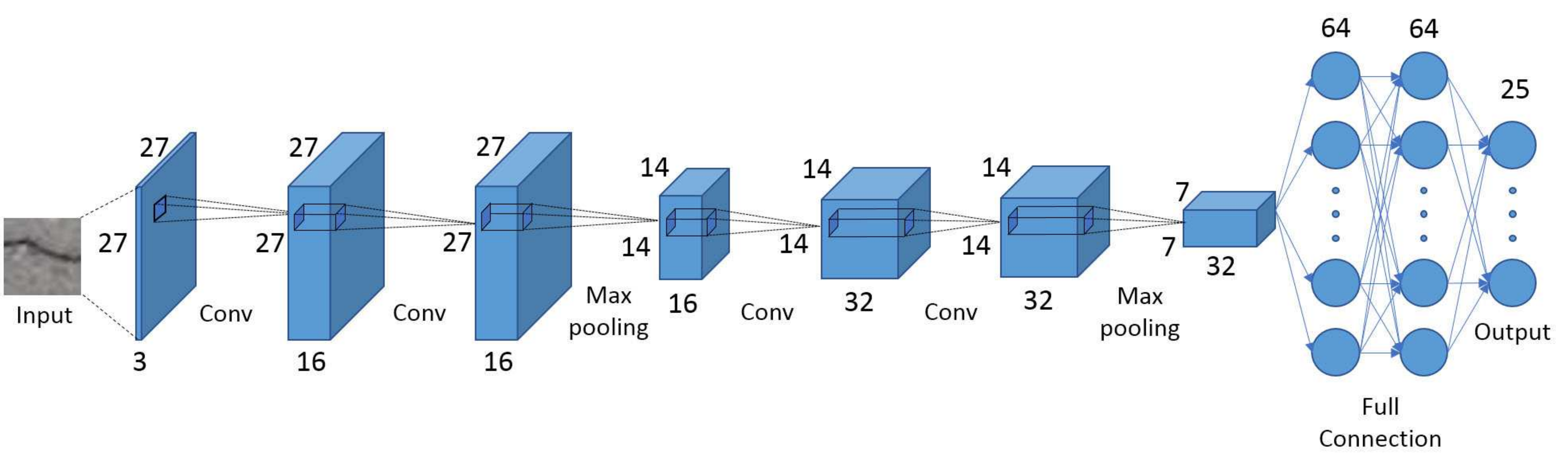}
\caption{An illustration of the CNN architecture. The leftmost image is the input patch with 3 channels. Other cubes indicate the feature maps obtained from convolution (Conv) or max pooling. All convolutional layers are with kernel of $3\times3$, stride 1 and zeros padding. Max pooling is performed with stride 2 over a $2\times2$ window.}
\label{Fig. CNN_architecture}
\end{figure*}

\begin{table*}[htbp]
\renewcommand{\arraystretch}{1}
\caption{Architecture of the CNN. Layer Names are Followed by Numbers of Feature Maps. \protect\\ Square Brackets Specify Kernel Size, Stride and Padding.}
\label{Table II}
\centering
\begin{tabular}{cp{0.1cm} c p{0.1cm}cp{0.1cm}cp{0.1cm}cp{0.1cm}cp{0.1cm}cp{0.1cm}cp{0.1cm}cp{0.1cm}c}
\toprule
\multirow{2}[0]{*}{Input} & \multirow{2}[0]{*}{$\rightarrow$} & conv16 & \multirow{2}[0]{*}{$\rightarrow$} & conv16 & \multirow{2}[0]{*}{$\rightarrow$} & maxpool & \multirow{2}[0]{*}{$\rightarrow$} & conv32 & \multirow{2}[0]{*}{$\rightarrow$} & conv32 & \multirow{2}[0]{*}{$\rightarrow$} & maxpool & \multirow{2}[0]{*}{$\rightarrow$} & \multirow{2}*{FC64} & \multirow{2}[0]{*}{$\rightarrow$} & \multirow{2}[0]{*}{FC64} & \multirow{2}*{$\rightarrow$} & \multirow{2}[0]{*}{FC25} \\ & & [3$\times$3,1,1] & & [3$\times$3,1,1] & & [2$\times$2,2,0] & & [3$\times$3,1,1] & & [3$\times$3,1,1] & & [2$\times$2,2,0]  \\
\bottomrule
\end{tabular}
\end{table*}

As shown in Fig.~1, the output of the network is a prediction of the structure centered in the input patch. For generalization, we use $s$ to denote the size of the output structure. Thus $s=5$ as shown in Fig.~1(c). The $s \times s$ window is flattened into $s^2$ neurons in the network, as the output layer of the CNN shown in Fig.~4, and is fully connected to the second last layer. Thus the structured prediction problem is modeled as an multi-label problem and $s^2$ labels are from the corresponding ground truth. Details about the multi-label problem is discussed in Section IV.

For a multi-label classification problem, sigmoid function is the commonly used activation function for the output. Except the output units, all hidden layers are equipped with Rectified Linear Units (ReLU) \cite{nair2010rectified} for dealing with nonlinearity.

\subsection{Training and Testing}
In a multi-label problem, all output units are not mutually exclusive, which means that more than one positive output may exist at the same time. In this case, cross entropy is commonly used in the loss function, which is defined as:
$$L=-\sum_{i=1}^{s^2}(y_i \log \hat{y_i} + (1-y_i) \log (1-\hat{y_i}))$$
where $y_i$ and $\hat{y_i}$ are the label and prediction of the $i$th output unit, respectively; $s^2$ is the number of labels. To prevent the network from overfitting, weight decay is adopted in the paper to penalize the large weights. The $L_2$ penalty term is added to the loss function and loss $L'$ used for optimization is:
$$L'=L + \beta \cdot \frac{1}{2} \sum_{j}{W_j^2}$$
where $L$ is the cross entropy function, $\beta$ is the $L_2$ penalty factor and $W_j$ is the $j$th weight in the network, including weights in convolutional layers and FC layers. $\beta$ is set to 0.0005 in the experiment.

Dropout \cite{Srivastava2014Dropout} is also a simple and effective approach for preventing overfitting. During training, value of hidden neurons is set to zero randomly. It is applied on the first two FC layers and the dropout ratio is set to 0.5.

Weights in convolutional layers and FC layers are initialized using Xavier method \cite{glorot2010understanding}. Adam \cite{Kingma2014Adam} is adopted as the optimizer with default setting in the original paper (learning rate set to 0.001). The batch size for each iteration is set to 256, and 20,000 iterations (43 epochs) in AigleRN and 30,000 iterations (13 epochs) in CFD are conducted.

In testing, each pixel is used for generating an input patch, so input patches overlap, leading to overlap of output windows. Typically, $s^2$ decisions can be obtained from a pixel except that it is near the boundary. Then for a testing image, all outputs from a pixel are summed and normalized to [0,1]. As a result, we can obtain a probability map from the results of all the pixels of an image.

\section{Results}

The network is programmed under TensorFlow. All experiments are conducted on the workstation with Intel Xeon E5-2690 2.9GHz CPU, 64GB RAM and Nvidia Quadro K5000 GPU.

\subsection{Evaluation}
We test the performance of the network on test images. For evaluation, precision ($Pr$), recall ($Re$) and F1 score ($F1$) are introduced, which are commonly used in classification problems, and defined as:
$$Pr= \frac{TP}{TP+FP}$$
$$Re= \frac{TP}{TP+FN}$$
$$F1= \frac{2\times Pr \times Re}{Pr+Re}$$
where $TP$, $FP$, $FN$ are the numbers of true positive, false positive and false negative respectively. Considering that the ground truth is subjective manually labels and there are transitional areas between crack pixels and non-crack pixels in real images, small distance between the detected result and the reference, which is 2 pixels in \cite{amhaz2016automatic} and 5 pixels in \cite{shi2016automatic}, is accepted for evaluation. We accept 2 pixels in the paper.

\subsection{Results on CFD}
Exemplar detections on CFD are shown in Fig.~5. The output of the network is a probability map as shown on the third row. Whiter color indicates that the pixel is more likely to be a crack pixel. Then the decision probability is set to 0.5 to remove the pixels with low probability, and a binary output is obtained. It can be observed that only few pixels are removed from binarization. The reason is that the contrast between the detected crack pixels and the detected non-crack pixels in the probability map is high. According to the output images in Fig.~5, our method can effectively deal with complex background and complex crack topology.

\begin{figure}[!t]
\centering
\includegraphics[width=2.8cm]{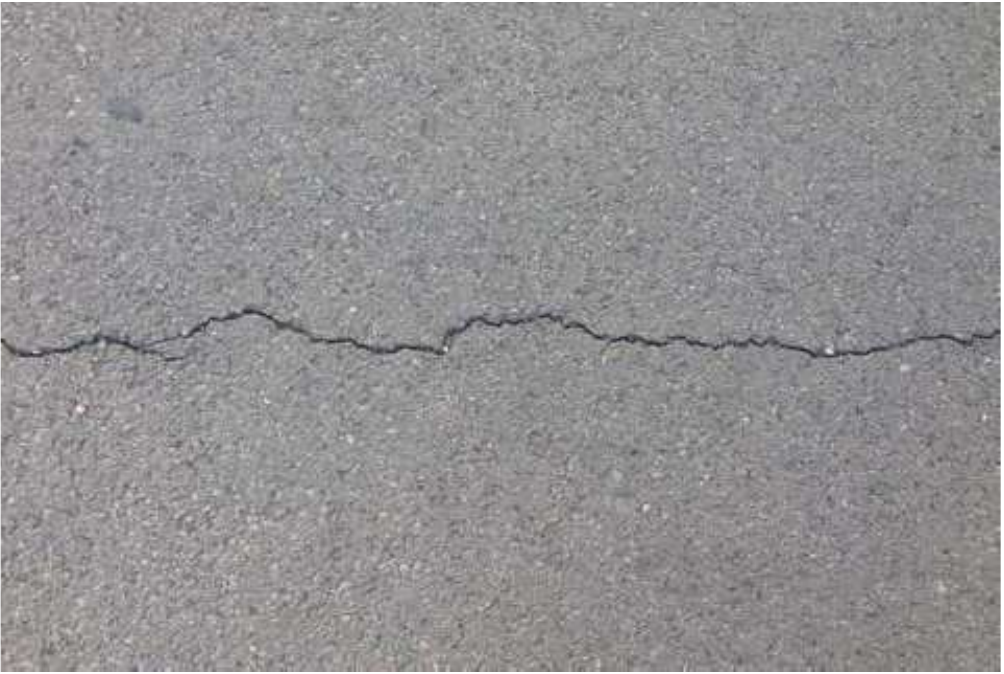}
\includegraphics[width=2.8cm]{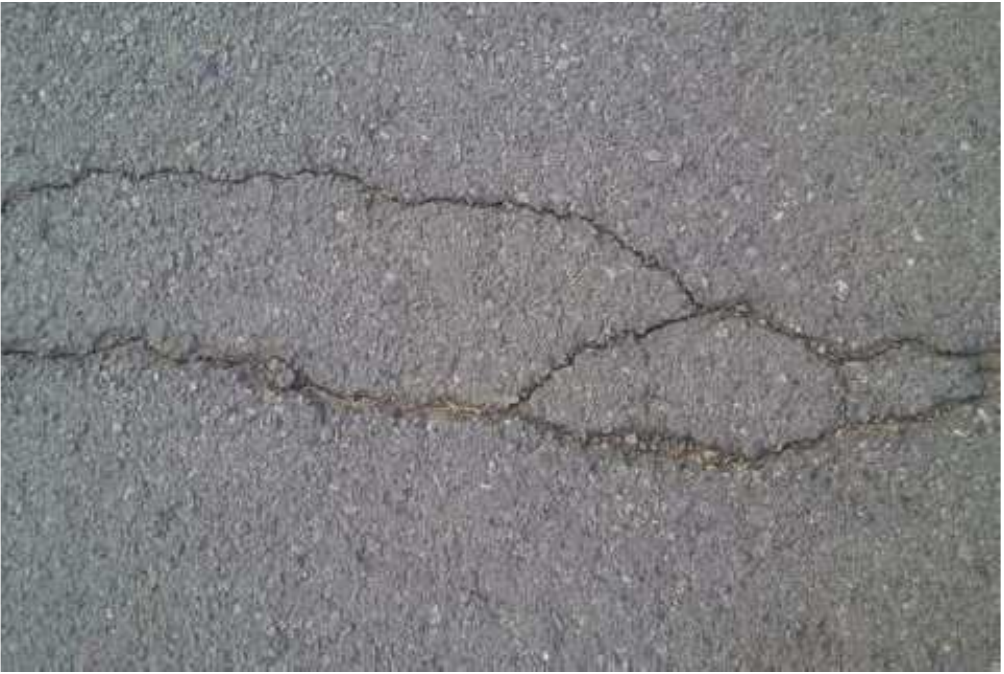}
\includegraphics[width=2.8cm]{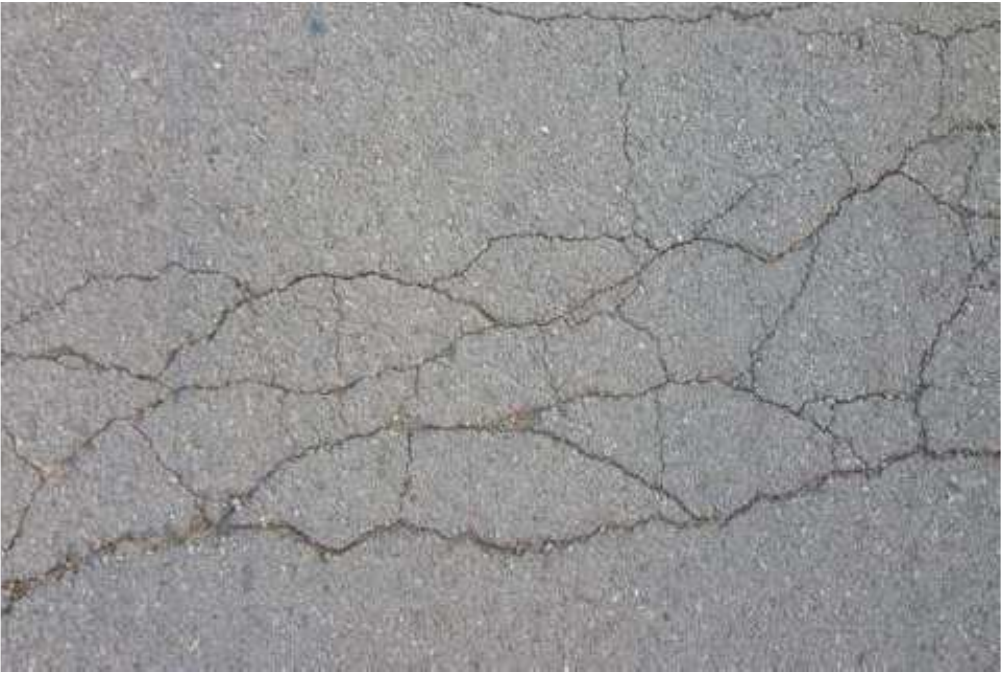}\\
~\\
\includegraphics[width=2.8cm]{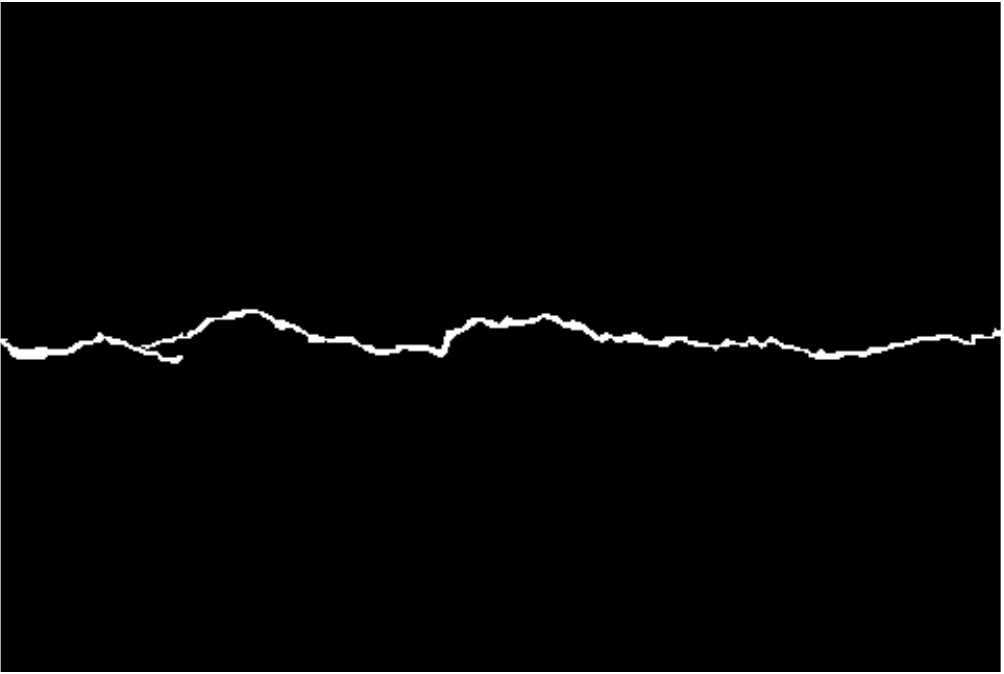}
\includegraphics[width=2.8cm]{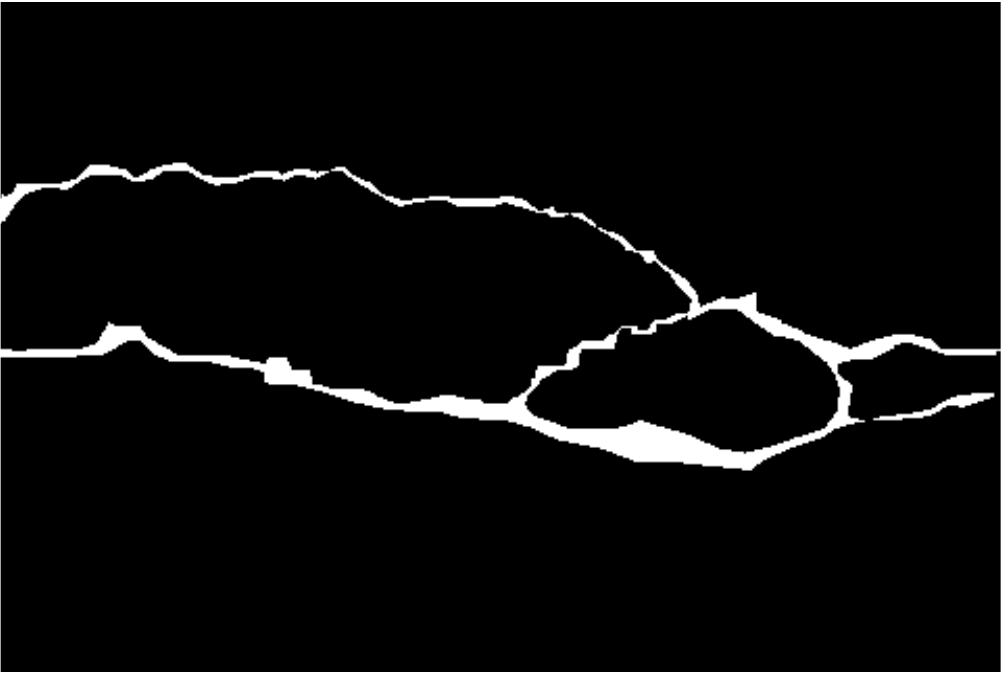}
\includegraphics[width=2.8cm]{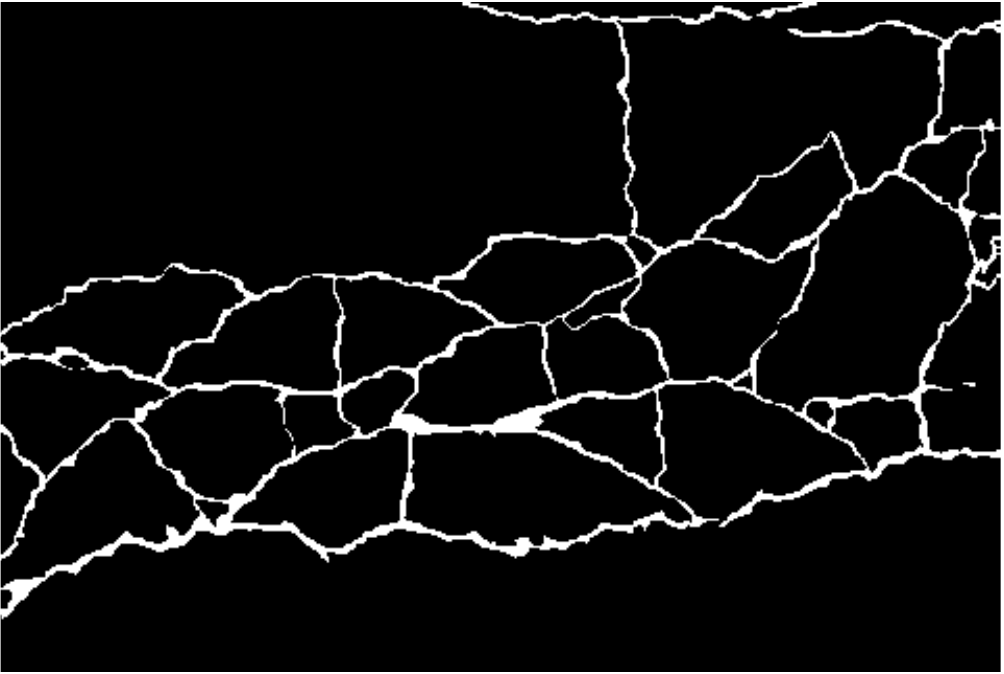}\\
~\\
\includegraphics[width=2.8cm]{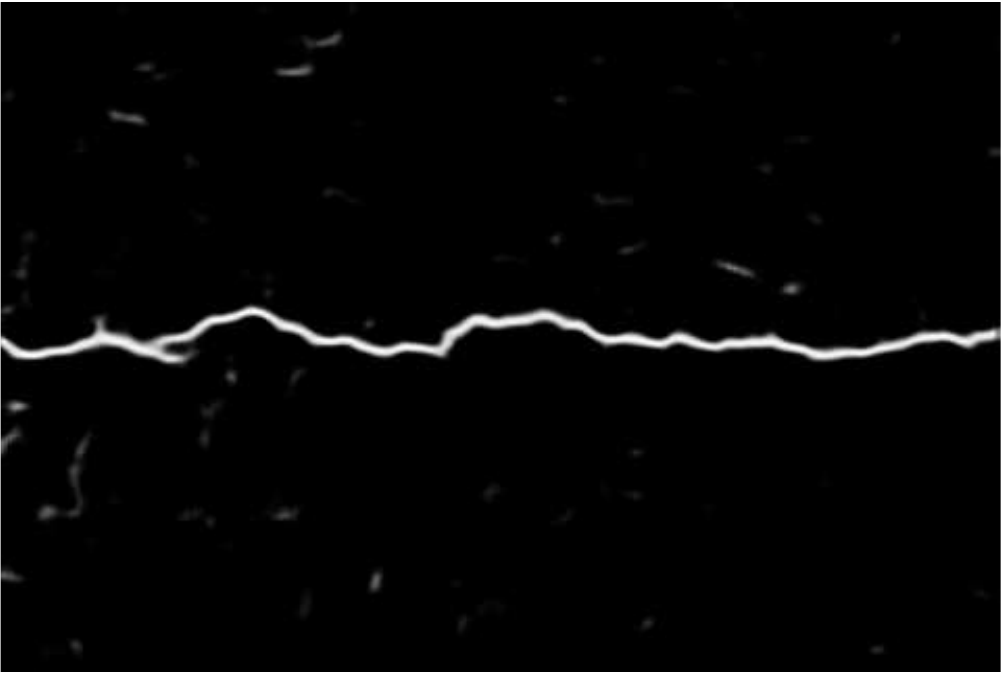}
\includegraphics[width=2.8cm]{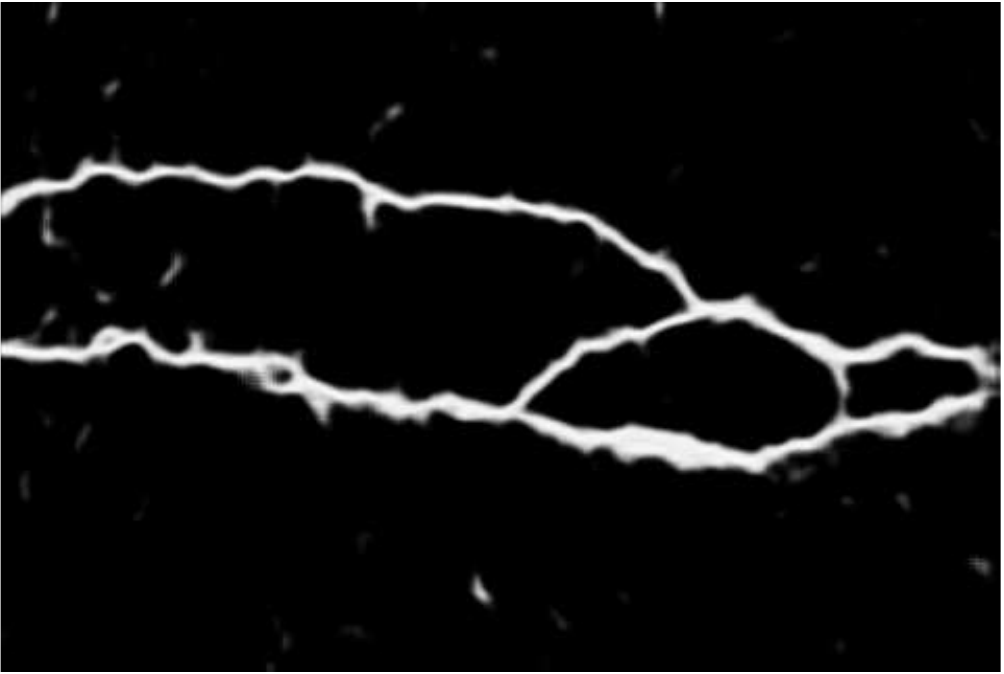}
\includegraphics[width=2.8cm]{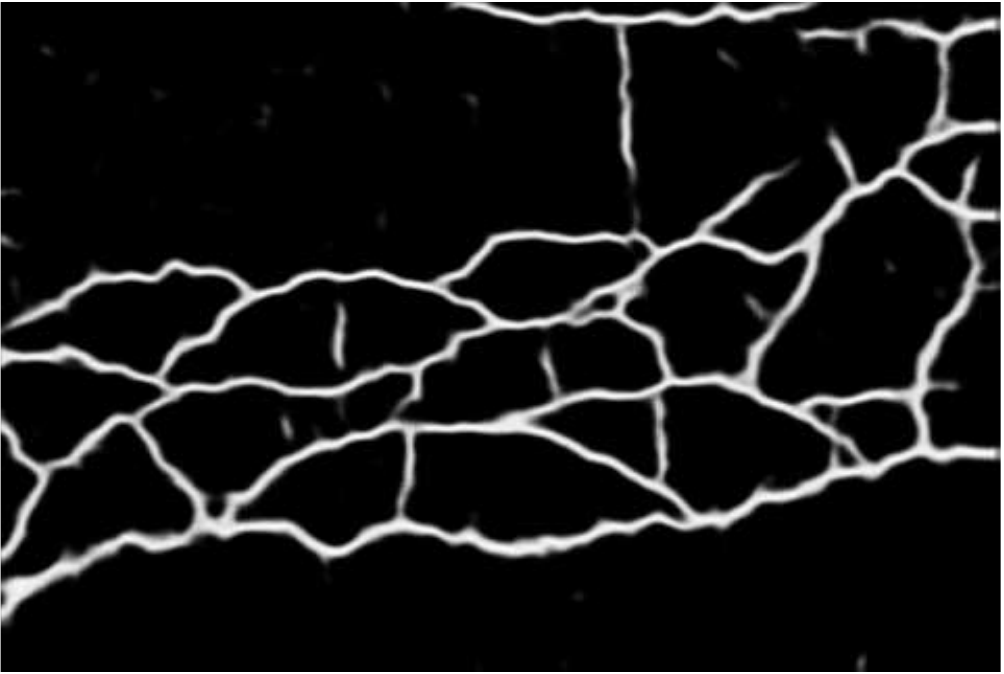}\\
~\\
\includegraphics[width=2.8cm]{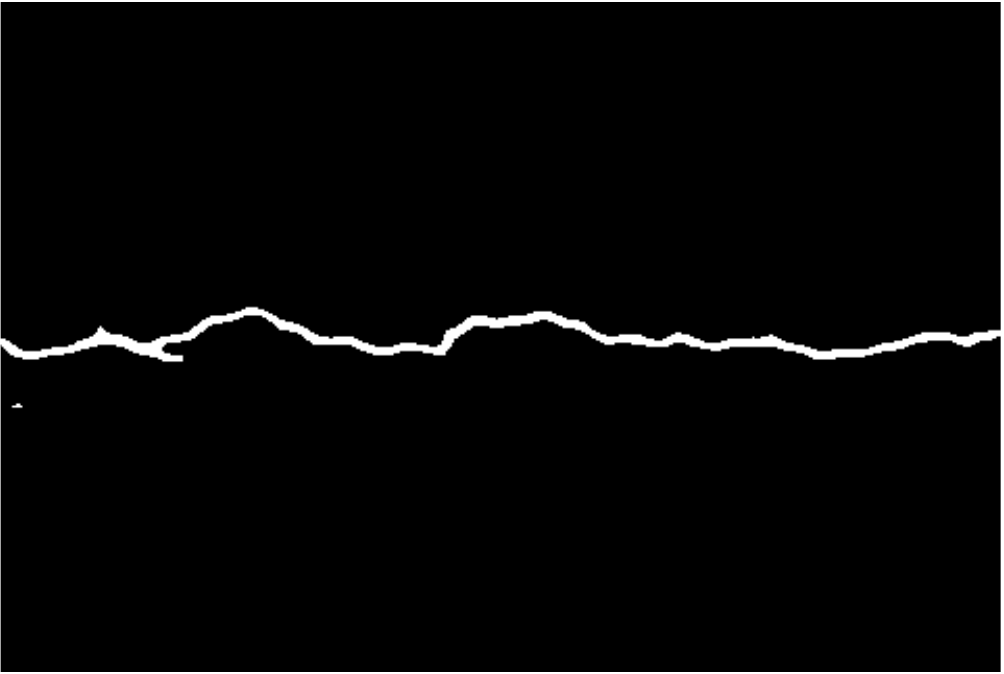}
\includegraphics[width=2.8cm]{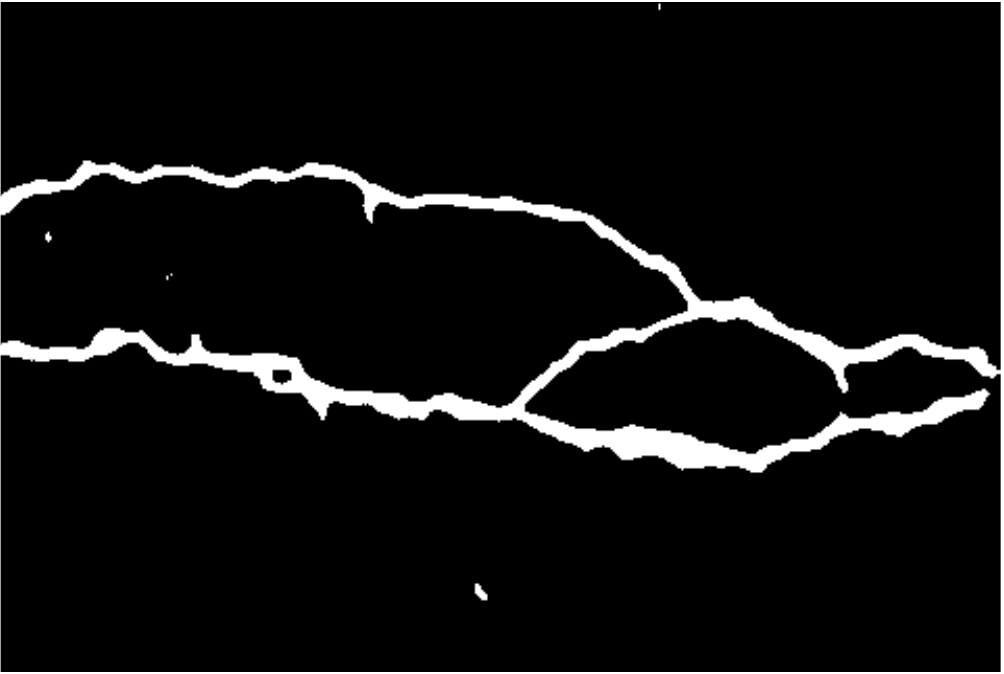}
\includegraphics[width=2.8cm]{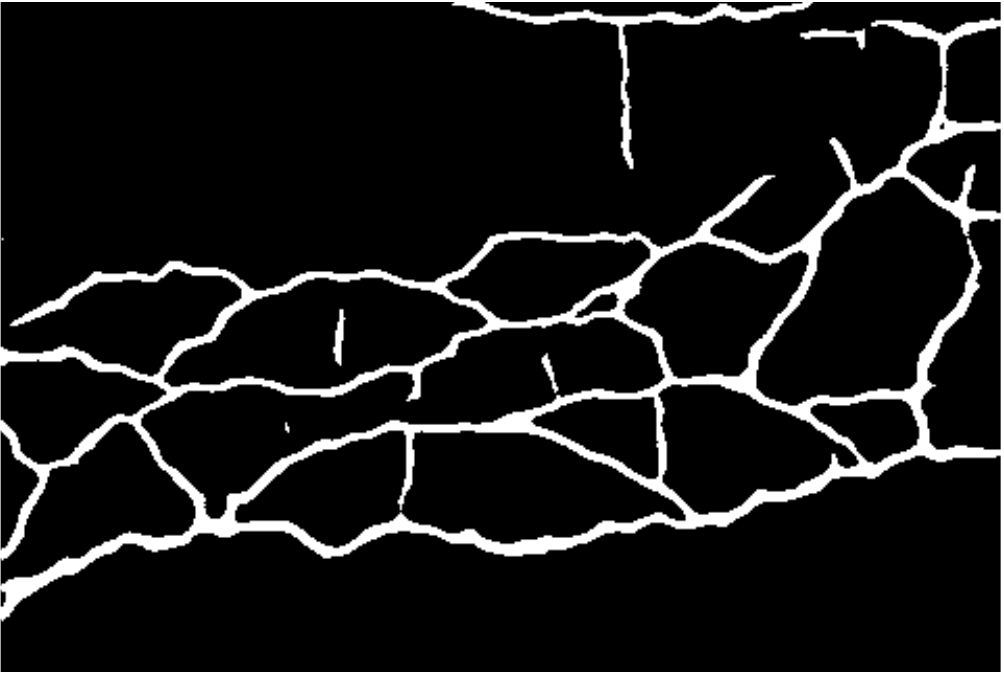}\\
\label{Fig. localization_binarization}
\caption{Part of crack detection results by the proposed method (from top to bottom: original image, ground truth, probability map, binary output).}
\end{figure}

Fig.~6 and Table III shows the comparison with some other algorithms. It can be observed that the two traditional methods are not suitable for crack detection as they are sensitive to noises. Since both CrackForest and the proposed method are supervised learning methods, we generate the same training and testing databases for them. Notice that crack width is overestimated in CrackForest, leading to the high recall but low precision as shown in Table III. Evaluation results shown in Table III indicate that the proposed method outperforms CrackForest.

\begin{figure*}[!t]
\centering
\includegraphics[width=2.8cm]{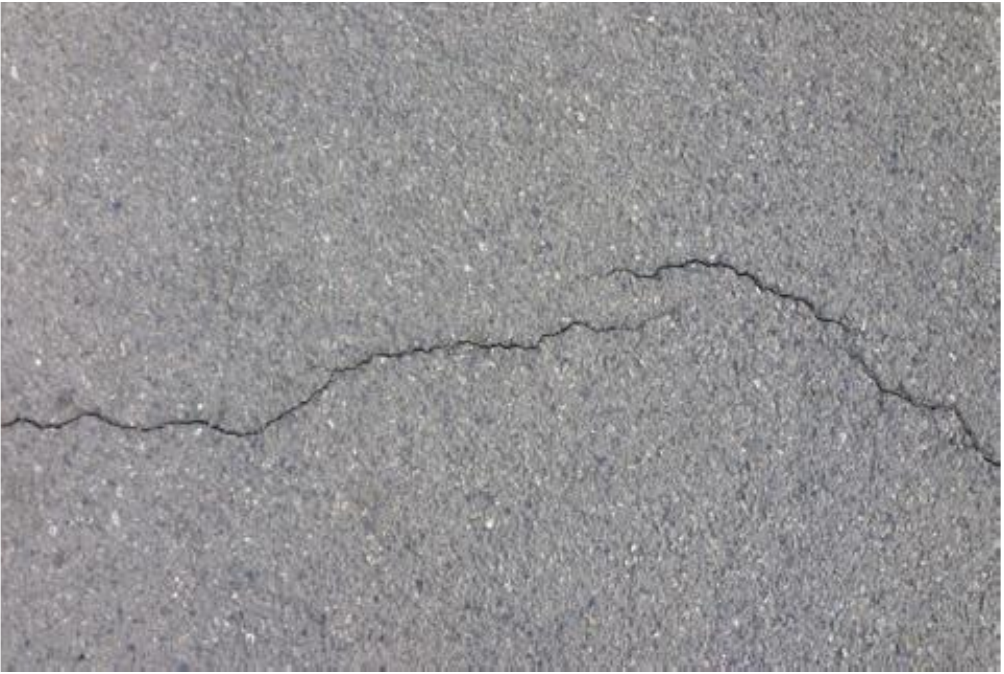}
\includegraphics[width=2.8cm]{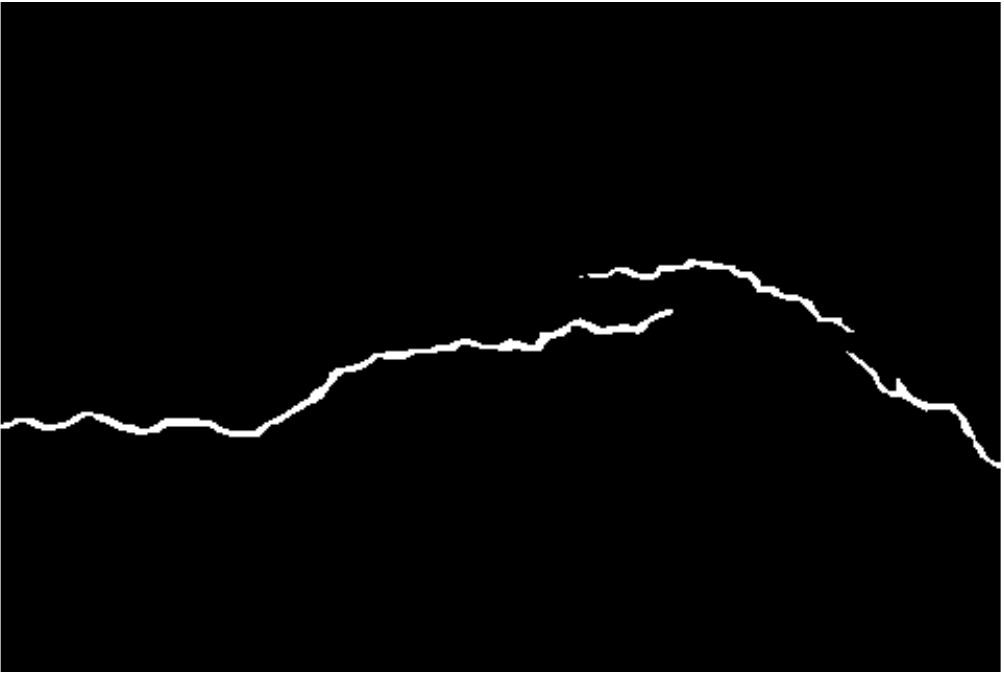}
\includegraphics[width=2.8cm]{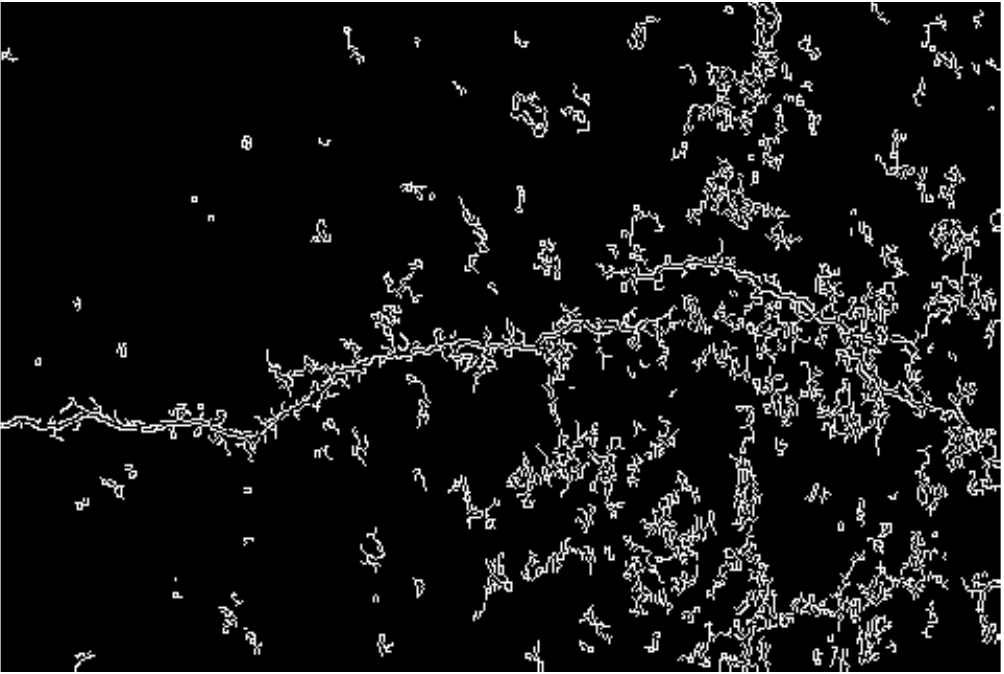}
\includegraphics[width=2.8cm]{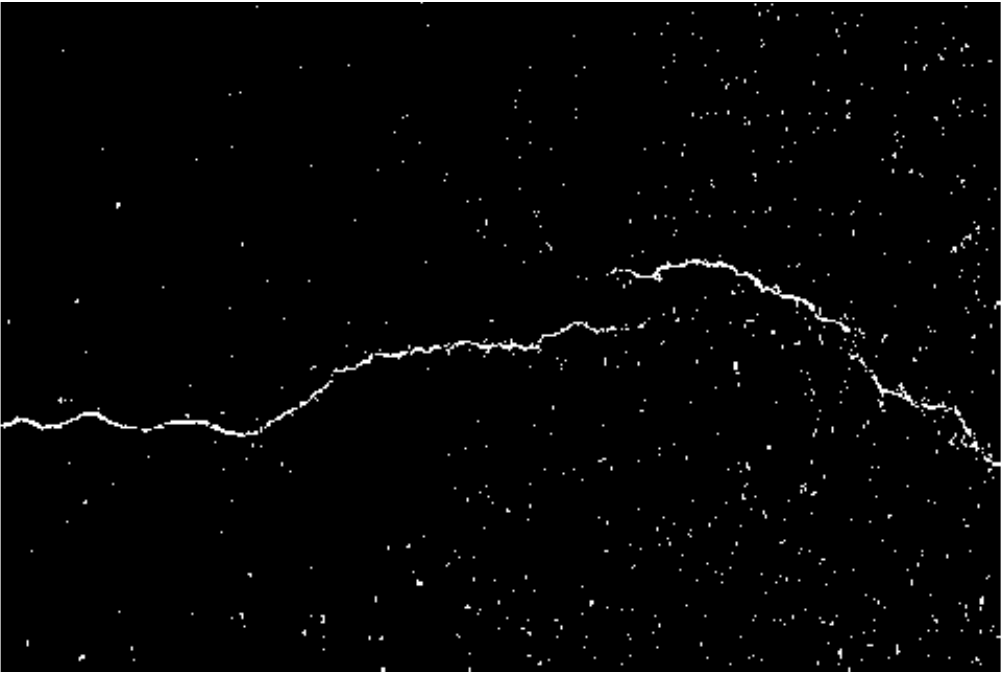}
\includegraphics[width=2.8cm]{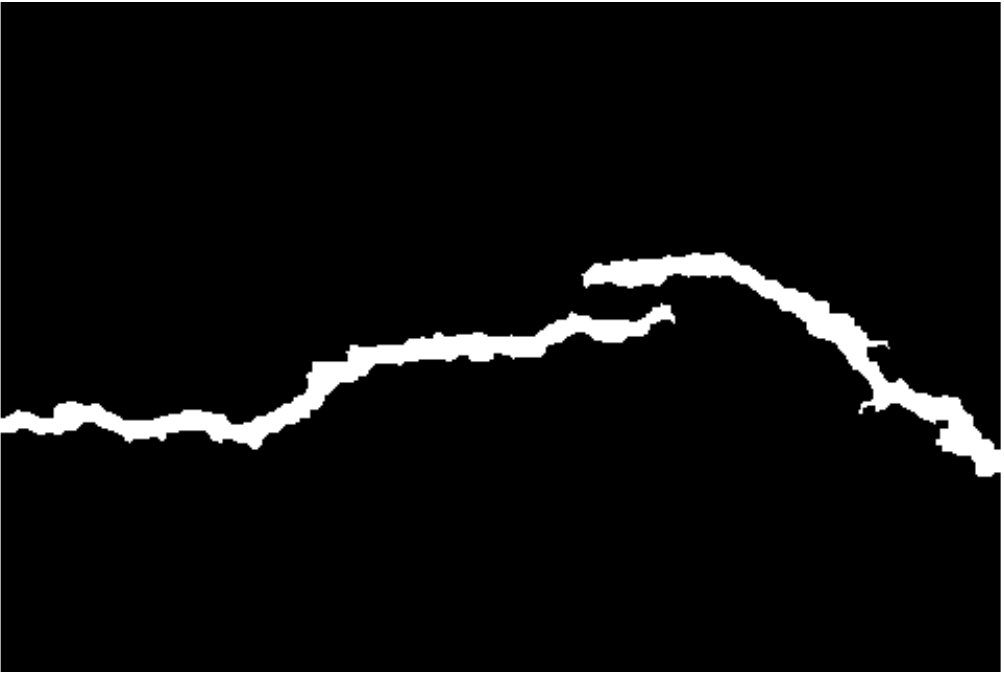}
\includegraphics[width=2.8cm]{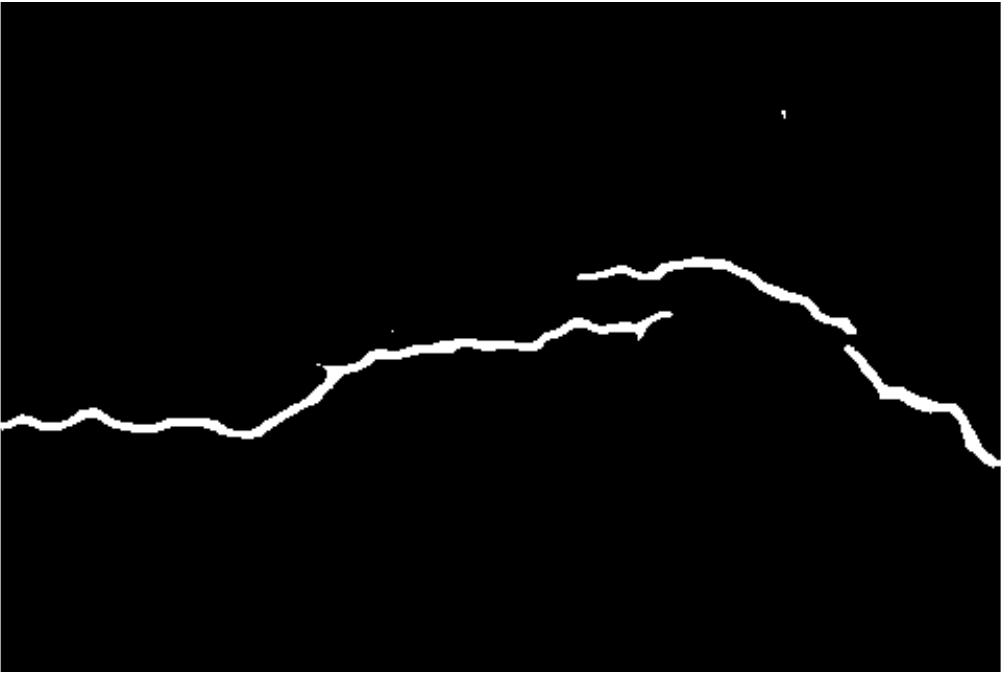}\\
~\\
\includegraphics[width=2.8cm]{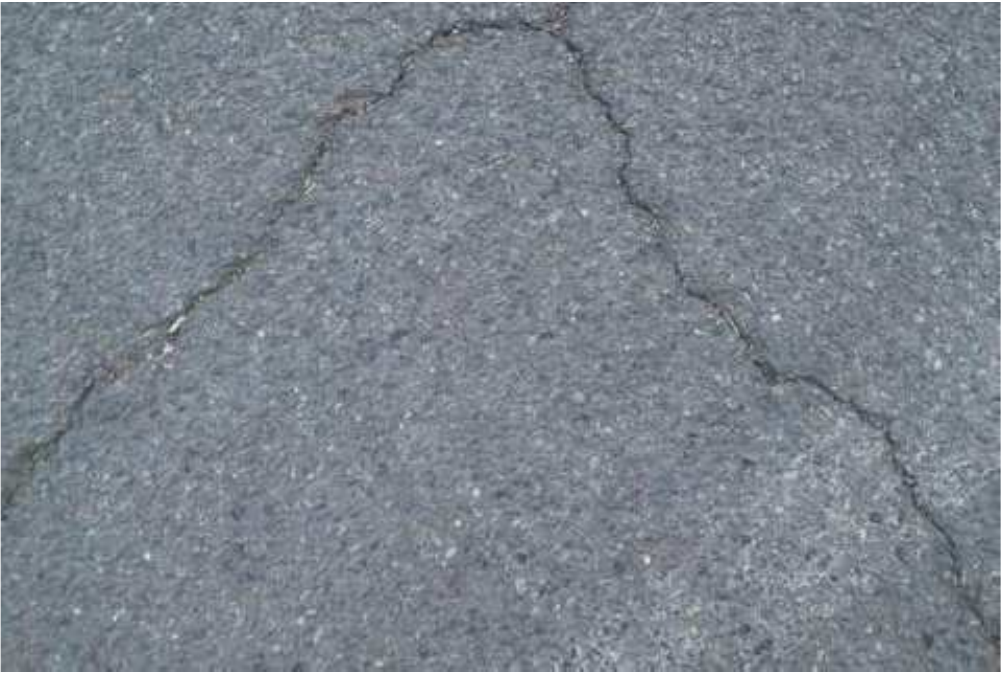}
\includegraphics[width=2.8cm]{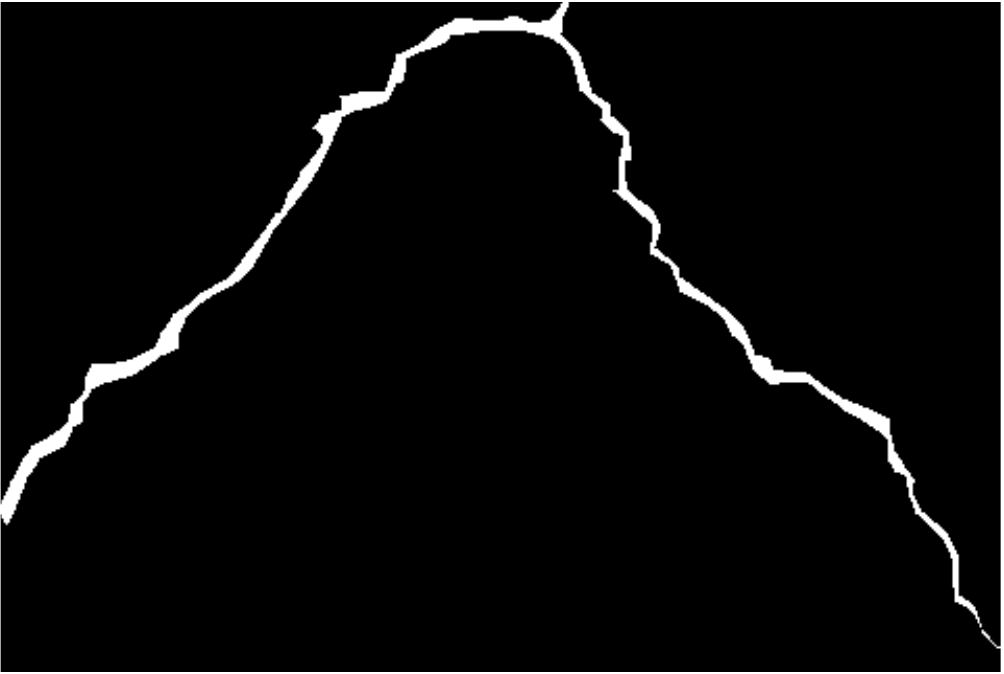}
\includegraphics[width=2.8cm]{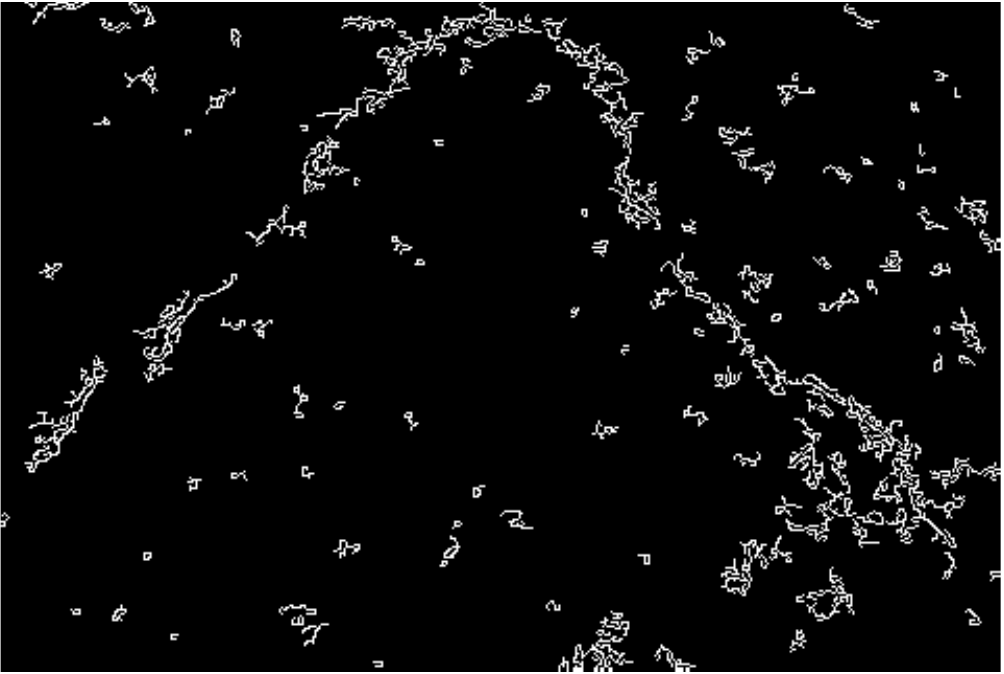}
\includegraphics[width=2.8cm]{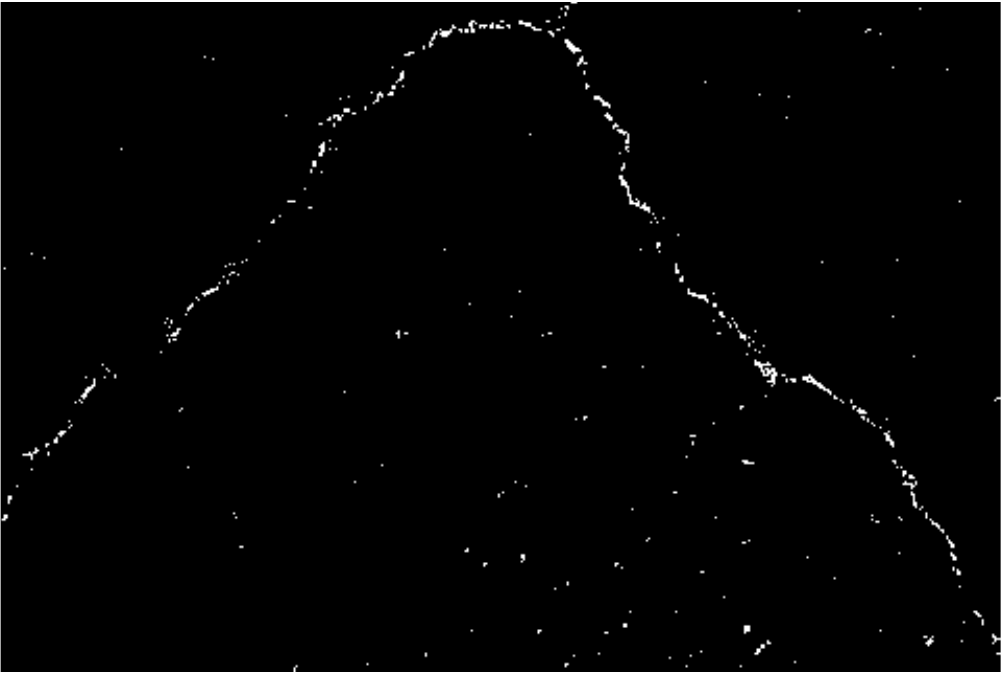}
\includegraphics[width=2.8cm]{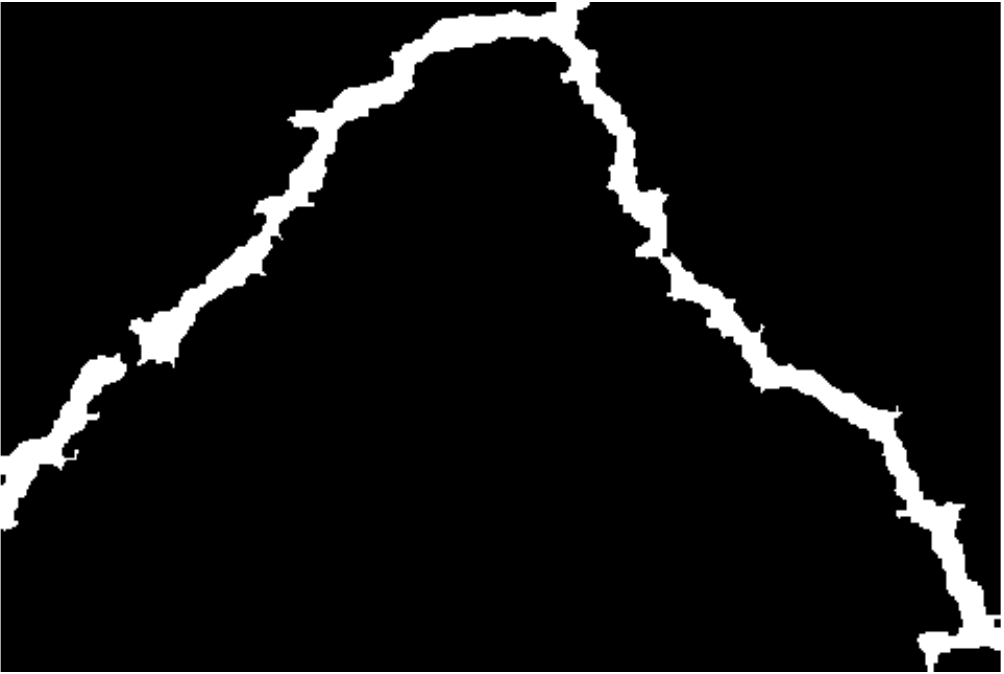}
\includegraphics[width=2.8cm]{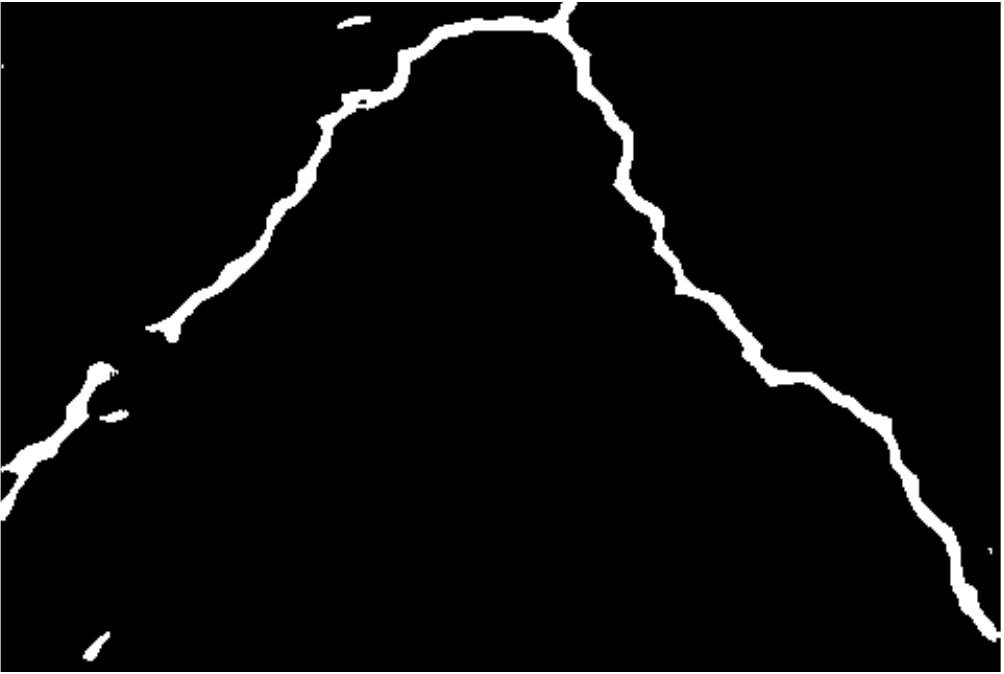}\\
~\\
\includegraphics[width=2.8cm]{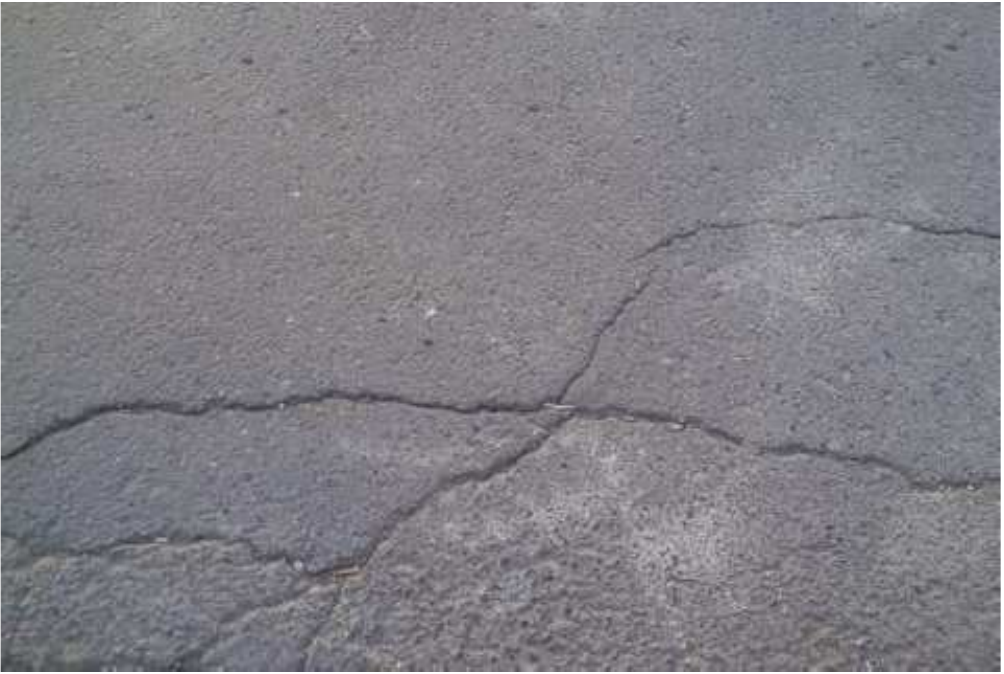}
\includegraphics[width=2.8cm]{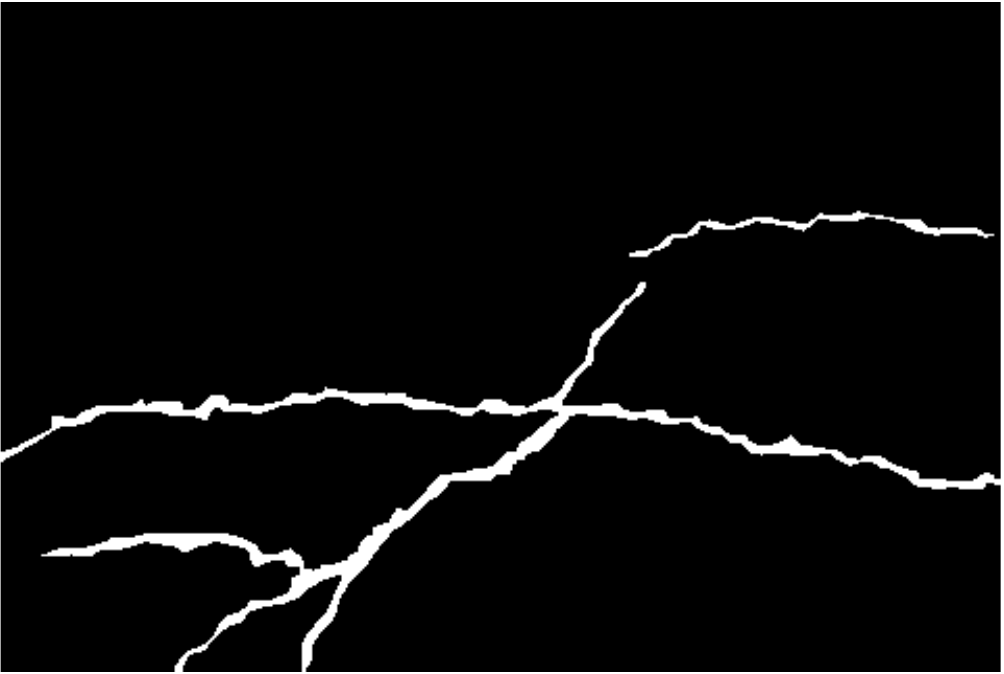}
\includegraphics[width=2.8cm]{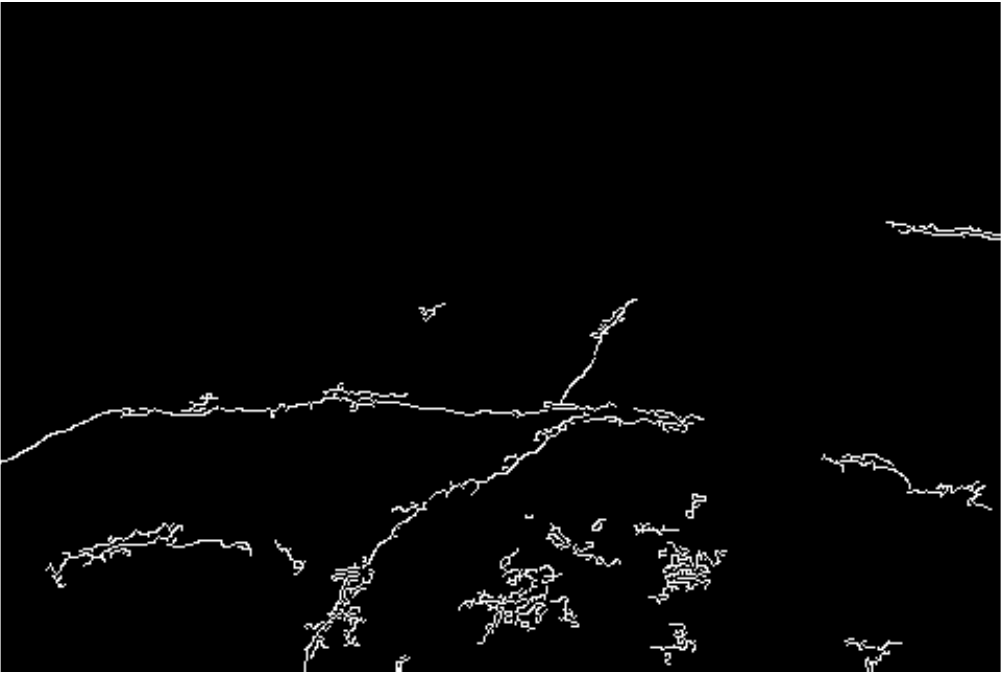}
\includegraphics[width=2.8cm]{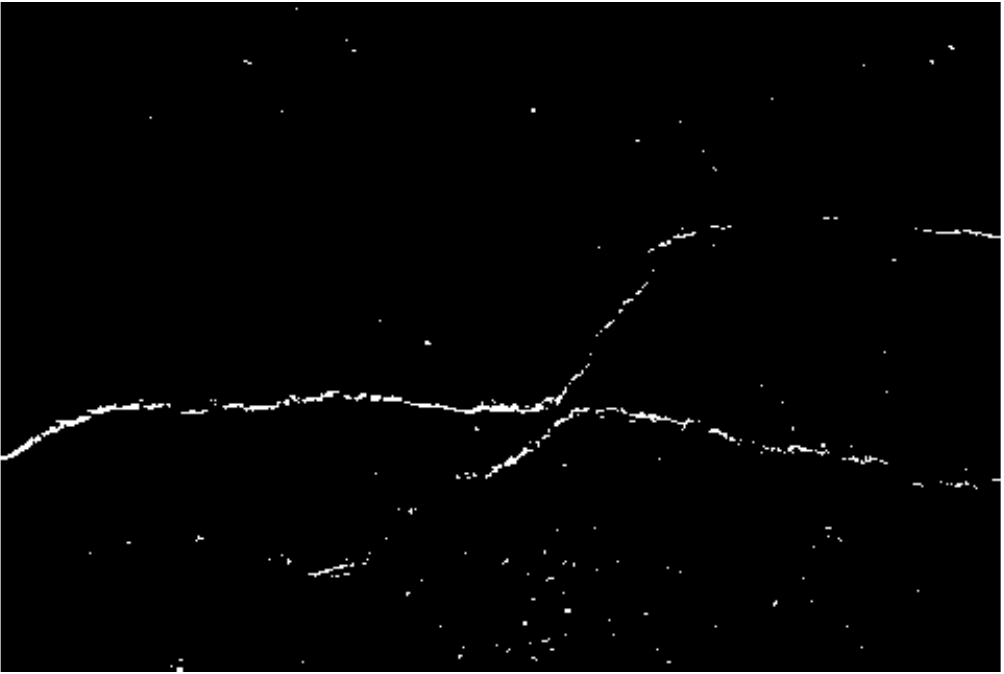}
\includegraphics[width=2.8cm]{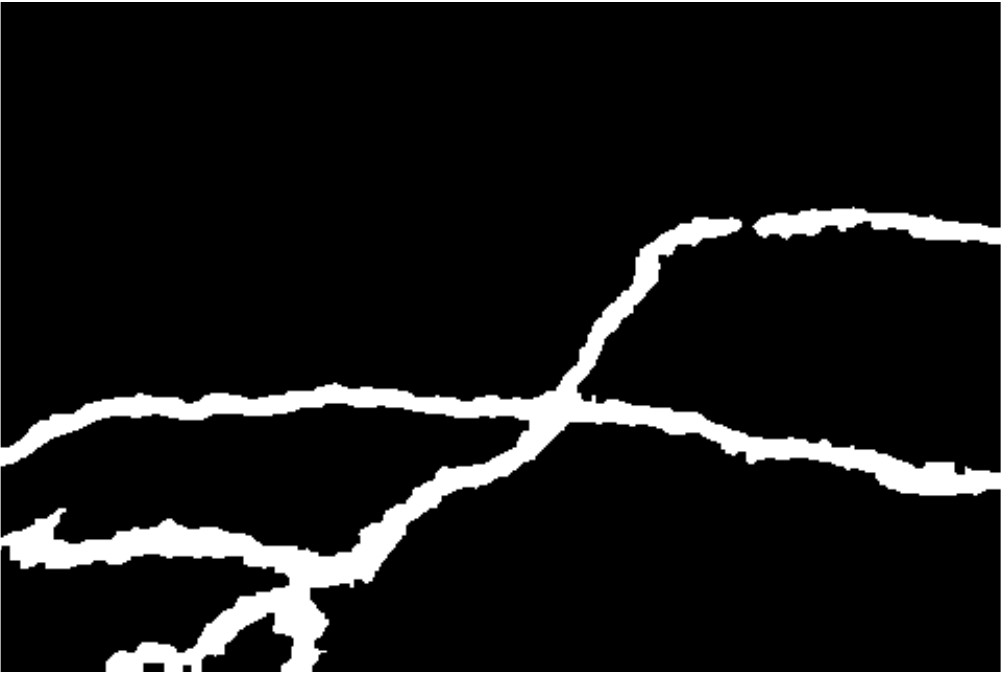}
\includegraphics[width=2.8cm]{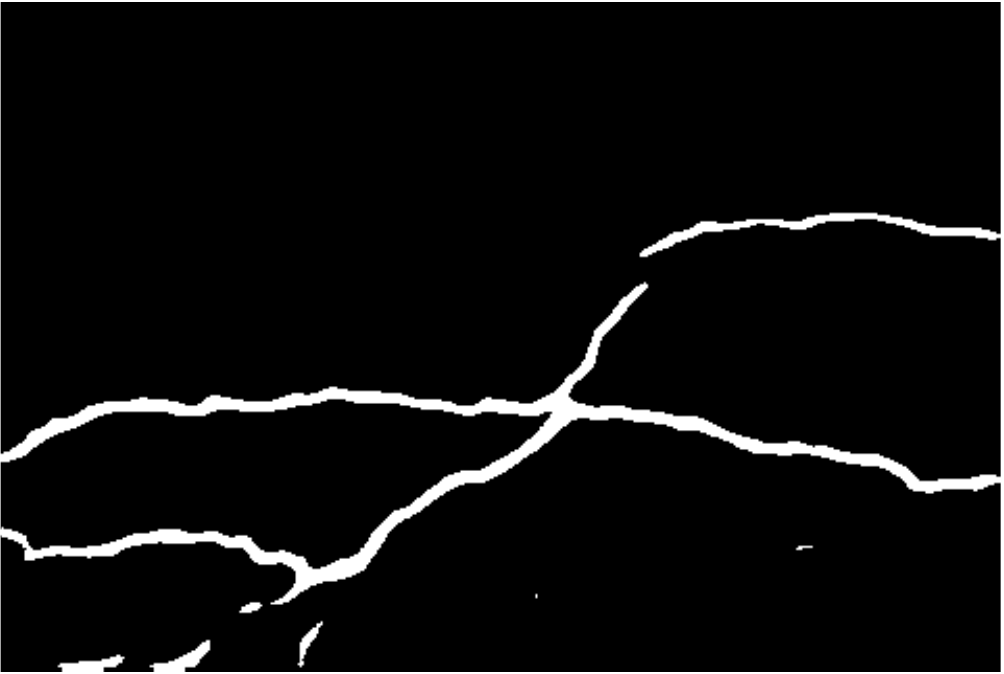}
\caption{Results comparing on CFD (from left to right: original image, ground truth, Canny, local thresholding, CrackForest, the proposed method).}
\label{Fig. CFD}
\end{figure*}

\begin{table}[!t]
\renewcommand{\arraystretch}{1.3}
\caption{Crack Detection Results Evaluation on CFD}
\label{Table III}
\centering
\begin{tabular}{c| c c c}
\hline
\space & $Pr$ & $Re$ & $F1$\\
\hline
Canny & 0.4377 & 0.7307 & 0.4570\\
Local thresholding & 0.7727 & 0.8274 & 0.7418\\
CrackForest & 0.7466 & {\bfseries 0.9514} & 0.8318 \\
The proposed method & {\bfseries 0.9119} & 0.9481 & {\bfseries 0.9244}\\
\hline
\end{tabular}
\end{table}

\subsection{Results on AigleRN}

Fig.~7 and Table IV present the results of different algorithms on AigleRN. It is again observed that the two traditional methods are sensitive to noise and fail to find the continuous cracks. FFA can find some cracks in a local scale, but can hardly find all the continuous cracks. MPS is more effective to find the continuous cracks on a global view. Compared to MPS, our proposed method can find the cracks with better precision and higher recall. From image results, it is clear that the proposed method outperforms the others.

\begin{figure*}[!t]
\centering
\includegraphics[width=2.4cm]{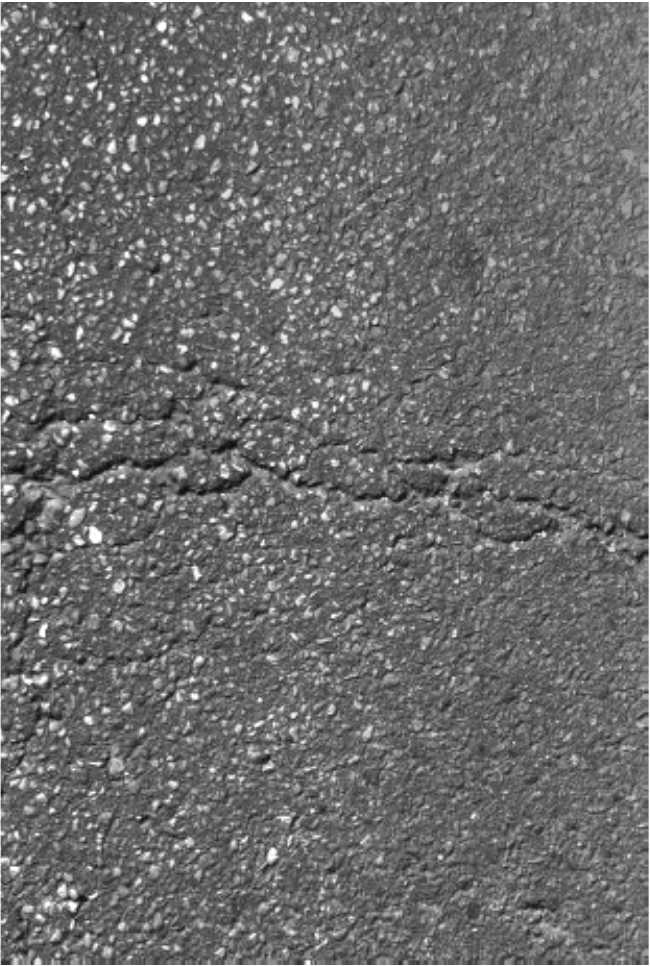}
\includegraphics[width=2.4cm]{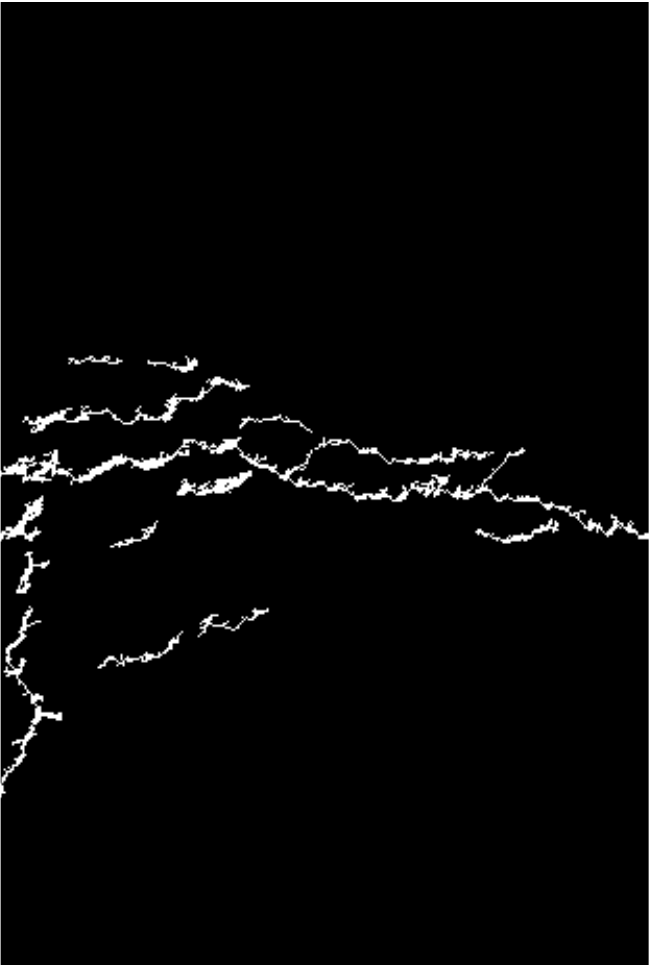}
\includegraphics[width=2.4cm]{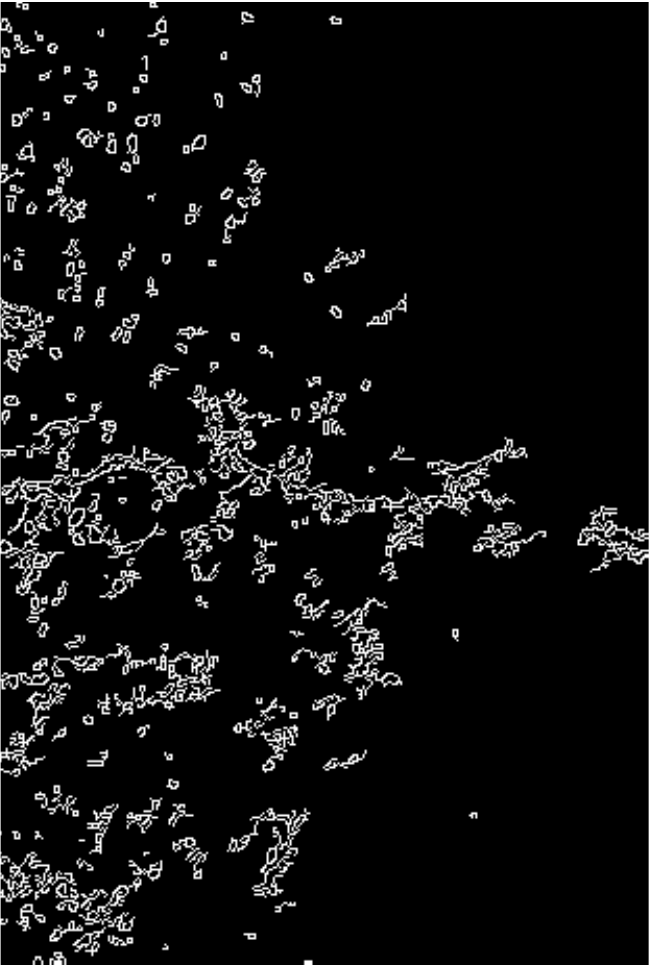}
\includegraphics[width=2.4cm]{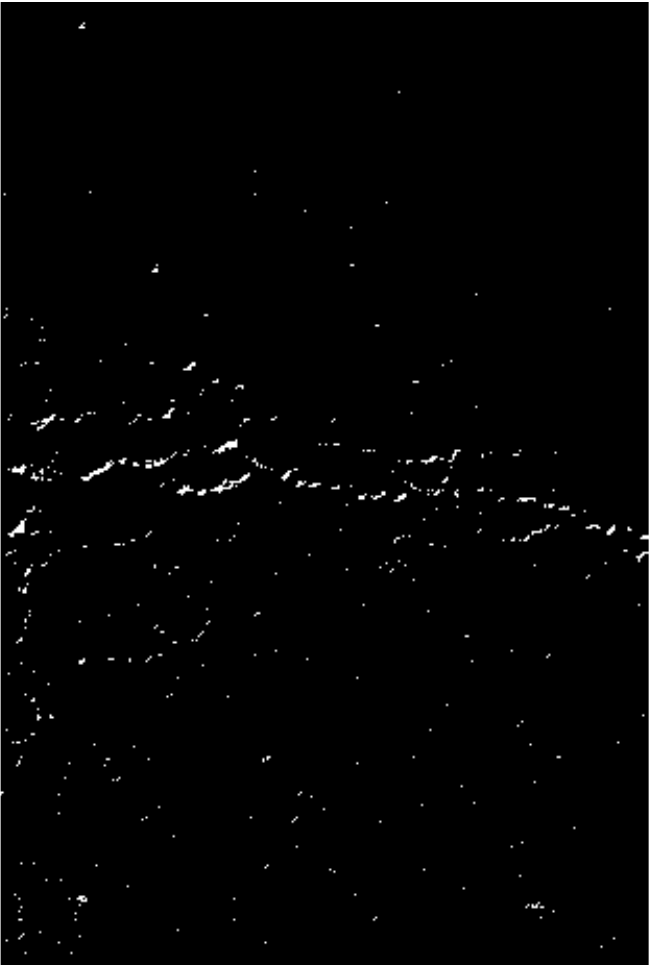}
\includegraphics[width=2.4cm]{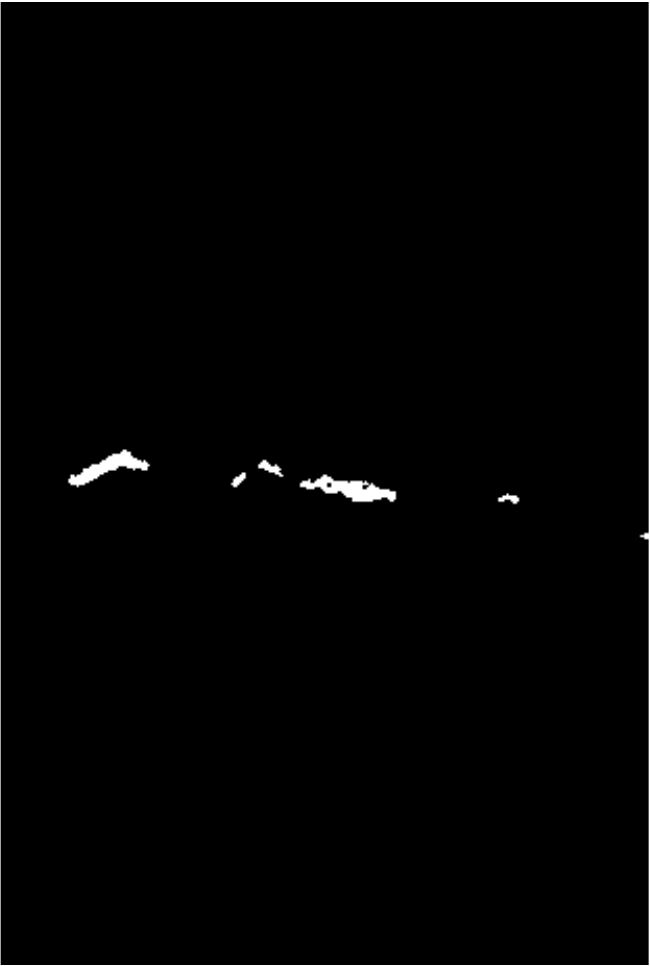}
\includegraphics[width=2.4cm]{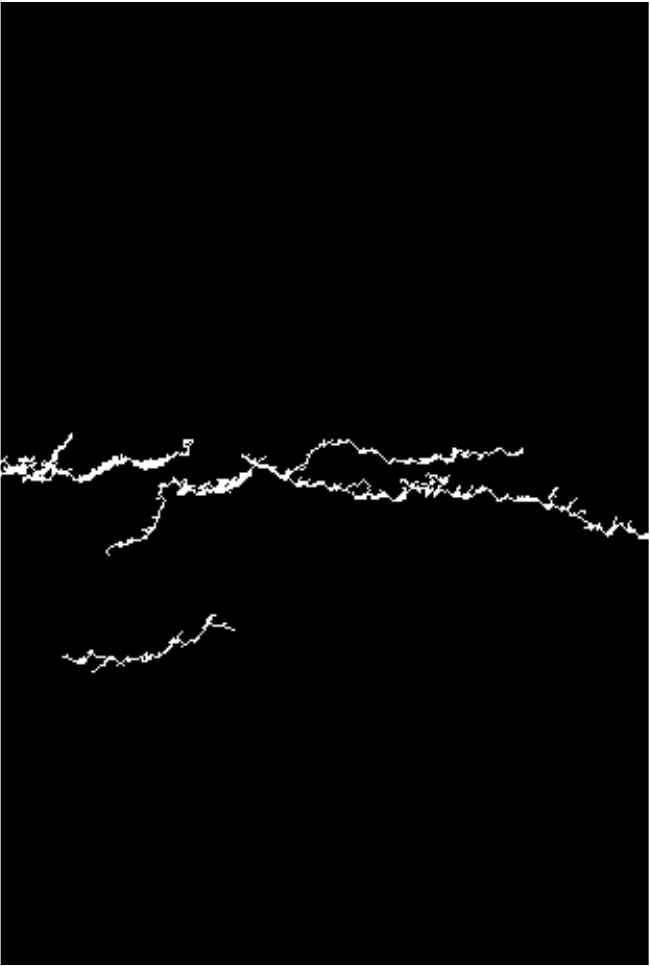}
\includegraphics[width=2.4cm]{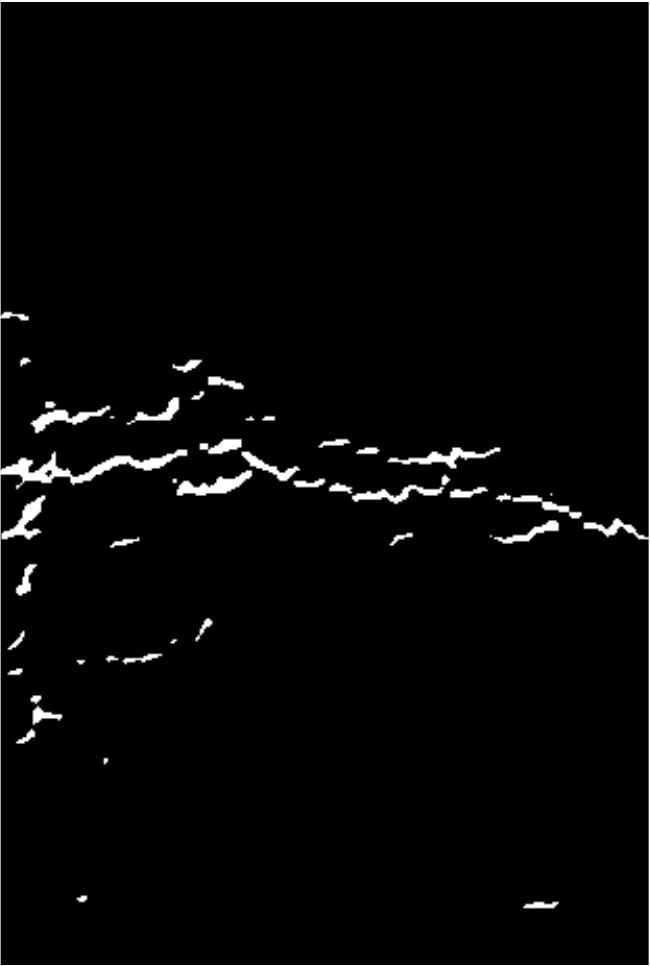}\\
~\\
\includegraphics[width=2.4cm]{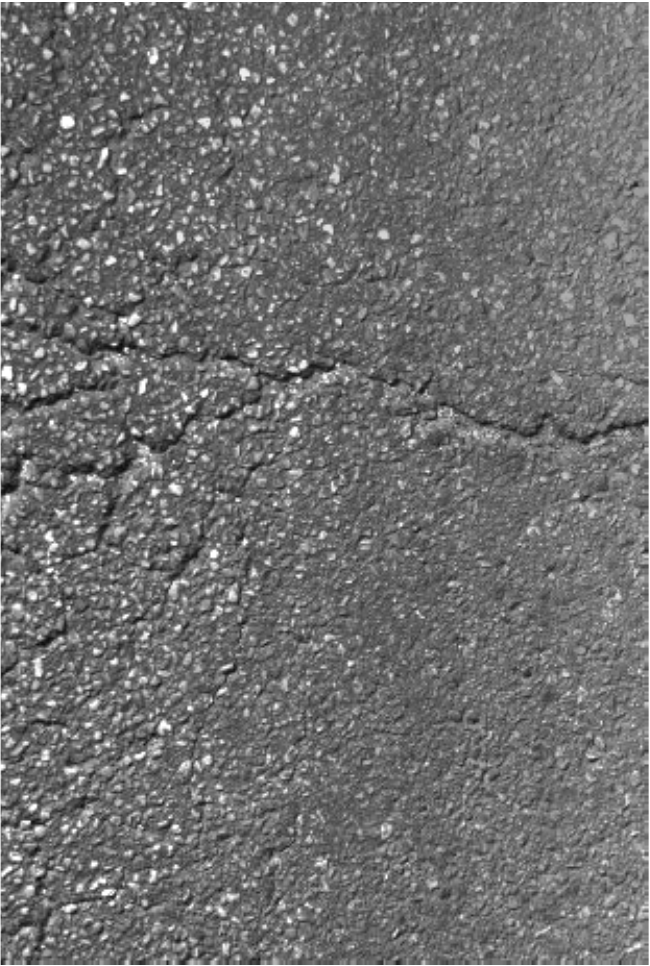}
\includegraphics[width=2.4cm]{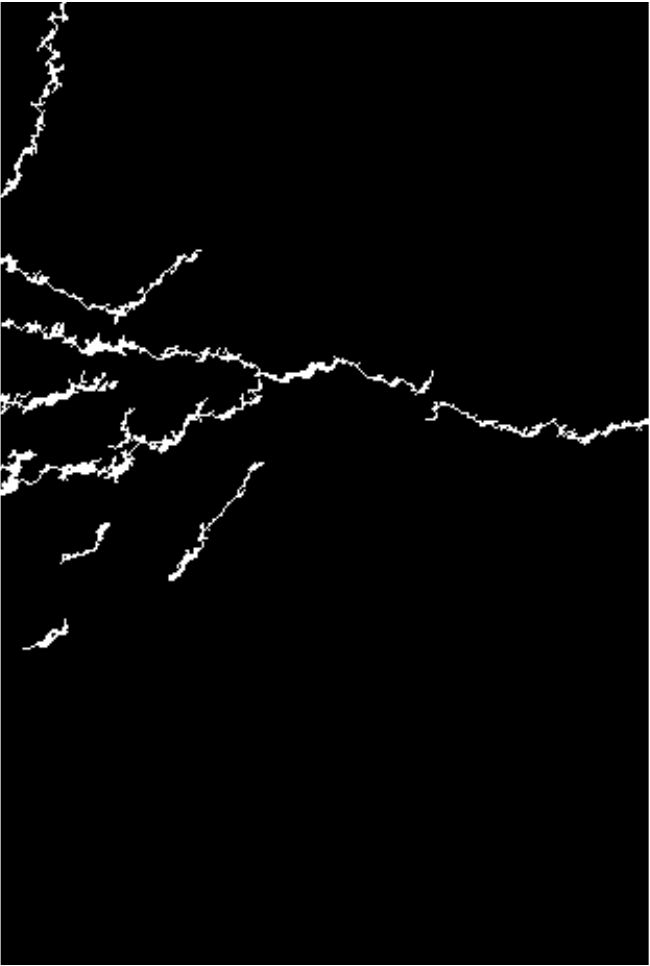}
\includegraphics[width=2.4cm]{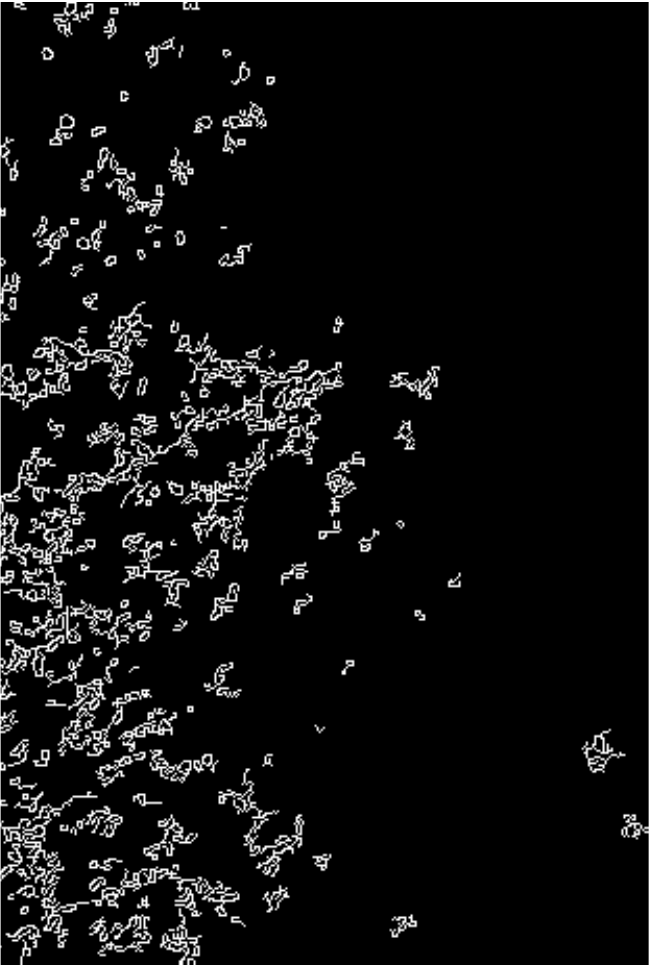}
\includegraphics[width=2.4cm]{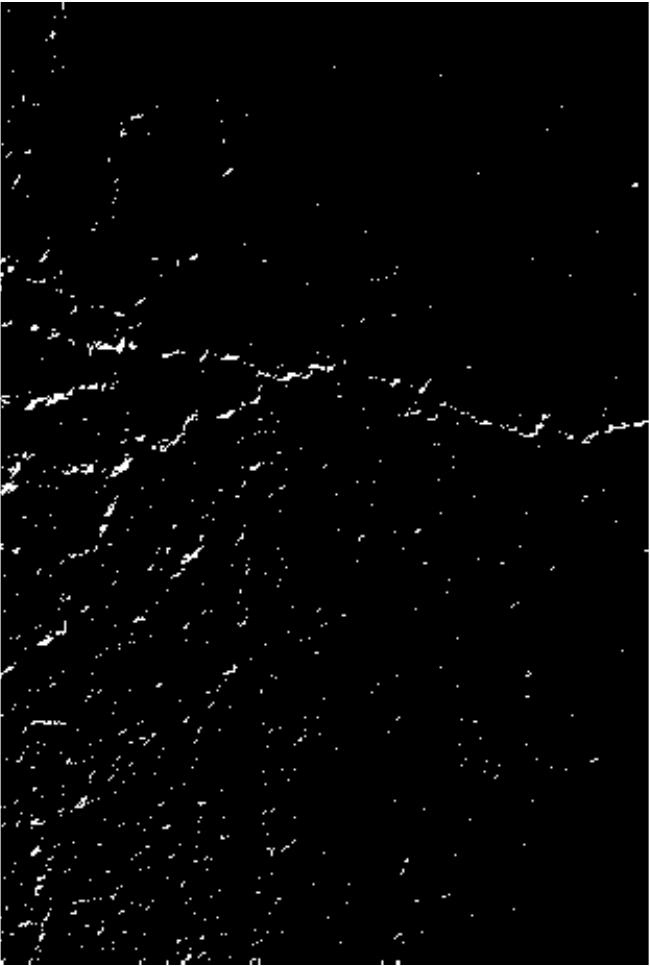}
\includegraphics[width=2.4cm]{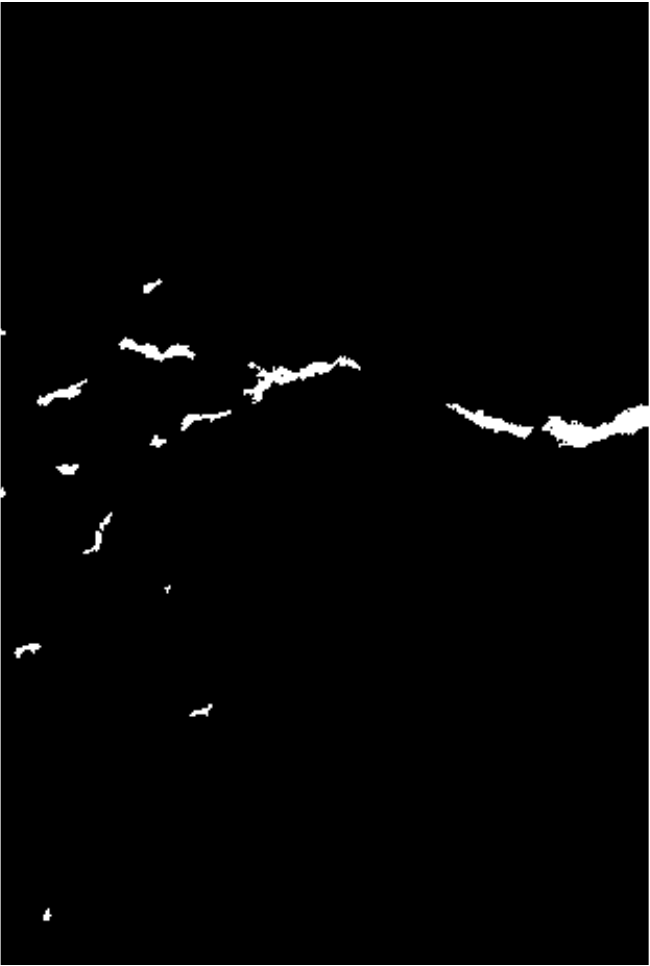}
\includegraphics[width=2.4cm]{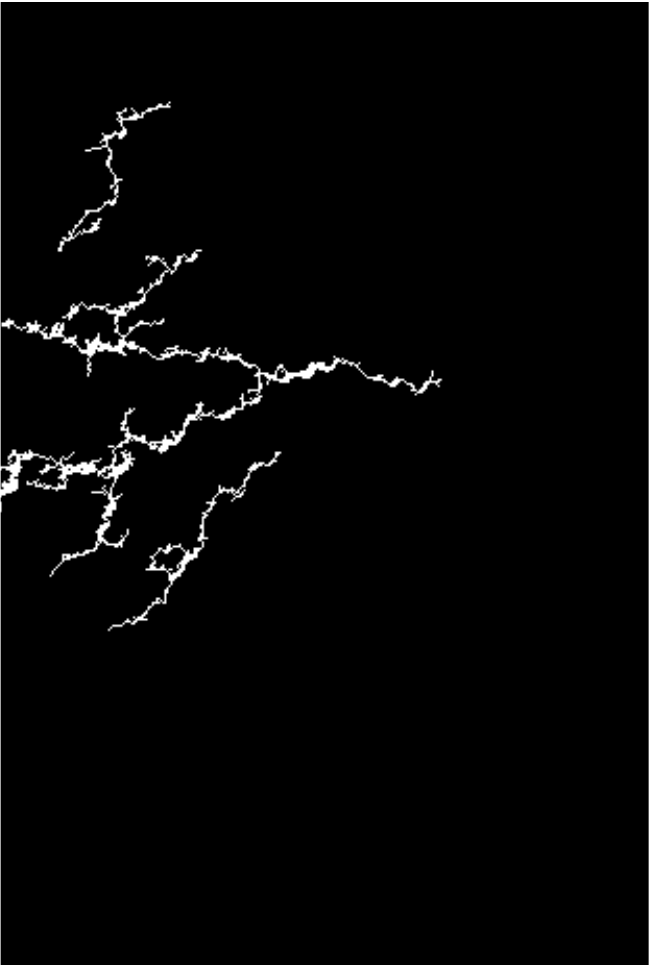}
\includegraphics[width=2.4cm]{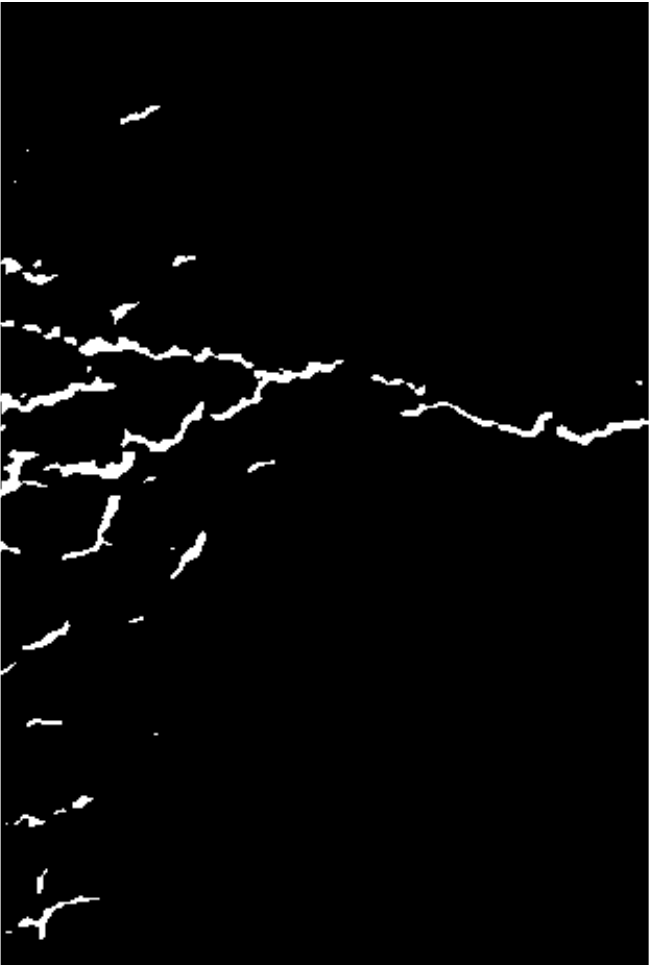}\\
~\\
\includegraphics[width=2.4cm]{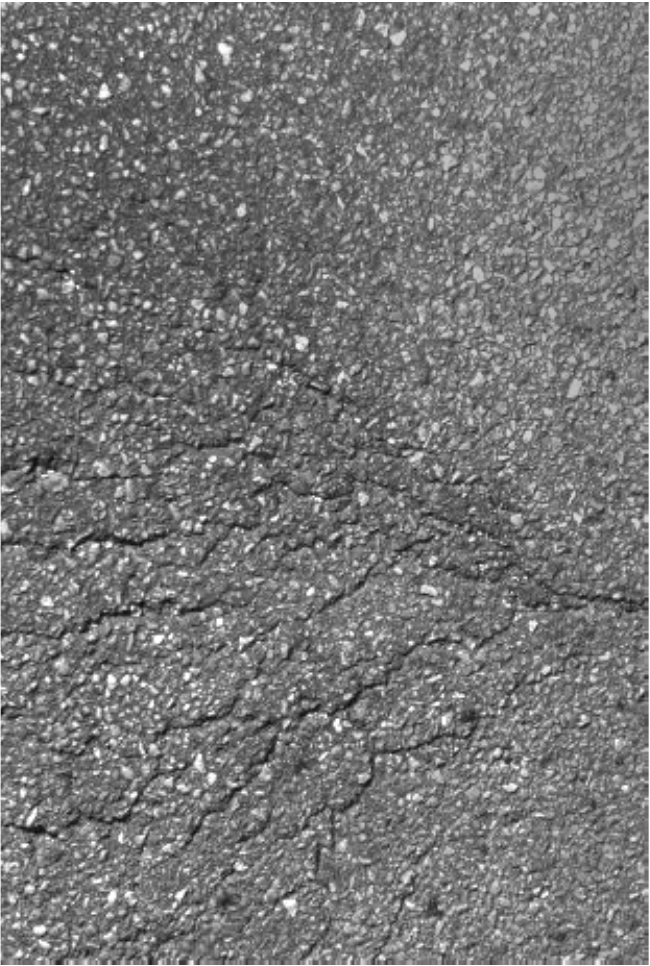}
\includegraphics[width=2.4cm]{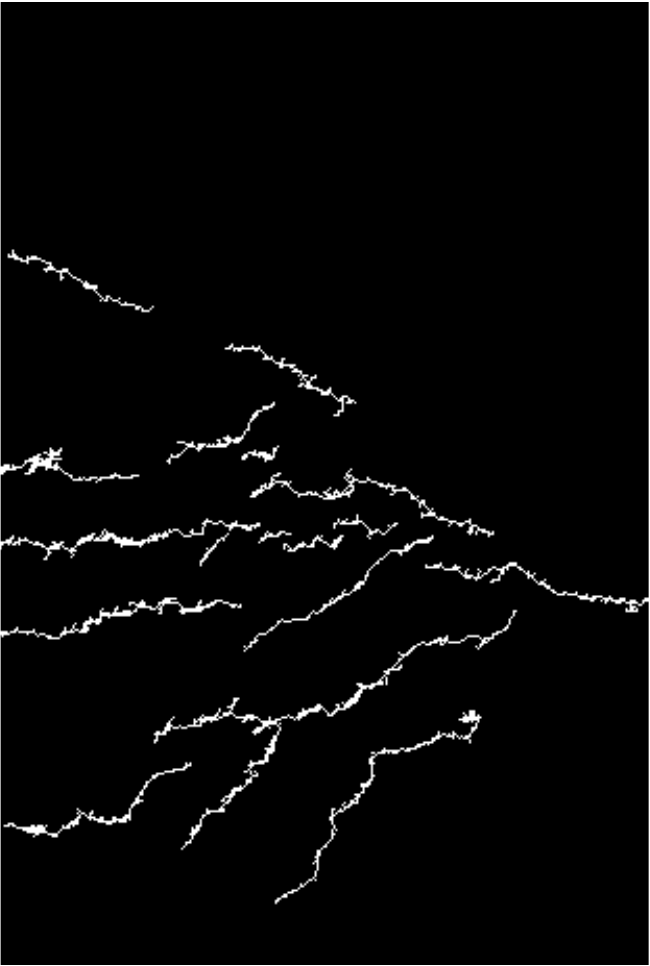}
\includegraphics[width=2.4cm]{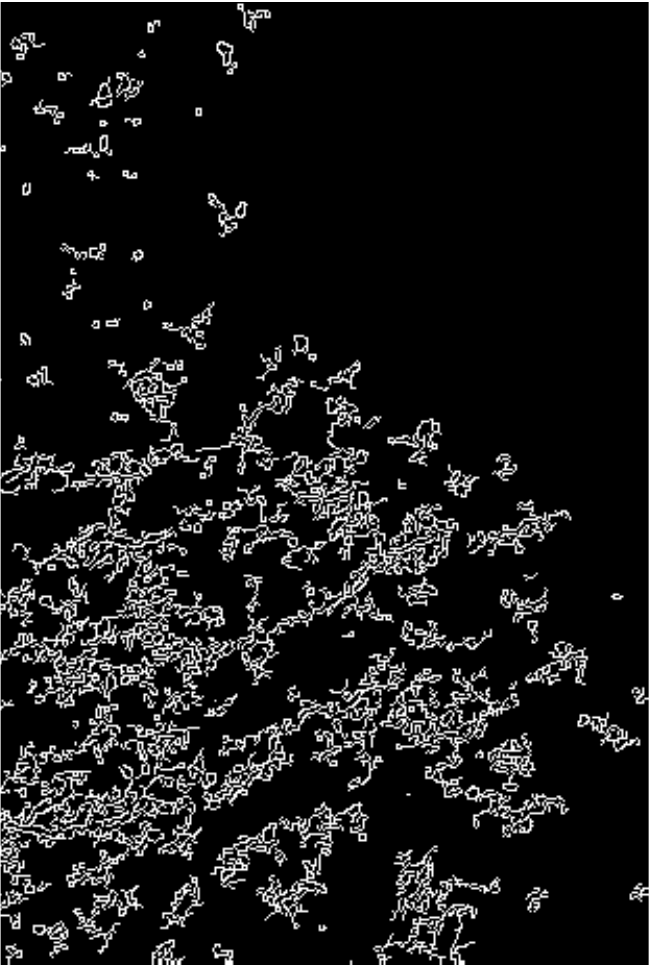}
\includegraphics[width=2.4cm]{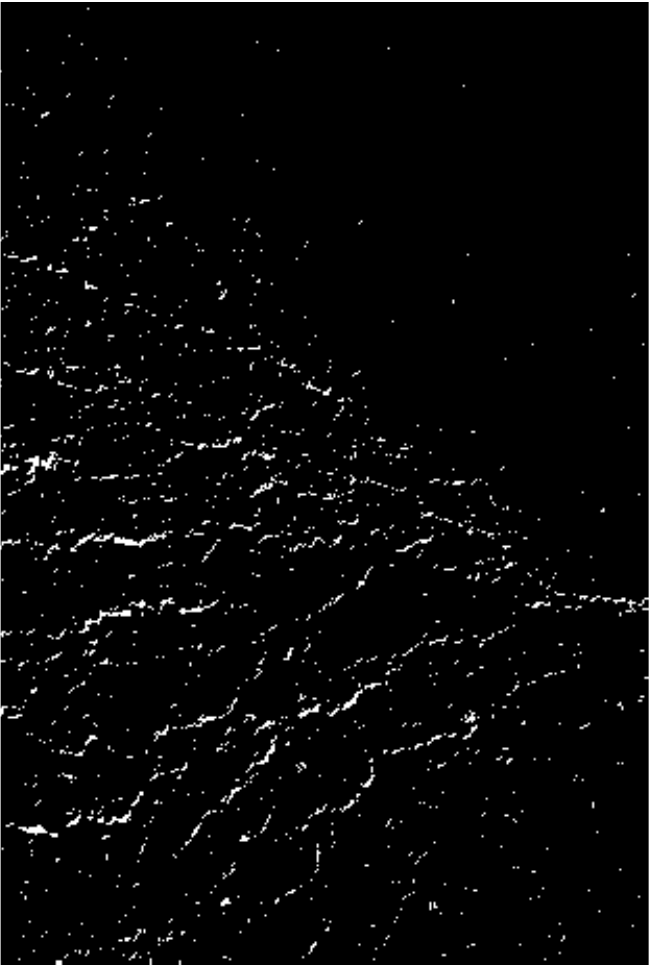}
\includegraphics[width=2.4cm]{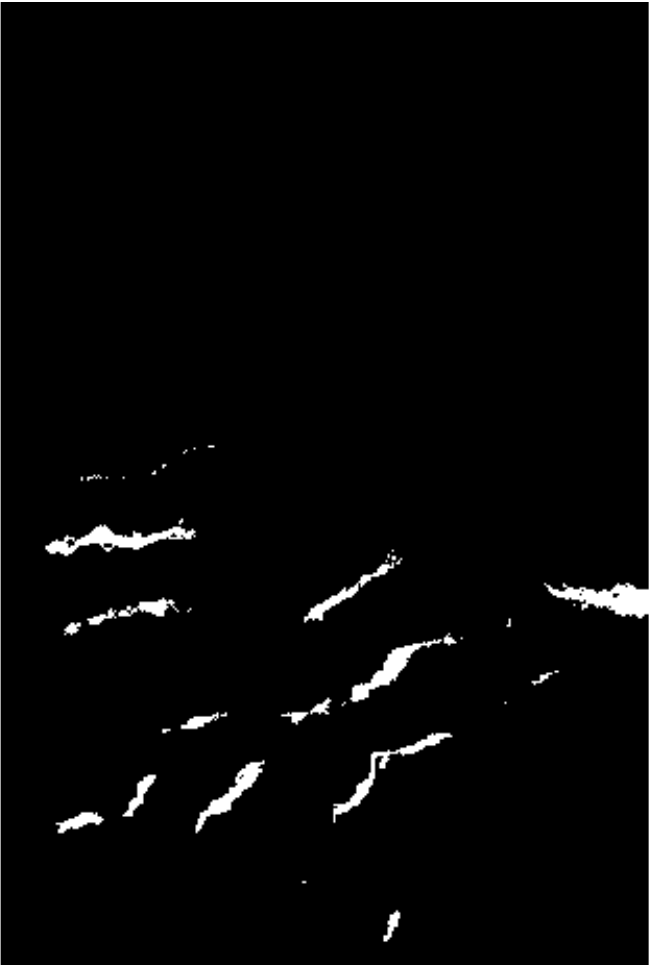}
\includegraphics[width=2.4cm]{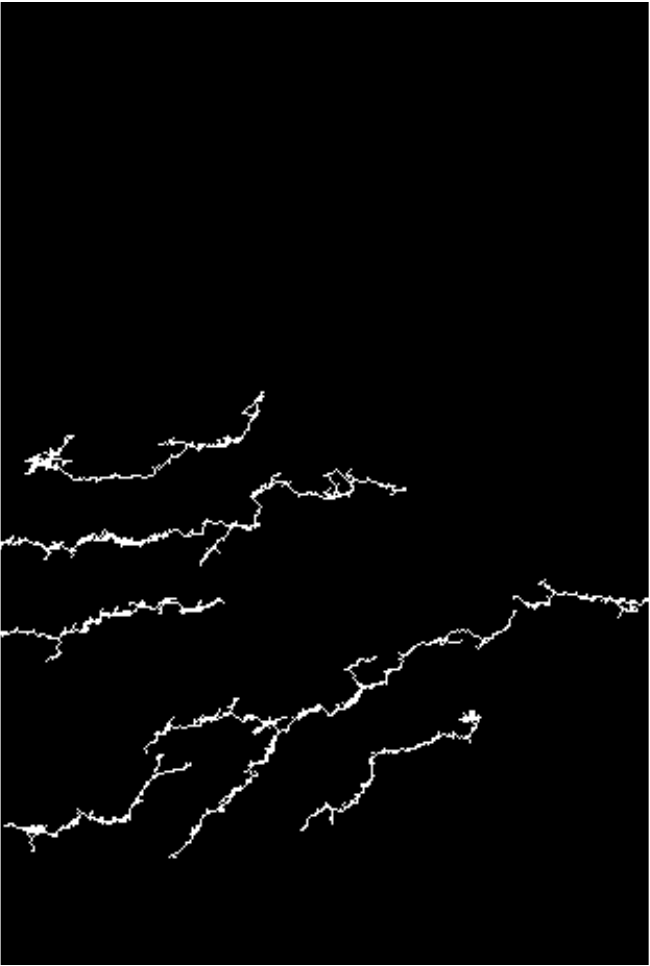}
\includegraphics[width=2.4cm]{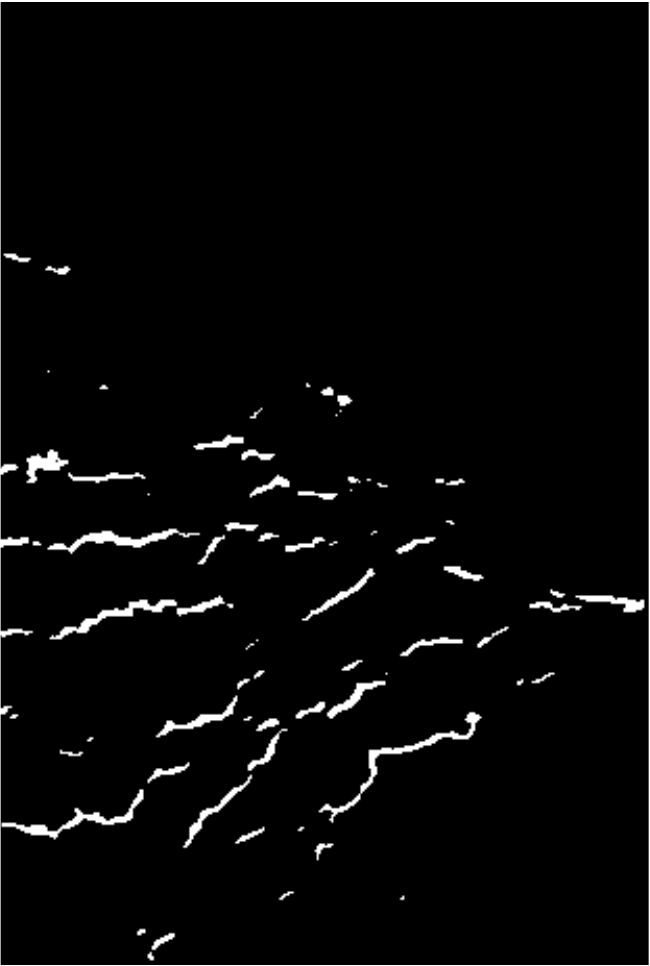}\\
~\\
\includegraphics[width=2.4cm]{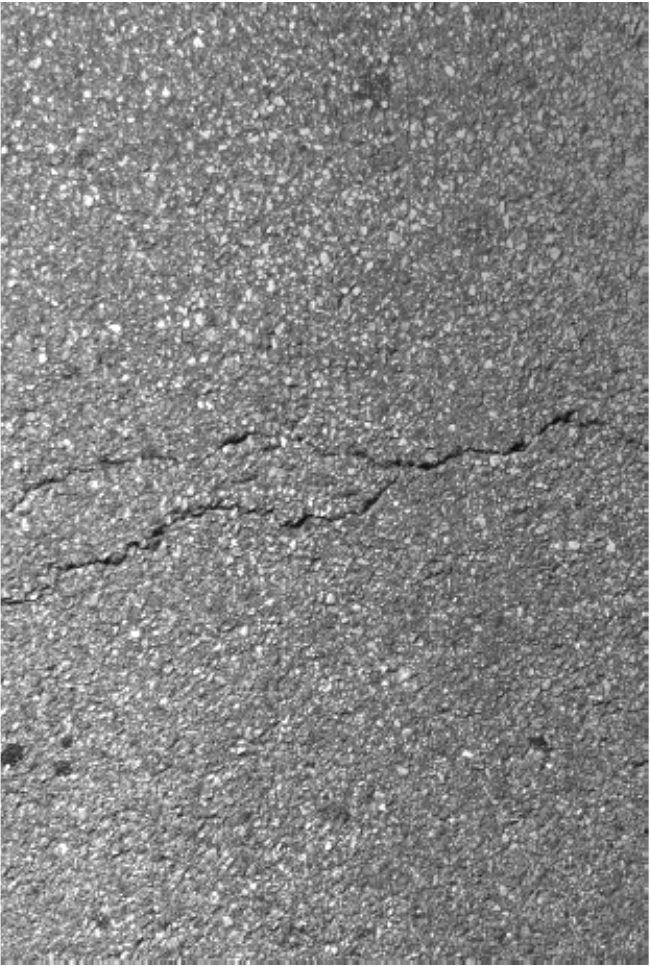}
\includegraphics[width=2.4cm]{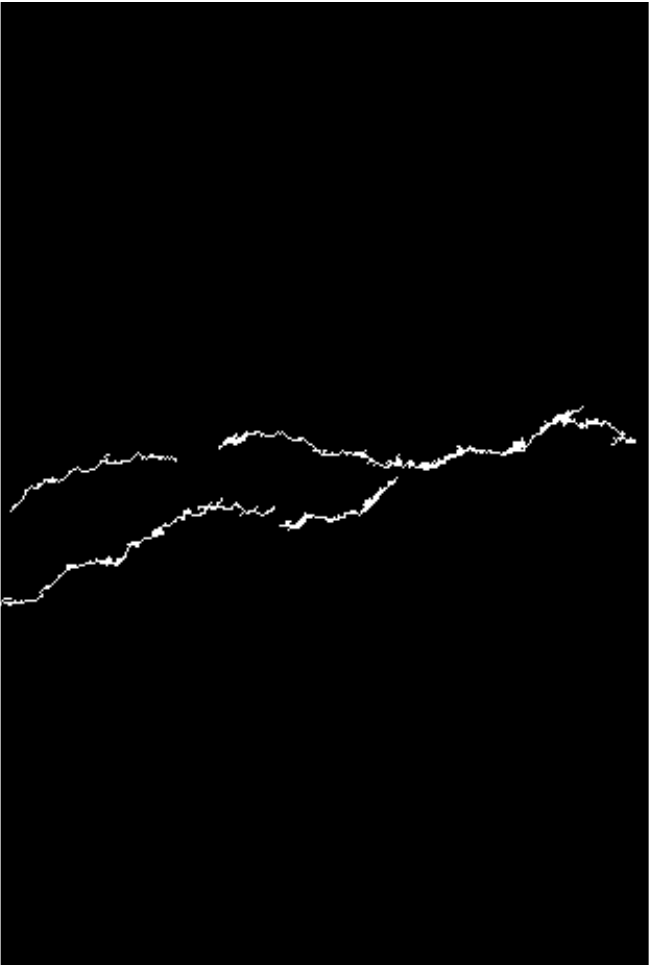}
\includegraphics[width=2.4cm]{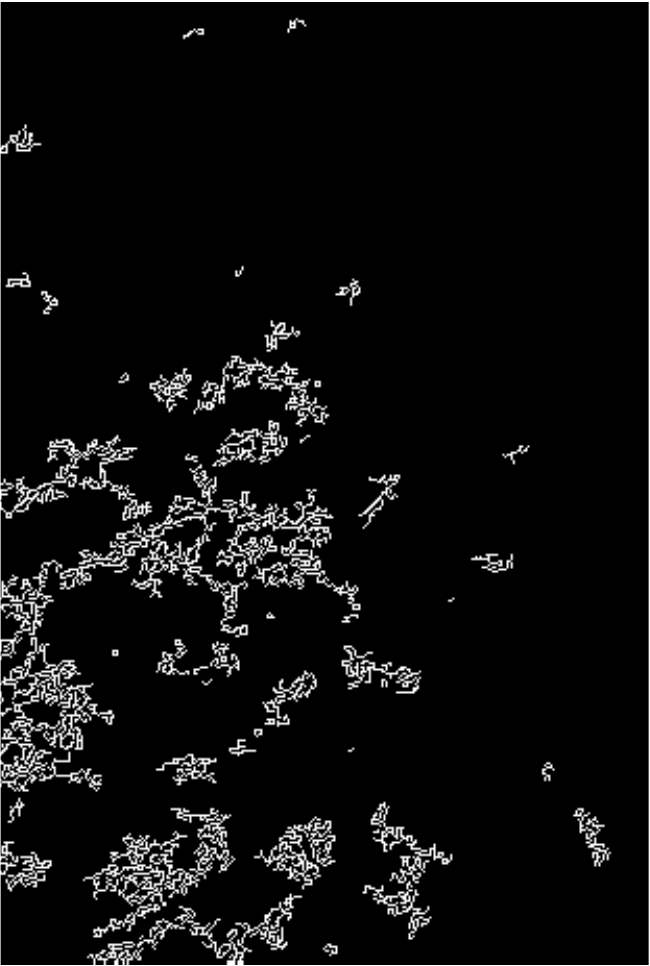}
\includegraphics[width=2.4cm]{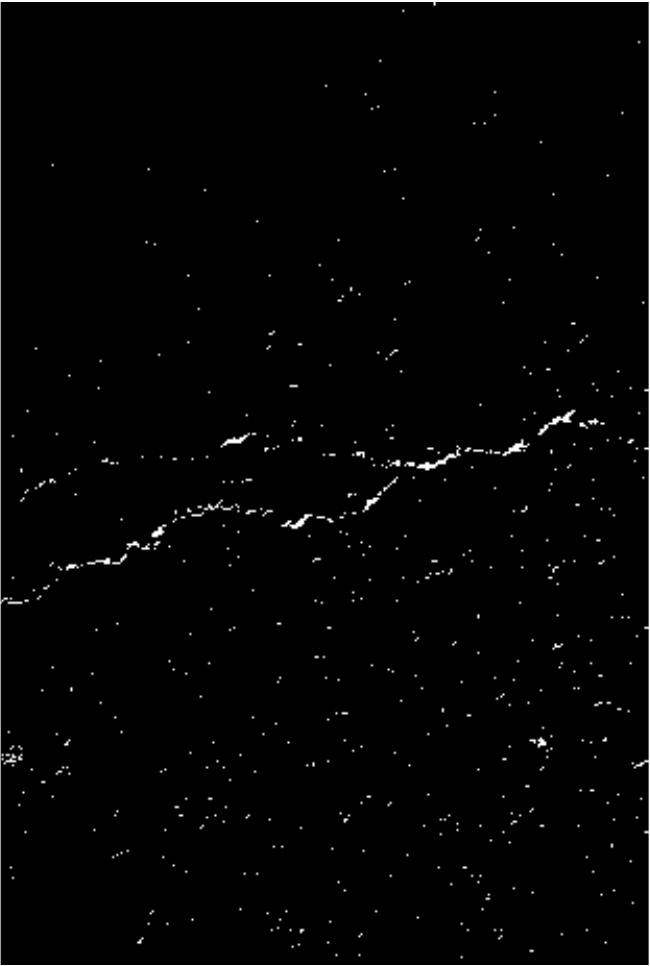}
\includegraphics[width=2.4cm]{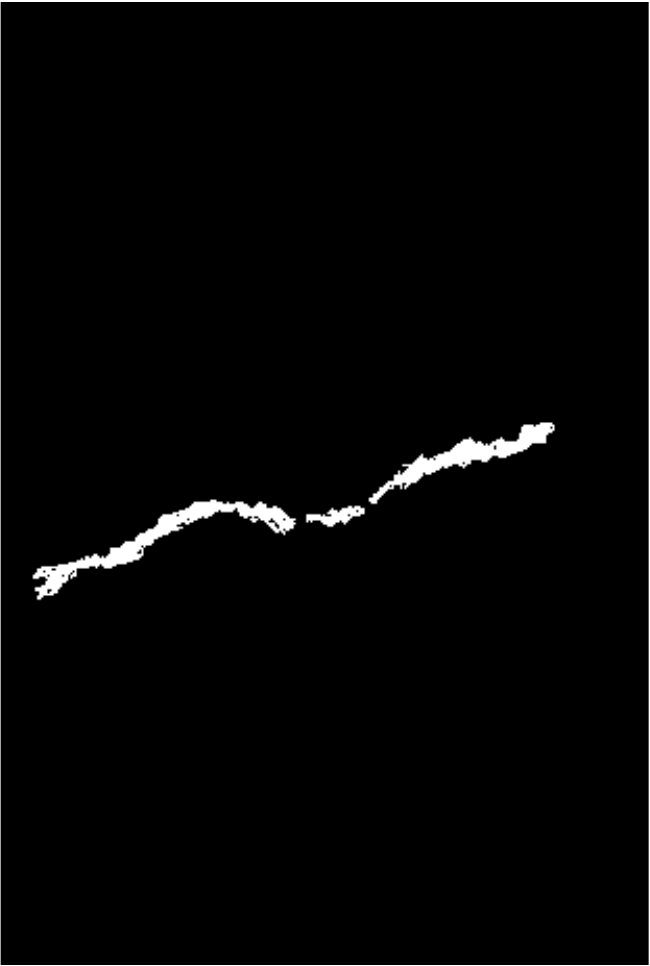}
\includegraphics[width=2.4cm]{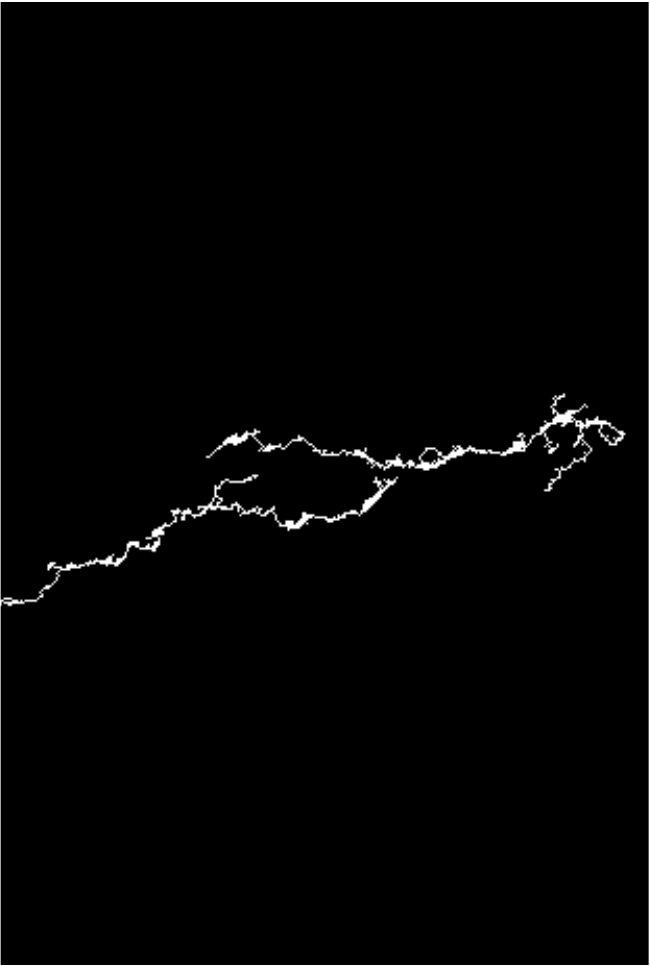}
\includegraphics[width=2.4cm]{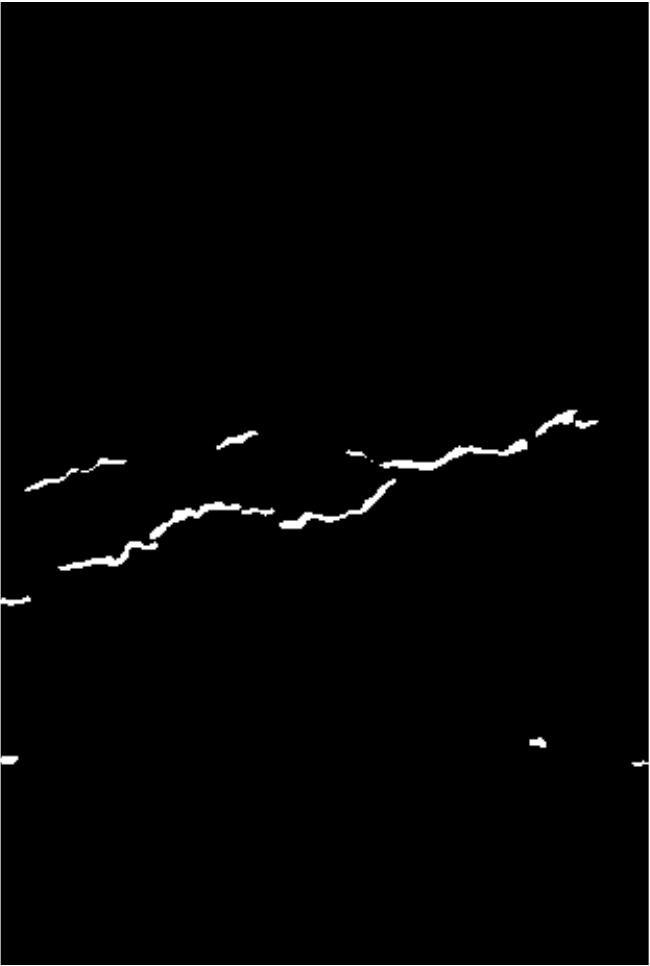}\\

\caption{Results of different methods on AigleRN (from left to right: original image, ground truth, Canny, local thresholding, FFA, MPS, the proposed method).}
\label{Fig. AigleRN}
\end{figure*}

\begin{table}[!t]
\renewcommand{\arraystretch}{1.3}
\caption{Crack Detection Results Evaluation on AigleRN}
\label{Table IV}
\centering
\begin{tabular}{c| c c c}
\hline
\space & $Pr$ & $Re$ & $F1$\\
\hline
Canny & 0.1989 & 0.6753 & 0.2881 \\
Local thresholding & 0.5329 & 0.9345 & 0.6670 \\
FFA & 0.7688 & 0.6812 & 0.6817 \\
MPS & 0.8263 & 0.8410 & 0.8195 \\
The proposed method & {\bfseries 0.9178} & {\bfseries 0.8812} & {\bfseries 0.8954}\\
\hline

\end{tabular}
\end{table}

%

\section{Output Structure}

 Generally, CNN is used for classification \cite{krizhevsky2012imagenet}. Typical classification problems contain single-label classification and multi-label classification. In a single-label classification problem, a sample can only be classified into one class. For example, in \cite{zhang2016road}, crack detection is modeled as a binary classification problem and a pixel is classified as a crack pixel or a non-crack pixel based on its neighbouring information. However, in a multi-label classification problem, a sample can be classified into more than one class. Structured prediction based on CNN, modeled as a multi-label classification problem, is quite a new idea \cite{liskowski2016segmenting}. The method utilizes the information of a patch to predict not only the centered pixel but the centered structure. This operation takes the spacial relations between pixels into account, i.e., a crack pixel is more likely to be neighboured by another crack pixel and vice versa.

 In this experiment, we explore the performance of the structured prediction. The size of the output structure is denoted by $s$. Results presented are obtained for $s=1,3,5,7$, i.e., the number of output units is set to 1, 9, 25, 49, respectively. Other configurations remain unchanged. Specifically, for $s=1$, the model is simplified to the pixel classification problem similar to \cite{zhang2016road}. The performance with different sizes of output structure is shown in Fig.~8.

\begin{figure}[!t]
\centering
\includegraphics[width=8cm]{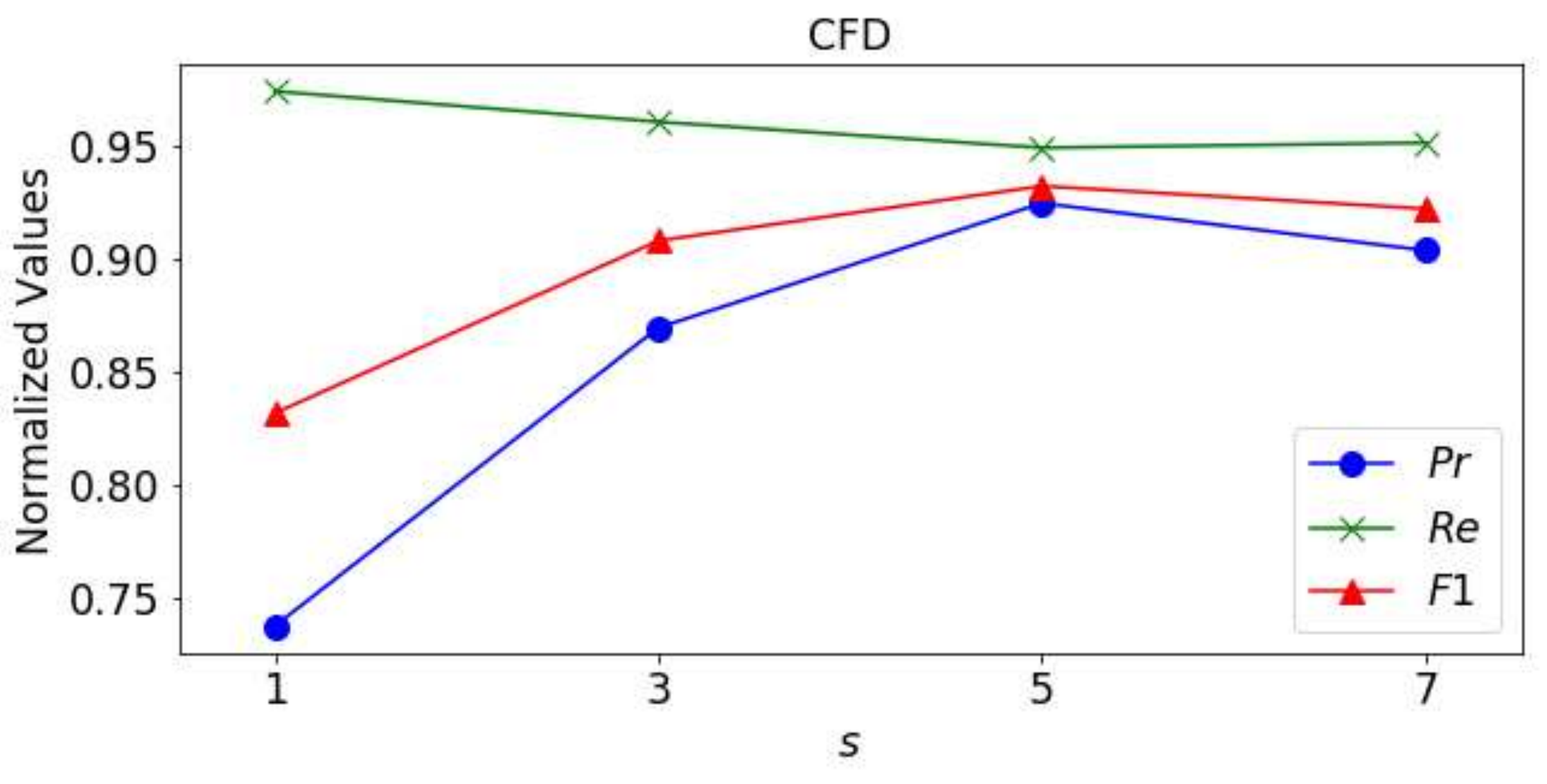}\\
\includegraphics[width=8cm]{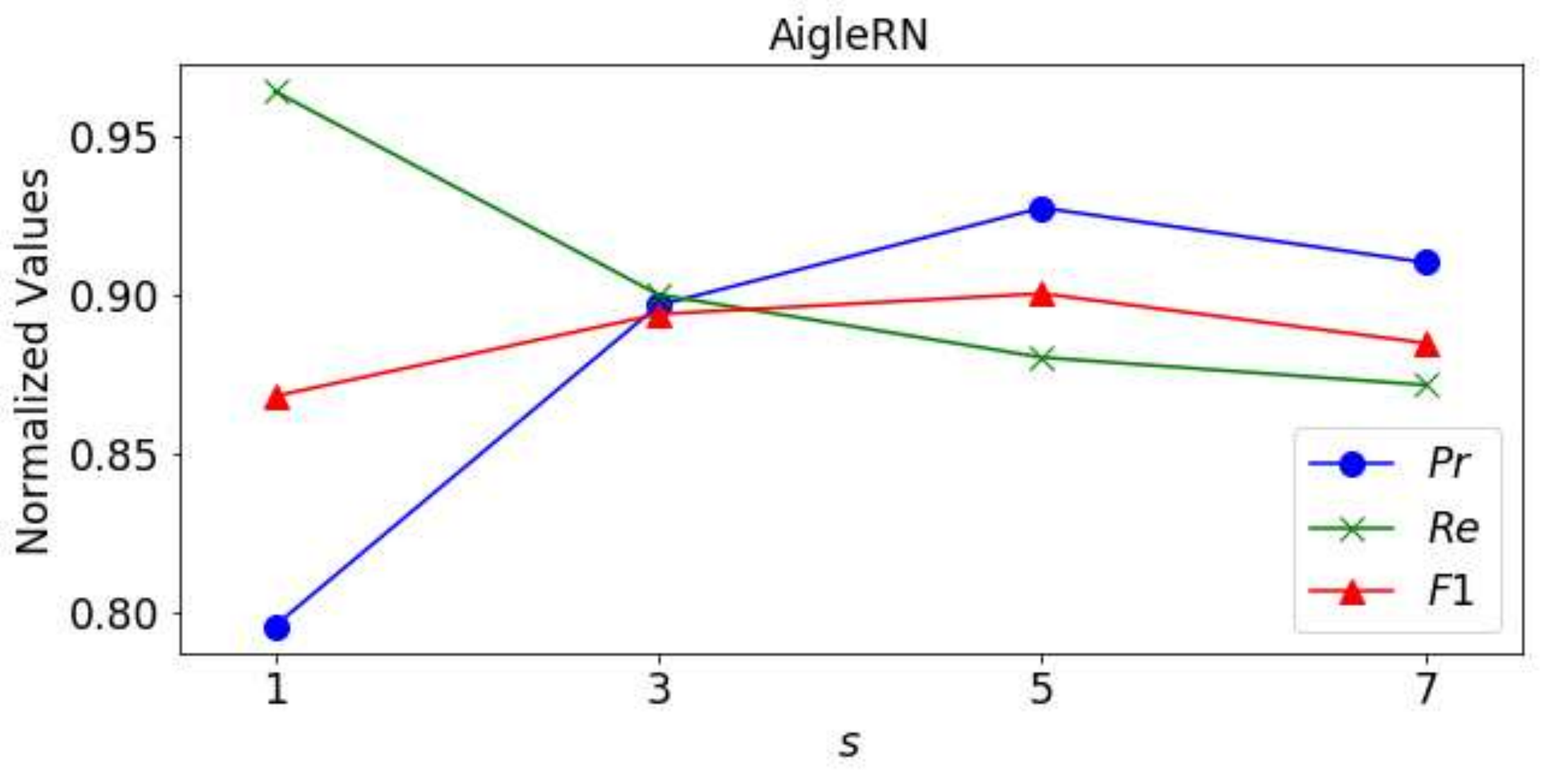}
\caption{Precision, recall and F1 score value variations with different sizes of output structure. Testing on CFD and AigleRN respectively.}
\label{Fig. structure list}
\end{figure}

From Fig.~8, it can be observed that all results for structured prediction, particularly for $s=5$, are better than that for pixel classification. To identify the difference between the structured prediction and the pixel classification, we present the probability maps for $s=1$ and $s=5$, as shown in Fig.~9. Both of them can detect the cracks but pixel classification is much more sensitive to noises. This is natural when considering the procedure of patches extraction and classification. Classifying a pixel based on its neighbouring information is the original design. However, the network is essentially classifying the patches in the procedure. Thus, limited by samples, it may be trapped in overfitting and misunderstand the classification problem. For example, if the mean of the extracted patches distinguishes between positive samples and negative samples, CNN may essentially turn to a classifier of patches based on the mean. The mean is a good feature for patches but is not enough to classify the pixel, since it is sensitive to the interference like non-uniform illumination. It is very confusing as we can not ensure what is CNN based on to classify the patches. Structured prediction can solve this problem to some extent, as it aims to predict several outputs at the same time. The outputs between patch to patch differ slightly based on the labels. Therefore, the network can not rely on single feature to accurately predict all outputs of a patch. This prevents the network from misunderstanding the problem.


\begin{figure}[!t]
\centering
\includegraphics[width=2.8cm]{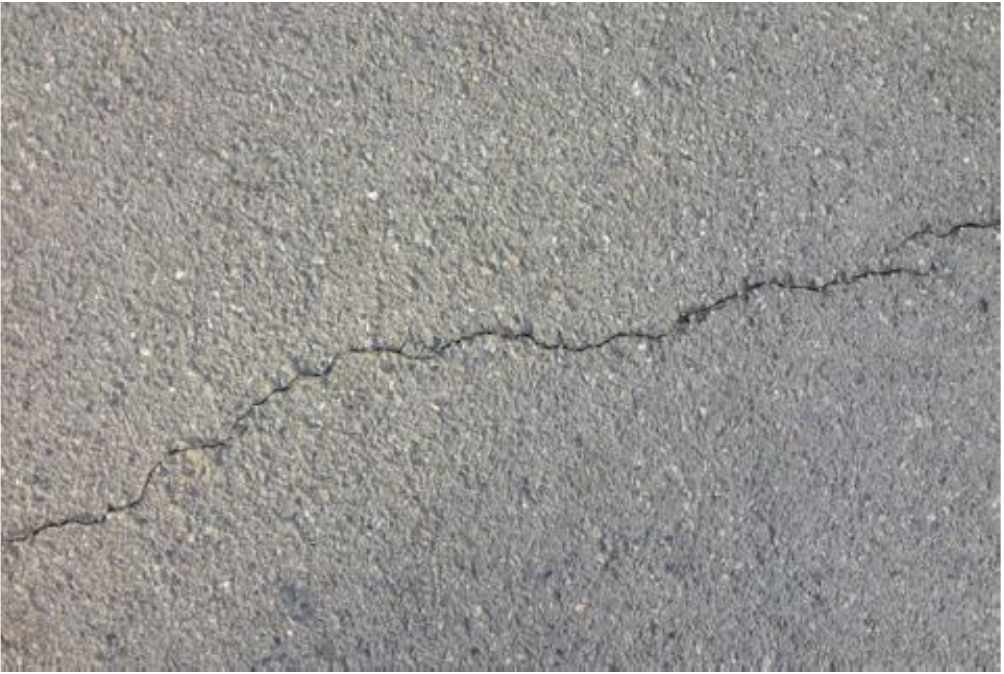}
\includegraphics[width=2.8cm]{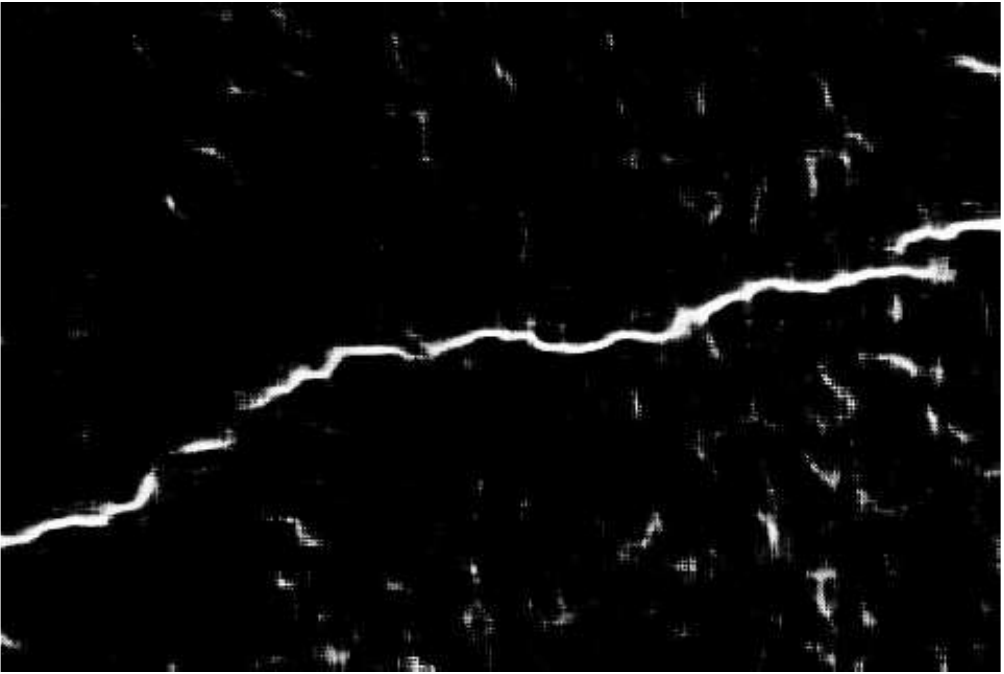}
\includegraphics[width=2.8cm]{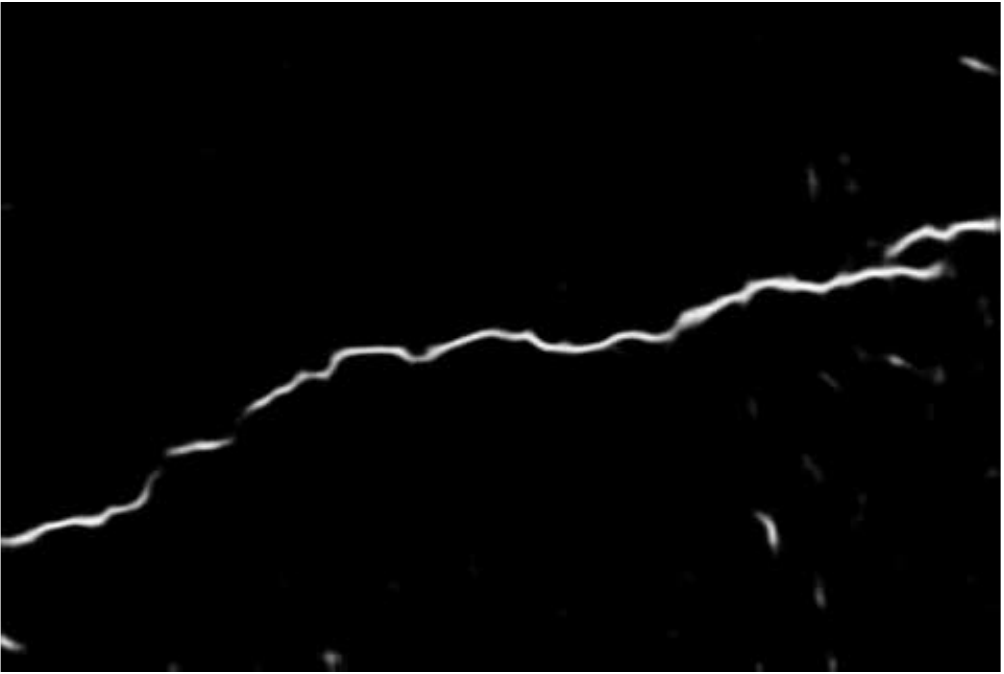}
\caption{Example probability maps with $s=1$ (middle) and $s=5$ (right) for the original image (left).}
\label{structure size image}
\end{figure}

Another explanation is for the testing procedure. In pixel classification, only one decision is obtained from the network for a pixel. However, in structured prediction, $s^2$ decisions can be obtained for a pixel by scanning all pixels through the whole image. These decisions are summed to obtain a more accurate result. Besides, normalizing the whole output image to [0,1] gives a global view to smooth the detected pixels.

\section{Ratio of Positive to Negative Samples}

Applying structured prediction directly can hardly obtain a satisfying result. As is shown in Table I, crack pixels are much fewer than non-crack pixels in typical crack images. The ratio of positive to negative samples is down to 1:65 in CFD, and even down to 1:98.5 in AigleRN. In such a classification task with severely imbalanced samples, the ratio of positive to negative has a great impact on the training results. To clarify the problem, assume that the natural ratio(i.e., 1:65) is used in training, if the predicted results of CNN are all negative, i.e., it predicts all pixels to be non-crack pixels, it can obtain a high accuracy as $65/(65+1)=0.985$. Thus if no treatment is conducted to deal with imbalanced data, the network prefers predicting negative results in training, which would lead to overestimation of the number of non-crack pixels in testing, as shown in Fig.~11.

Typically, in a multi-label problem, one output unit represents one class. Specifically in our method, an output unit represents a pixel on the particular position. For output units, a significant characteristic is that if a crack pixel appears, it is likely to find crack pixels on other units. Based on this characteristic, we introduce the terms \emph{positive sample} and \emph{negative sample} that used in binary classification, as we just care about the probabilities of crack pixels and non-crack pixels appearing. The particular position is not so important. According to the patch extraction procedure, we define an input extracted from a crack pixel as a positive sample, and an input extracted from a non-crack pixel as a negative sample. Experiments are conducted to make a thorough investigation on the influence of the ratio.

In the experiment, the total number of positive samples and negative samples are kept constant, and the ratio of positive to negative samples is modified. The ratio $R$ is defined as:
$$R = \frac{\mbox{negative samples}}{\mbox{positive samples}}$$
For example, if we keep the number of total samples to be 120, $R=1$ means 60 positive samples and 60 negative samples, while $R=2$ means 40 positive samples and 80 negative samples. We set the number of total training samples to be 360,000 for CFD and 60,000 for AigleRN. All training samples are generated from training dataset randomly, and we test the model with different training conditions on the whole testing dataset. The result is shown in Fig.~10.

\begin{figure}[!t]
\centering
\includegraphics[width=8cm]{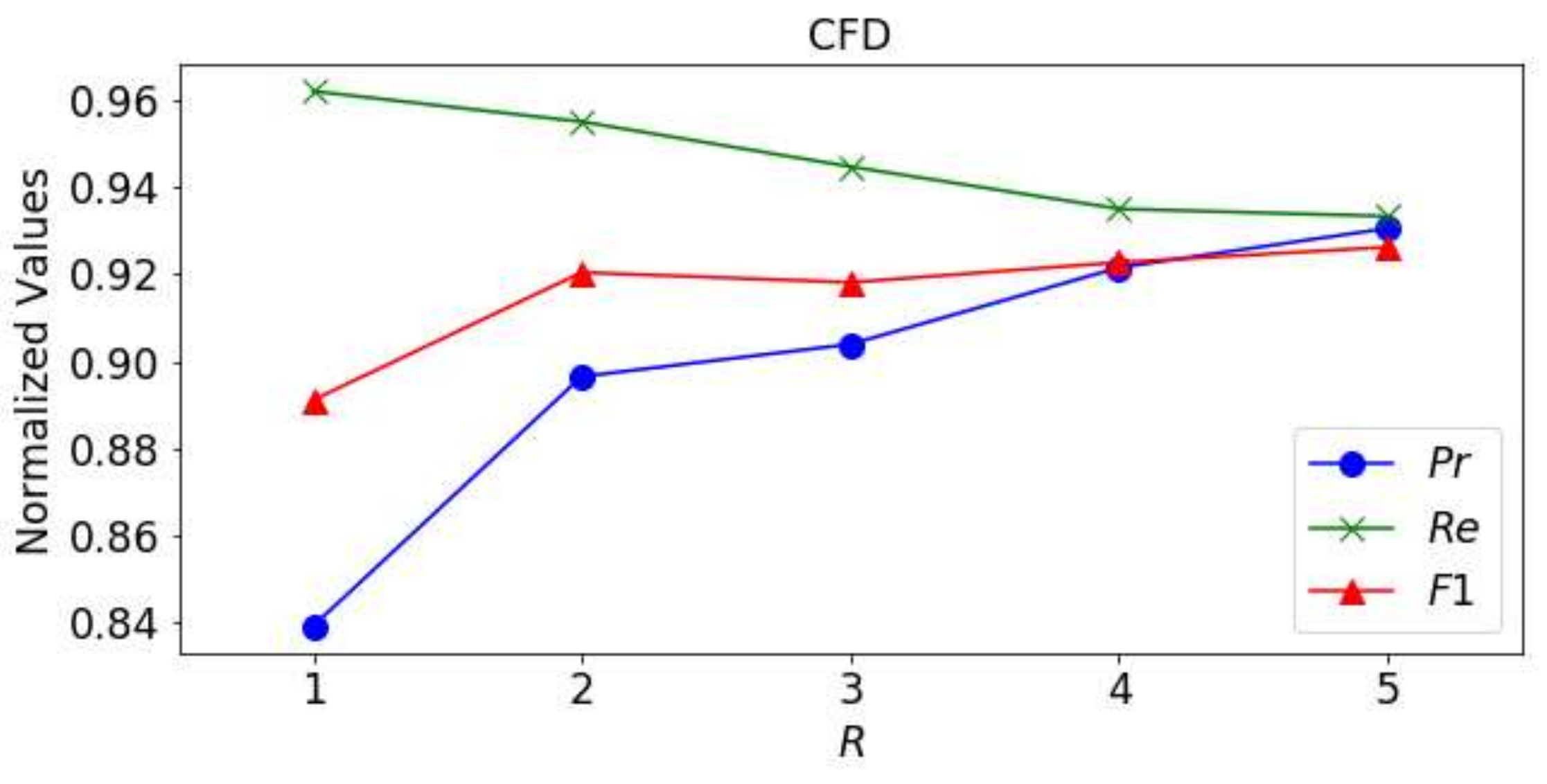}\\
\includegraphics[width=8cm]{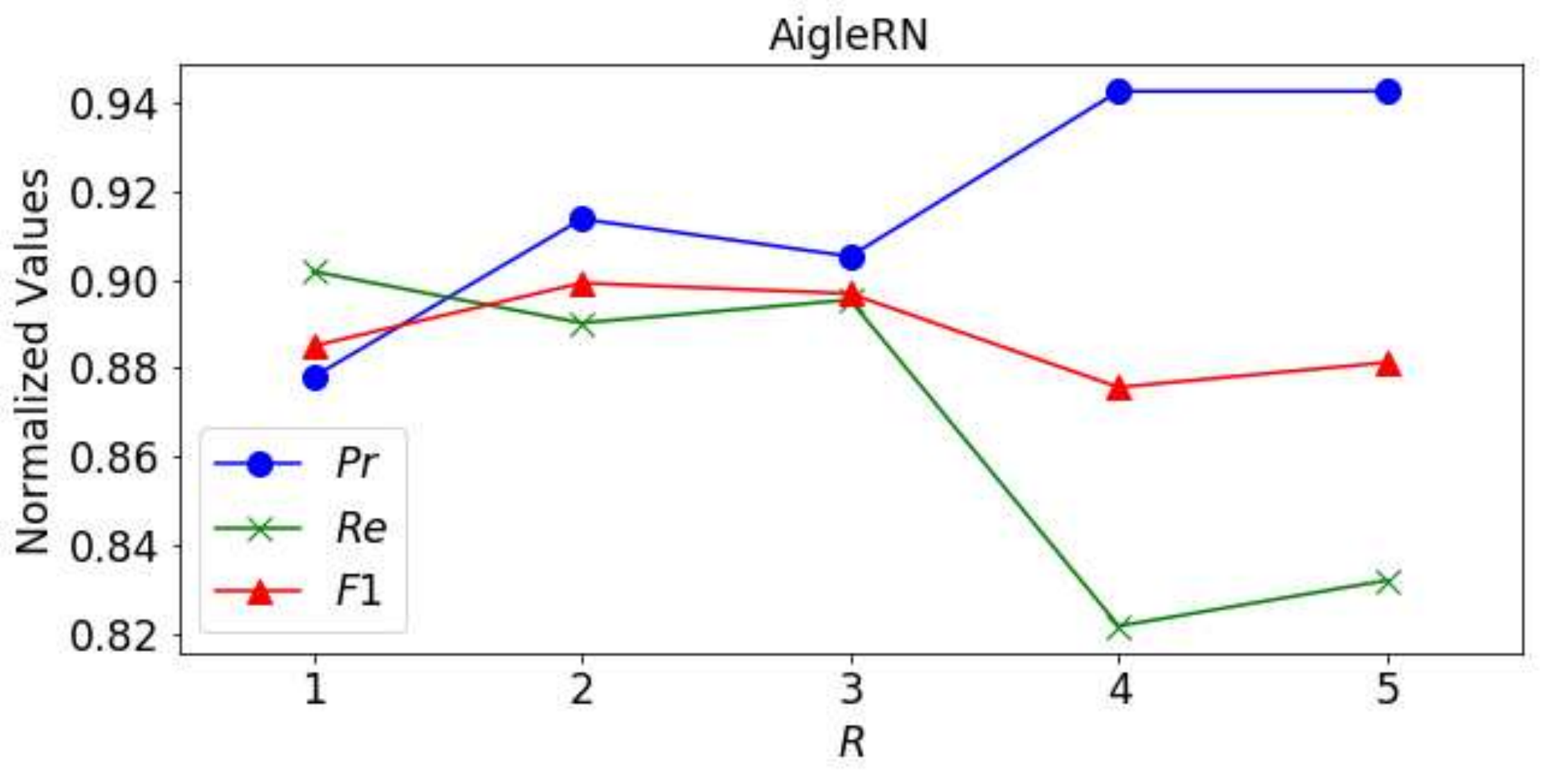}
\caption{Precision, recall and F1 score value variations with different ratios of positive to negative samples. Testing on CFD and AigleRN respectively.}
\label{Fig. ratio list}
\end{figure}

\begin{figure}[!t]
\centering
\subfigure[Original image] {\includegraphics[width=4.3cm]{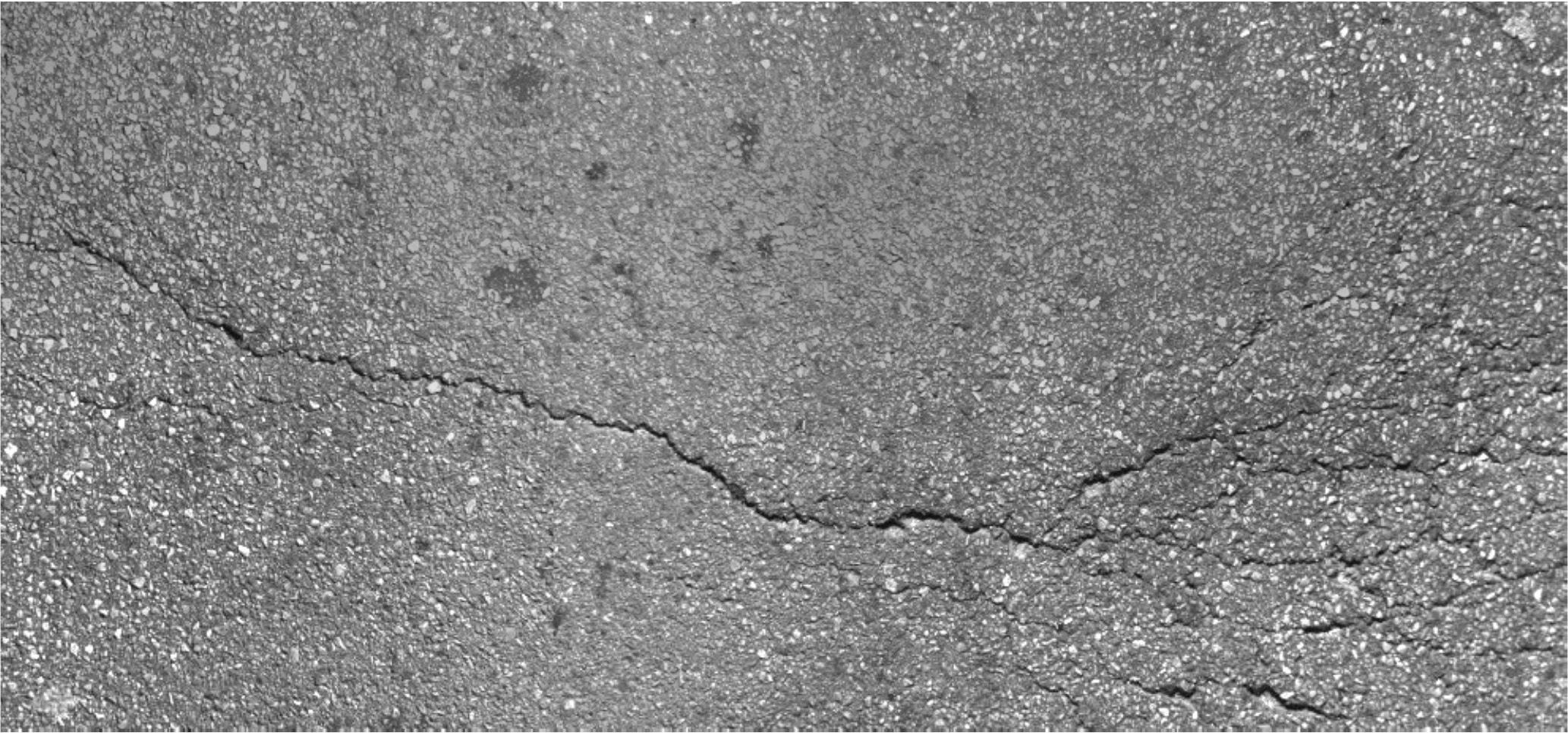}}
\subfigure[$R=1$] {\includegraphics[width=4.3cm]{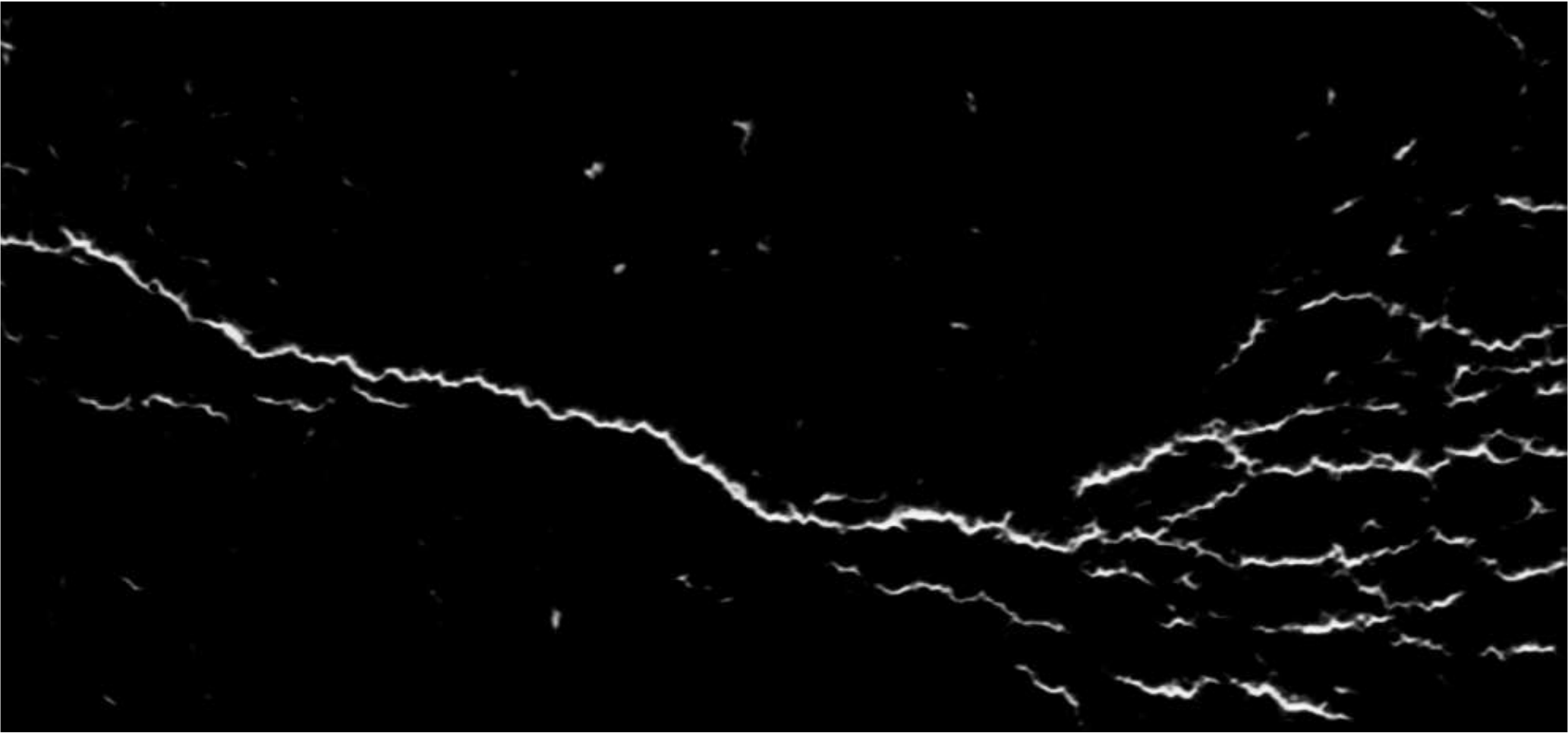}}\\
\subfigure[$R=3$] {\includegraphics[width=4.3cm]{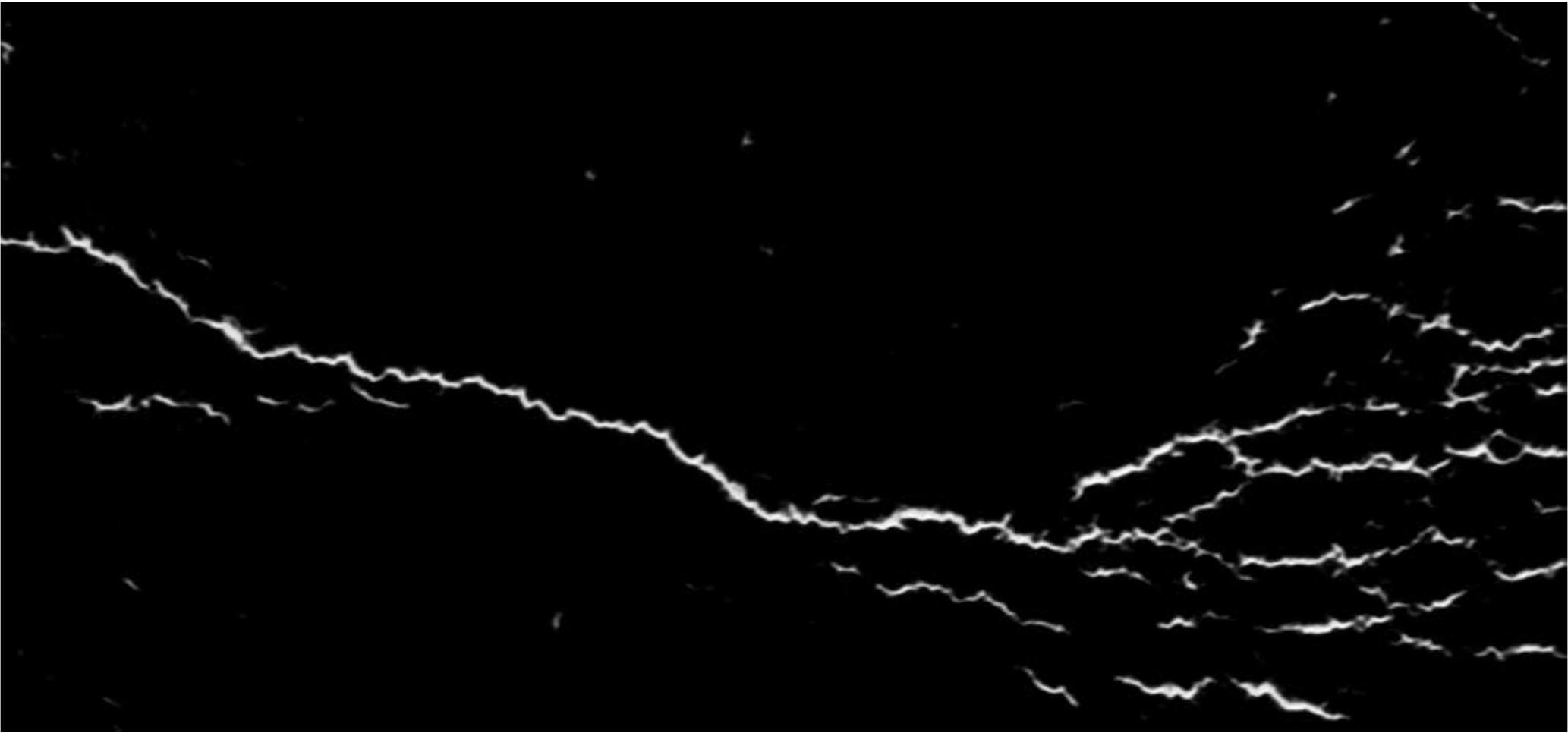}}
\subfigure[$R=5$] {\includegraphics[width=4.3cm]{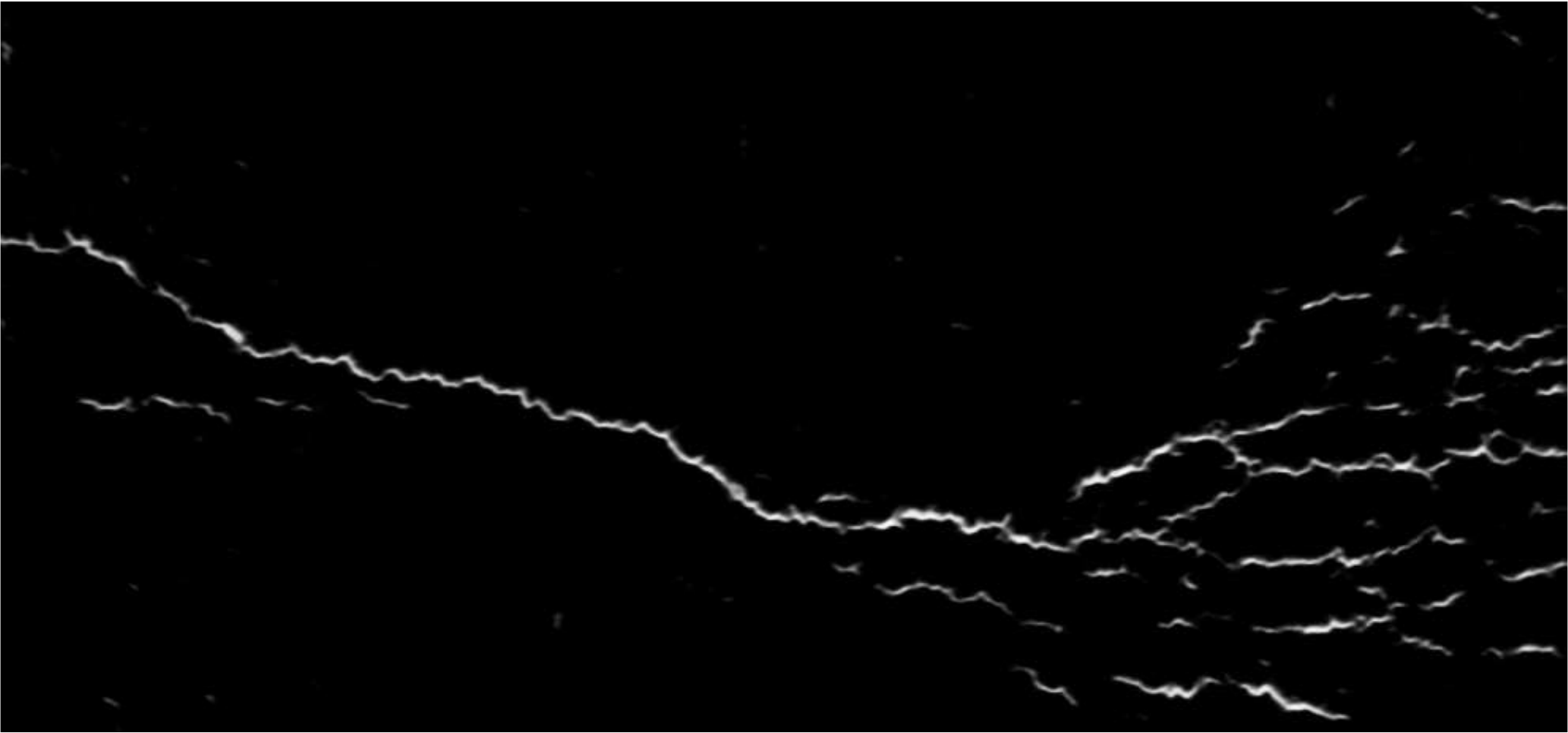}}\\
\subfigure[$R=59$] {\includegraphics[width=4.3cm]{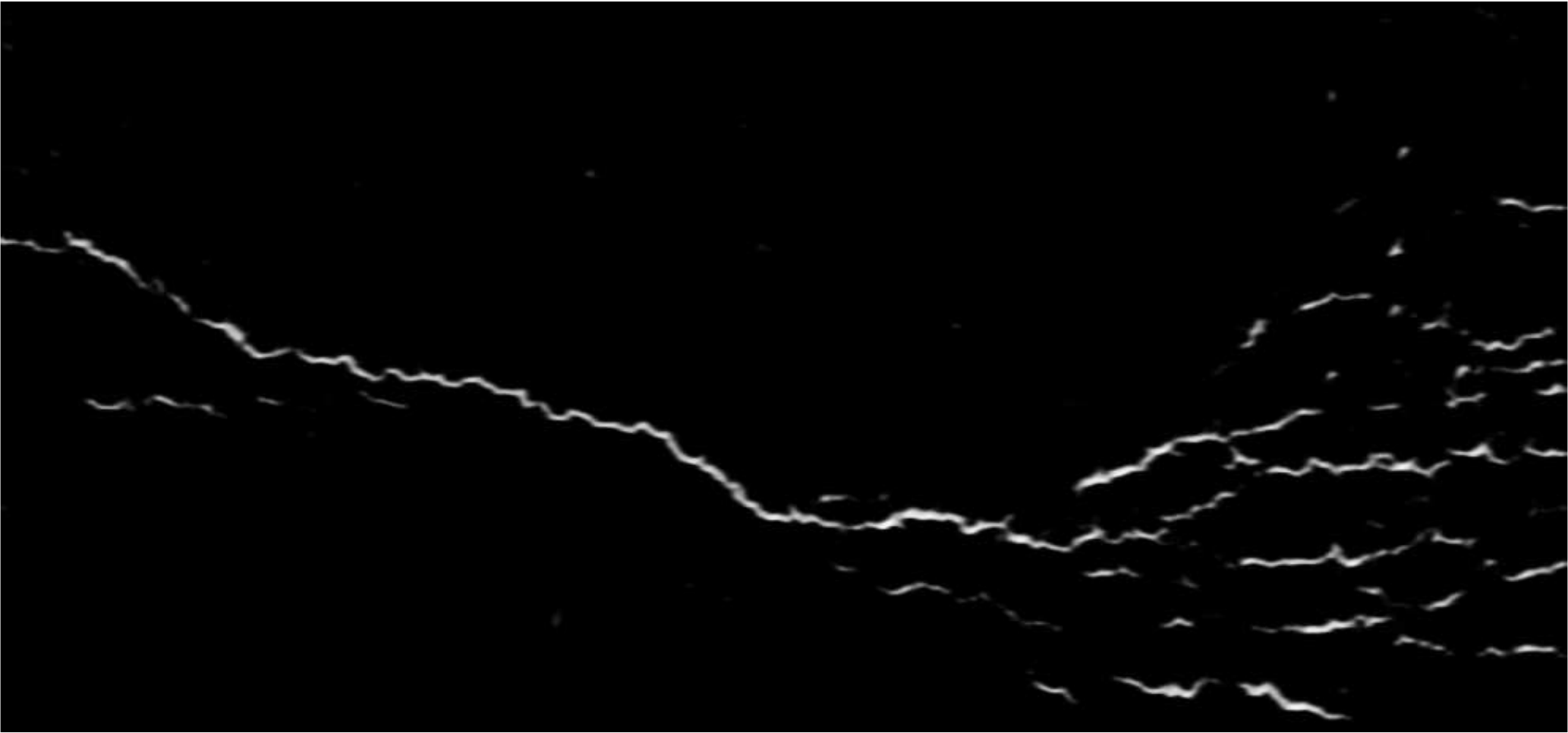}}
\subfigure[Natural($R=184$)] {\includegraphics[width=4.3cm]{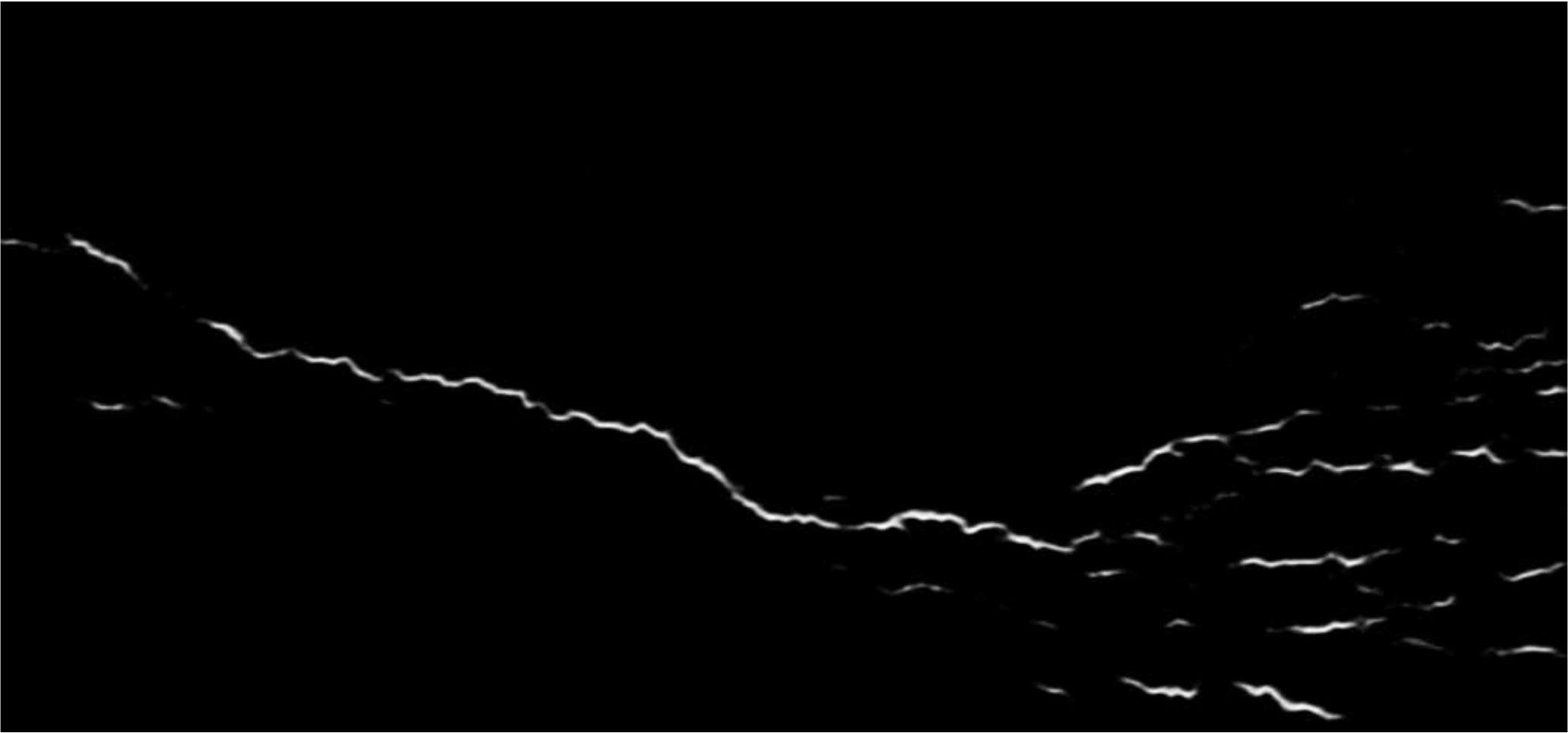}}
\caption{Results of the network with different ratios of positive to negative samples in training. \emph{Natural} means all samples are extracted randomly.}
\label{Fig. ratio examples}
\end{figure}

From Fig.~10, upward trend of precision and downward trend of recall can be clearly seen. The reason is that increasing negative samples in training would evidently increase the probability of negative prediction of the model, as is discussed above. This can be observed more clearly from Fig.~11. With higher $R$ in training, the model overestimate the number of the non-crack pixels, which leads to higher precision and lower recall. F1 score varies accordingly. Besides, precision and recall are computed according to manual labels, which are different from database to database. Since cracks in ground truth of CFD are wider than those of AigleRN generally, the best ratios for different database are different. From Fig.~10, the acceptable range of $R$ is $2 \leqslant R \leqslant 5$ for CFD and $2 \leqslant R \leqslant 3$ for AigleRN.

\begin{figure*}[!t]
\centering
\includegraphics[width=4.4cm]{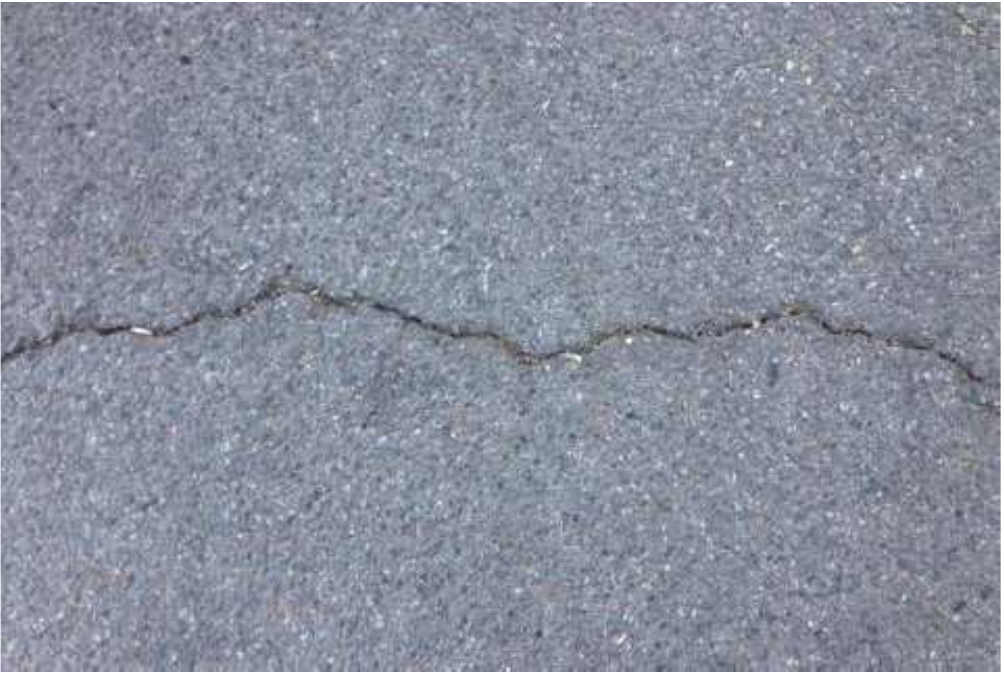}
\includegraphics[width=4.4cm]{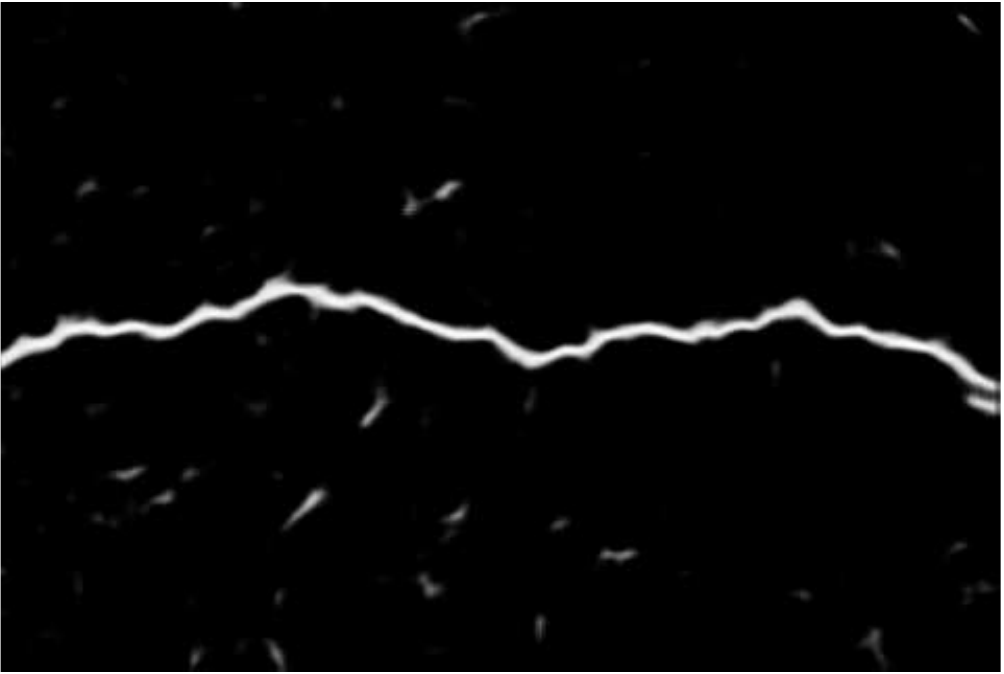}
\includegraphics[width=4.4cm]{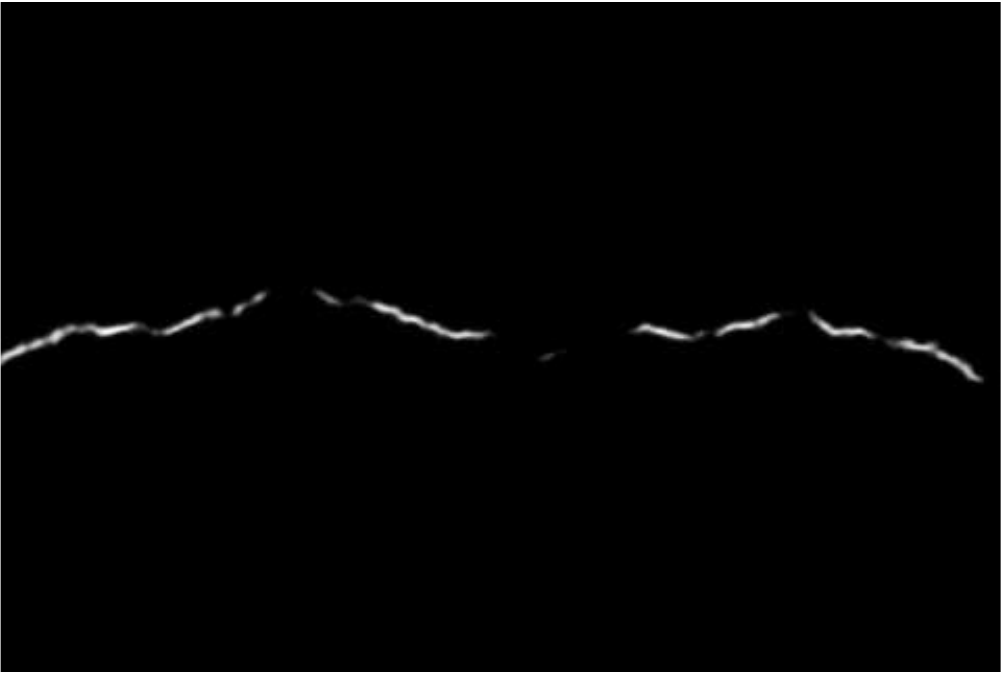}
\includegraphics[width=4.4cm]{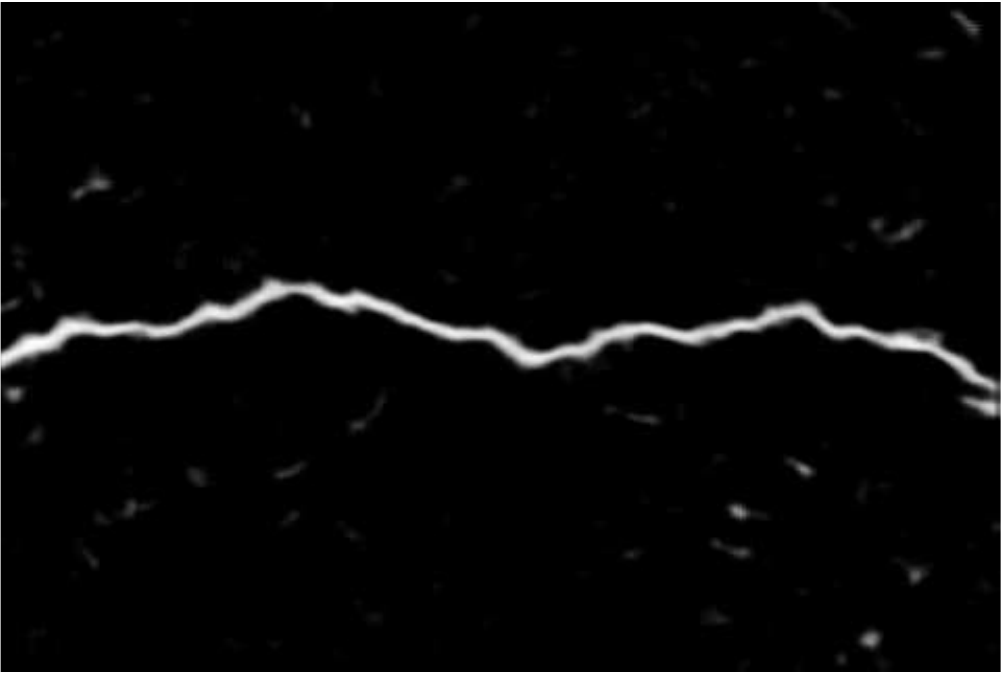}\\
~\\
\includegraphics[width=4.4cm]{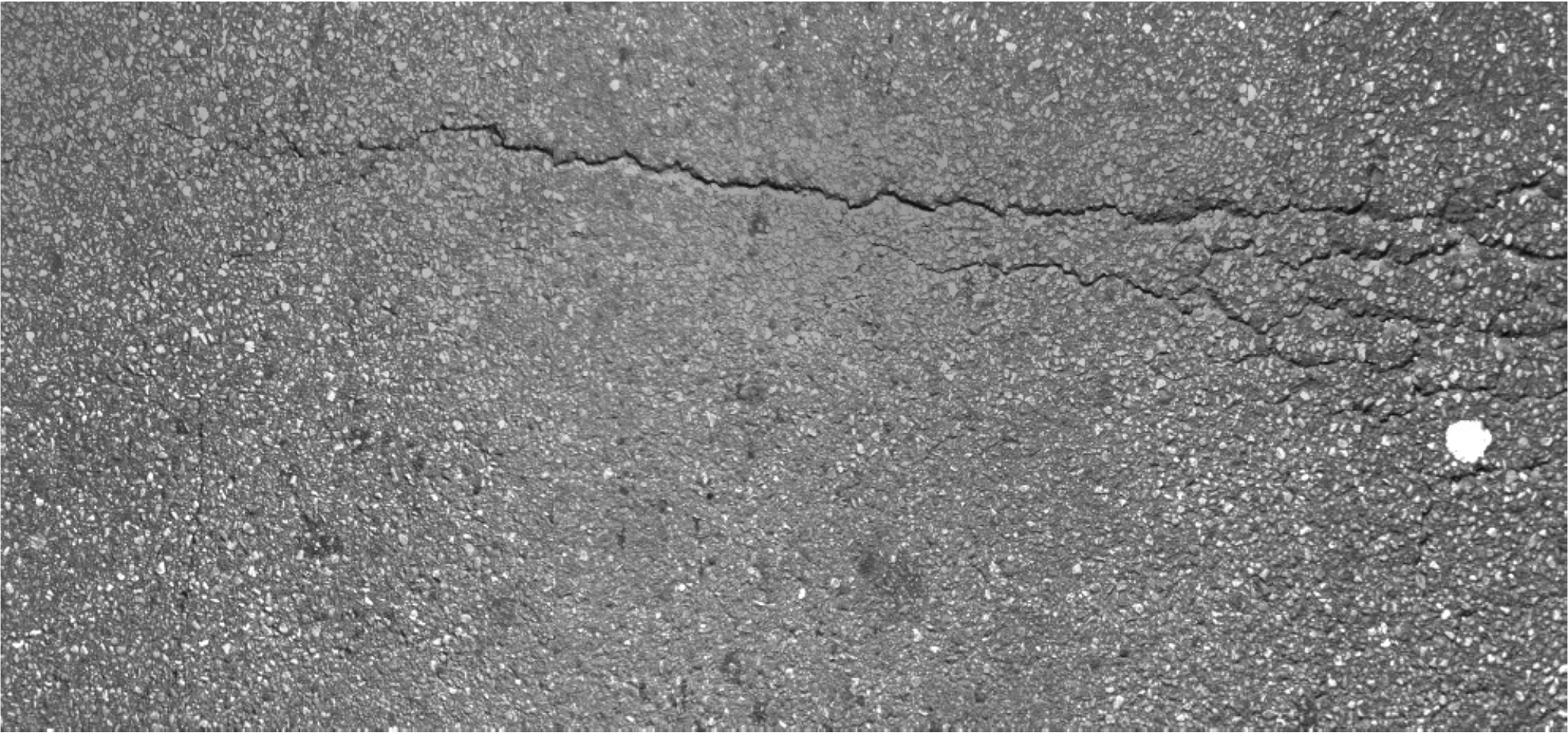}
\includegraphics[width=4.4cm]{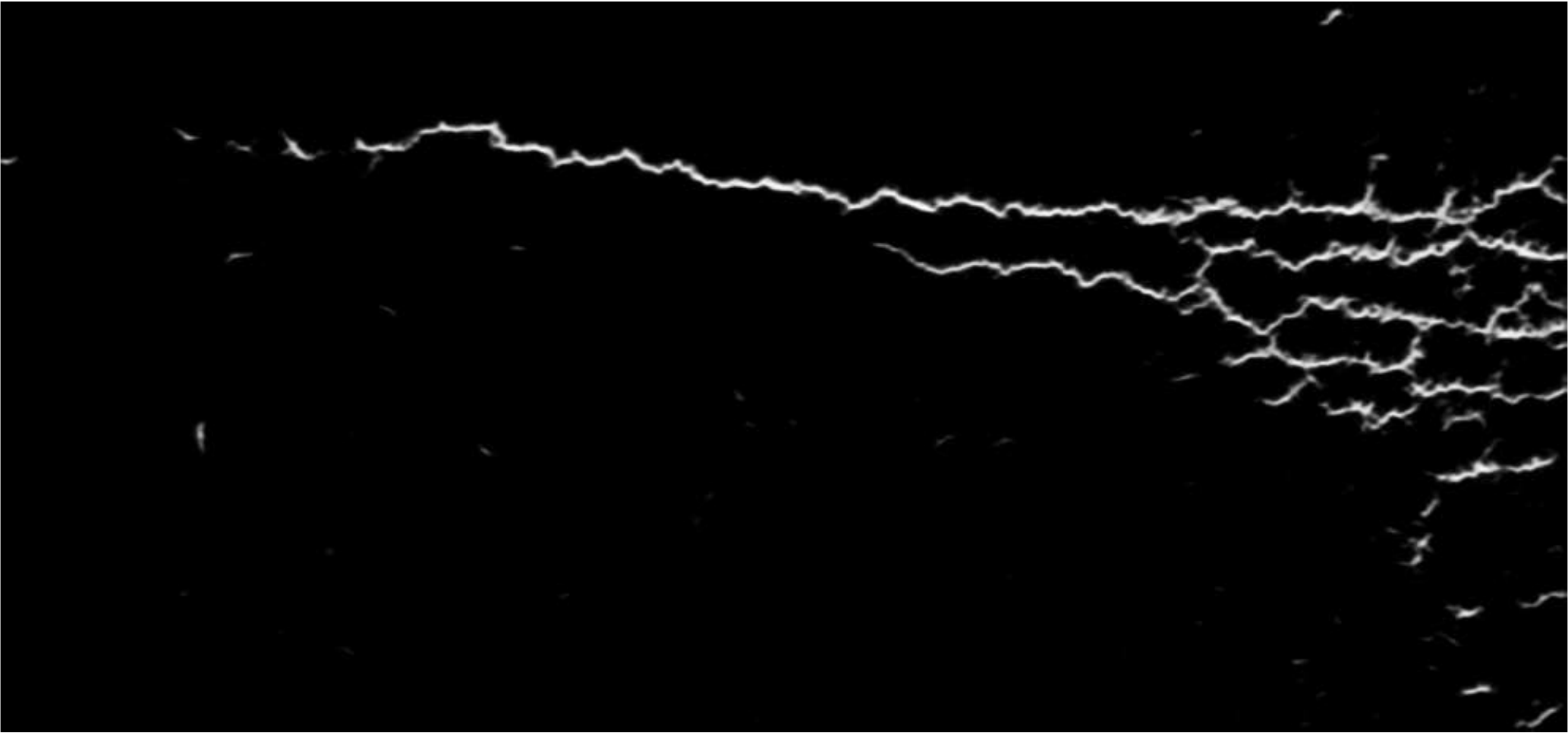}
\includegraphics[width=4.4cm]{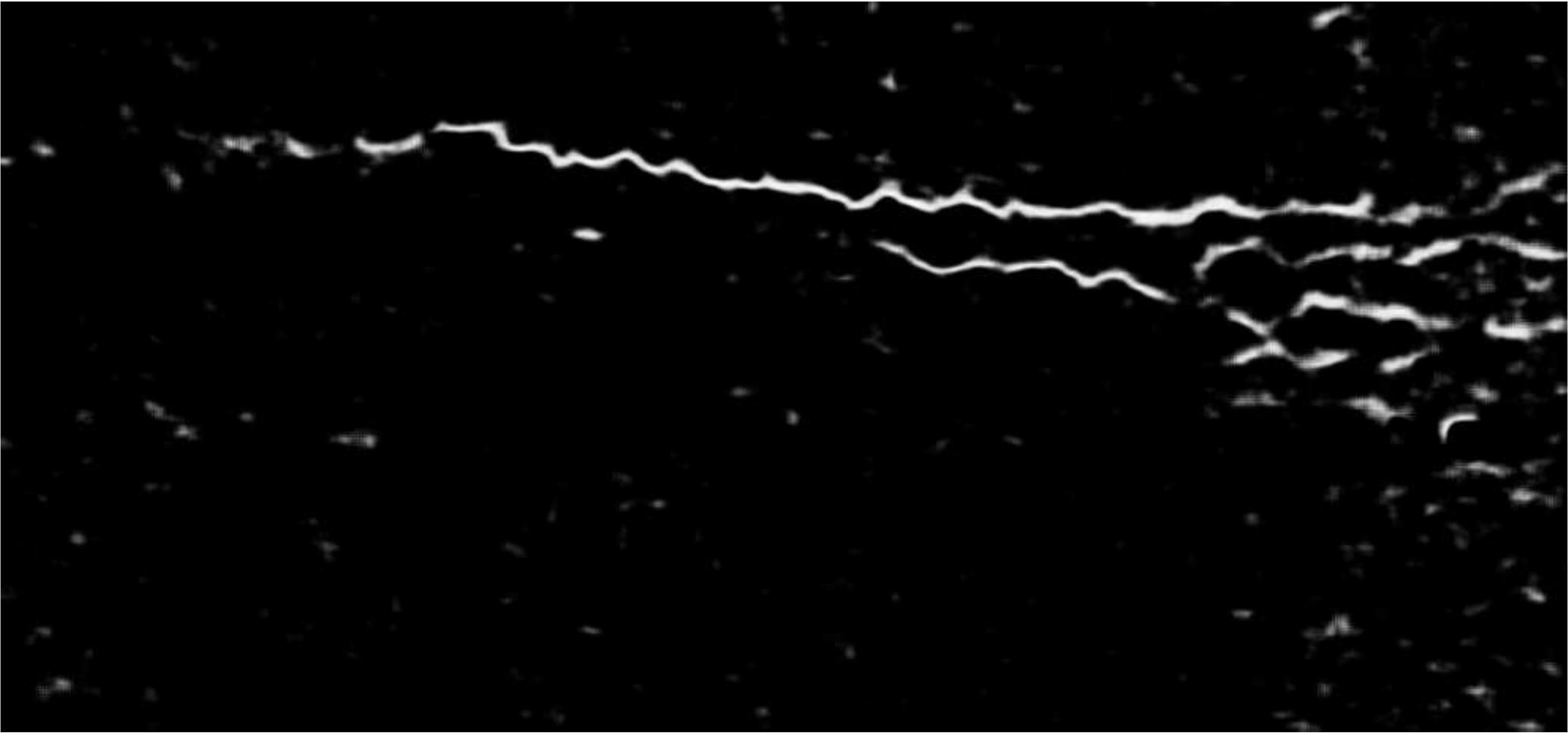}
\includegraphics[width=4.4cm]{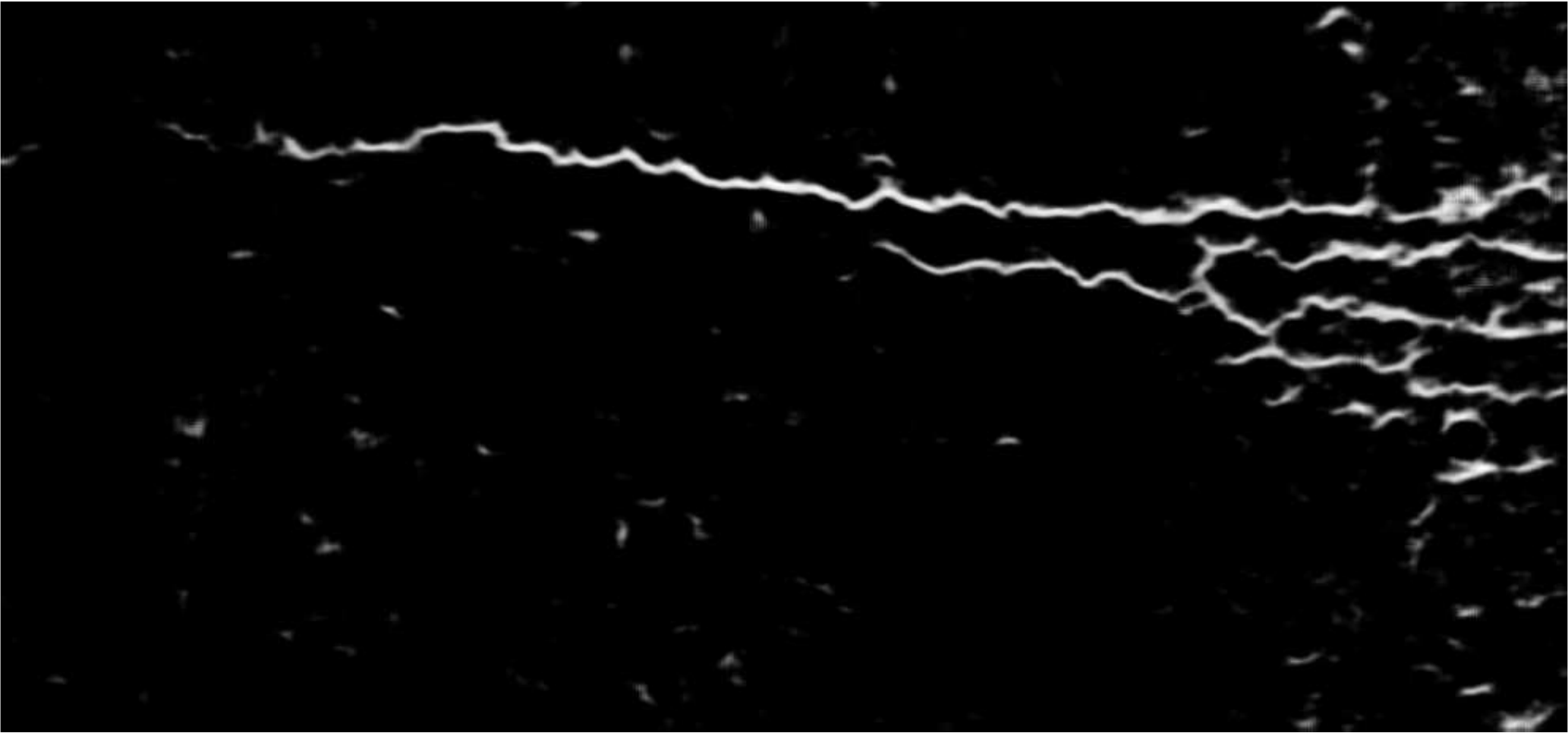}\\
\caption{Examples of cross database testing. From left to right on the top (training/testing): original image, CFD/CFD, AigleRN/CFD, Hybrid/CFD. From left to right on the bottom: original image, AigleRN/AigleRN, CFD/AigleRN, Hybrid/AigleRN. All outputs are probability maps.}
\label{Fig. cross testing}
\end{figure*}

\section{Generalization Study}

From results shown in Section III, we can see that the model works efficiently on CFD database and AigleRN database, respectively. In this section, the generalization of the model is discussed. In practical applications, different data acquisition systems bring different types of data. Besides, conditions of pavements differ from road to road. We conduct cross database testing on the two databases to confirm the validation of the model generalization.

We conduct three experiments: training on CFD and testing on AigleRN, training on AigleRN and testing on CFD, and training and testing on a hybrid database. In the first two experiments, the training and testing databases generated previously are used. For the hybrid database, we take half of both training databases to generate a new training database and test on the other images. The examples of results are shown in Fig.~12 and the evaluations are shown in Table V.

\begin{table}[!t]
\renewcommand{\arraystretch}{1.3}
\caption{Cross Database Testing}
\label{Table cross testing}
\centering
\begin{tabular}{c| c | c}
\hline
\diagbox{Train}{Test} & AigleRN & CFD\\
\hline
\multirow{3}*{AigleRN} & $Pr=0.9178$ & $Pr=0.9651$\\
& $Re=0.8812$ & $Re=0.3832$\\
& $F1=0.8954$ & $F1=0.4812$\\
\hline
\multirow{3}*{CFD} & $Pr=0.6488$ & $Pr=0.9119$\\
& $Re=0.8819$ & $Re=0.9481$\\
& $F1=0.7182$ & $F1=0.9244$\\
\hline
\multirow{3}*{Hybrid} & $Pr=0.9042$ & $Pr=0.9018$\\
& $Re=0.8448$ & $Re=0.9494$\\
& $F1=0.8677$ & $F1=0.9210$\\
\hline
\end{tabular}
\end{table}

It is observed that in cross testing, cracks detected by the model trained on AigleRN are very thin, leading to low recall and high precision, and the opposite is true for the model trained on CFD. In addition, a hybrid data training can make a trade-off. Moreover, results from hybrid data training suggest that with more data from different pavement conditions, CNN can learn better to predict cracks for different roads with the same architecture.

%

\section{Conclusion and Perspectives}

In this paper, an effective method based on CNN is proposed for pavement crack detection. Compared with other state-of-art methods, the proposed method shows a better performance in terms of dealing with different pavement texture. After training, the model can predict cracks close to the manual labels. This evidently shows that deep learning methods have potential for crack structure prediction. The network can learn from raw images without any preprocessing and output very satisfactory results.

With training and testing on different databases, we observe a good generalization of the model. With hybrid data training, the potential of a general model for different conditions of pavements may be brought out. Since CNN has a good learning ability for images, the network architecture can also be considered for other types of data, such as laser and 3D camera images, which are commonly obtained from systems of pavement distress detection.

Moreover, from cross testing we can see the influence of the manual labels to the detection results. As shown in experiments, thinner crack labels lead to thinner crack outputs. This is basically because labelling still limited by judgements of human experts. Our main perspective is to develop Semi-supervised or unsupervised learning methods to break through the limitations of human judgments.

%



%
%

\ifCLASSOPTIONcaptionsoff
  \newpage
\fi



\bibliographystyle{IEEEtran}
\bibliography{references}
\end{document}